\documentclass[sigconf]{acmart}  
\AtBeginDocument{%
  }

\copyrightyear{2026}
\acmYear{2026}
\setcopyright{cc}
\setcctype{by}
\acmConference[MM '26]{Proceedings of the 34th ACM International Conference on Multimedia}{November 10--14, 2026}{Rio de Janeiro, Brazil}
\acmBooktitle{Proceedings of the 34th ACM International Conference on Multimedia (MM '26), November 10--14, 2026, Rio de Janeiro, Brazil}
\acmDOI{10.1145/3767308.3835475}
\acmISBN{979-8-4007-2213-4/2026/11}

\acmSubmissionID{3214}

\usepackage{acronym}
\usepackage{bm}
\usepackage[table]{xcolor} 
\usepackage{multirow}      
\usepackage{algorithm}
\usepackage{algorithmicx}
\usepackage{algpseudocode}
\usepackage{colortbl}
\usepackage{overpic}
\usepackage{graphicx}  
\usepackage{pifont}    
\usepackage{makecell}  
\usepackage{gensymb}
\usepackage{placeins}
\usepackage{float}
\usepackage{cleveref}
\usepackage{subcaption}    
\usepackage{booktabs}      
\usepackage{cuted}
\usepackage{enumitem}
\usepackage{pbalance}

\acrodef{SSIM}[SSIM]{Structural Similarity Index Measure}
\acrodef{POV}[POV]{Point Of View}
\acrodef{GRU}[GRU]{Gated Recurrent Unit}
\acrodef{ODE}[ODE]{Ordinary Differential Equation}
\acrodef{OCP}[OCP]{Optimal Control Problem}
\acrodef{RK4}[RK4]{Runge-Kutta 4}

\newcommand{\gameplaytime}{140+}

\newcommand{\ourtodo}[1]{}

\newcommand{\scite}[1]{\cite{#1}}

\newif\ifarxiv
\makeatletter
\ifdefined\if@ACM@nonacm
    \if@ACM@nonacm
        \arxivtrue
    \else
        \arxivfalse
    \fi
\else
    \arxivfalse 
\fi
\makeatother

\ifarxiv
\else
\fi

\begin{document}

\title{TT4D: A Pipeline and Dataset for Table Tennis 4D Reconstruction From Monocular Videos}

\author{Nima Rahmanian}
\authornote{These authors contributed equally to this research.}
\affiliation{%
  \institution{University of California, Berkeley}
  \city{Berkeley}
  \state{CA}
  \country{USA}
}

\author{Daniel Kienzle}
\authornotemark[1]
\affiliation{%
  \institution{Universität Augsburg}
  \city{Augsburg}
  \country{Germany}
}

\author{Thomas Gossard}
\authornotemark[1]
\affiliation{%
  \institution{University of Tübingen}
  \city{Tübingen}
  \country{Germany}
}

\author{Dvij Kalaria}
\affiliation{%
  \institution{University of California, Berkeley}
  \city{Berkeley}
  \state{CA}
  \country{USA}
}

\author{Rainer Lienhart}
\affiliation{%
  \institution{Universität Augsburg}
  \city{Augsburg}
  \country{Germany}
}

\author{S. Shankar Sastry}
\affiliation{%
  \institution{University of California, Berkeley}
  \city{Berkeley}
  \state{CA}
  \country{USA}
}

\renewcommand{\shortauthors}{Nima Rahmanian et al.}

\begin{abstract} 
We present \textbf{TT4D}, a large-scale, high-fidelity table tennis dataset. 
It provides \gameplaytime{} hours of reconstructed singles and doubles gameplay from monocular broadcast videos, featuring annotations like high-quality camera calibrations, precise 3D ball positions, ball spin, time segmentation, and 3D human meshes over time. 
This rich data provides a new foundation for virtual replay, in-depth player analysis, and robot learning. 
The dataset's combination of scale and precision is achieved through a novel reconstruction pipeline. 
Prior methods first partition a game sequence into individual shot segments using the 2D ball track, and only then attempt reconstruction.
However, 2D-based time segmentation collapses under occlusion and varied camera viewpoints, preventing reliable reconstruction.
We invert this paradigm by first lifting the entire unsegmented 2D ball track to 3D through a learned lifting network.
This 3D trajectory then allows us to reliably perform time segmentation. 
The learned lifting network also infers the ball’s spin, handles unreliable ball detections, and successfully reconstructs the ball trajectory in cases of high occlusion. 
This lift-first design is necessary, as our pipeline is the only method capable of reconstructing table tennis gameplay from general-view broadcast monocular videos. 
We demonstrate the dataset’s fidelity through two downstream tasks: estimating the racket’s pose \& velocity at impact, and training a generative model of competitive rallies. 
The code and dataset can be found on our \href{https://nimsi123.github.io/tt4d/}{\textcolor{blue}{project website}}.
\end{abstract}    


\ifarxiv
\else
\fi

\begin{teaserfigure}
  \includegraphics[width=0.9\textwidth]{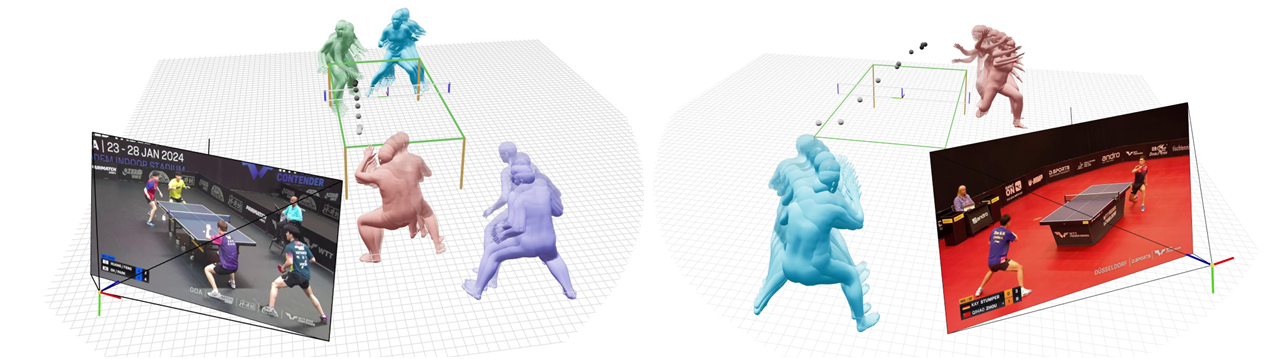}
  \Description{Two side-by-side 3D visualizations demonstrating the multimodal output of the Lift-First Pipeline. Each visualization shows a 2D broadcast video frame projected into a 3D grid space, originating from a camera coordinate axis. Within the 3D space, a green wireframe table tennis table is visible. A sequence of grey spheres maps the 3D ball positions across the table. Colored 3D human meshes are rendered with multiple overlapping translucent poses to illustrate their movement over time. The left visualization depicts a doubles match with four human meshes, while the right visualization depicts a singles match with two human meshes.}
  \caption{``Lift-First Pipeline" to generate TT4D, a \gameplaytime{} hour dataset that recovers camera parameters, 3D ball positions, ball spin, and 3D human meshes \textit{over time} from general-view broadcast videos.
  We lift full 2D ball tracks to 3D trajectories, bypassing fragile 2D-based time segmentation used in traditional methods.}
  \label{fig:teaser}
\end{teaserfigure}


\maketitle

\section{Introduction}
\label{sec:intro}
Online platforms host a vast and growing collection of high-quality competitive sports footage. This abundance of broadcast video makes 4D monocular-view reconstruction a scalable task enabling virtual replay, player analytics, and robot learning. \\
\noindent Table tennis, in particular, serves as a challenging testbed due to its high-speed and dynamic nature. A complete analysis goes beyond human mesh recovery and ball position reconstruction: it includes estimating ball spin, which strongly influences both flight (via the Magnus effect) and bounce behavior. 
Extracting these signals at scale from broadcast video is difficult: the ball is small, moves rapidly, and is routinely occluded by players. \\[0.66ex]
\noindent This constant occlusion makes time segmentation, the task of identifying the exact moment of a hit, the biggest challenge for reconstruction.
Existing pipelines ~\cite{etaat2025latte,gossard2025tt3d,ertner2024synthnet,kienzle2025towards,kienzle2026uplifting} follow a 2D-based strategy: first partition the sequence into shot segments using the 2D ball track, then lift each segment to 3D. 
Automated 2D-based segmentation, as in LATTE-MV~\cite{etaat2025latte} and TT3D~\cite{gossard2025tt3d}, often fails when the 2D track is interrupted by occlusions or corrupted by misdetections.
Manual segmentation ~\cite{kienzle2025towards, kienzle2026uplifting} allows for precise benchmarking but is not scalable, feasible only for small test sets. \\[0.66ex]
\noindent In this work, we introduce the \textbf{Lift-First Pipeline}, which fundamentally reverses this logic.
Our pipeline directly lifts the \textit{entire unsegmented} 2D ball track to a continuous 3D trajectory using a learned model.
Only then do we perform time segmentation in the unambiguous 3D domain.
Once this 3D trajectory is available, hit and bounce events can be reliably identified.
\\
Our Lift-First Pipeline is enabled by our core technical contribution: a novel \textbf{Full-Sequence Lifting Network}.
By taking a 3D-first approach, we can process long and complex unsegmented gameplay for the first time. 
This network is trained on a massive-scale synthetic dataset of 3 million rallies. \\[0.66ex]
We summarize our contributions as follows:
\begin{itemize}[leftmargin=*,itemsep=0pt,parsep=0pt,topsep=0pt,partopsep=0pt]
    \item \textbf{The ``Lift-First" Reconstruction Pipeline:} A new paradigm that decouples 3D reconstruction from fragile 2D-based time segmentation by first lifting the entire unsegmented sequence to 3D.
    \item \textbf{A Novel Full-Sequence Lifting Network:} The core technical method, enabled by a 3M points synthetic dataset. The network is the first to process unsegmented rallies and outputs the full 3D trajectory and dense per-frame 3D spin vectors.
    \item \textbf{The TT4D Dataset:} A \gameplaytime{} hour dataset generated with our pipeline, featuring precise 3D ball trajectories, 3D human meshes, and two annotations previously unavailable at scale: dense ball spin and robust 3D-derived time segmentations.
    \item \textbf{Novel Downstream Applications:} We demonstrate our dataset's high fidelity by (a) introducing a new racket stroke estimation method that recovers the racket's pose and velocity at impact and (b) training a generative Flow Matching model on competitive gameplay.
\end{itemize}
We will release the TT4D dataset, paving the way for new applications in computational sports science and robotics \cite{etaat2025latte,wang2012probabilisticmodelling,wang2017anticipatoryactionselection}.
\section{Related Work}
\label{sec:related_work}
\textbf{2D ball tracking} in table tennis is challenging due to its small size, fast motion, frequent occlusions, and motion blur. 
Recent methods rely on deep detectors~\cite{huang2019tracknet, vanzandycke2019realtime, komorowski2019deepball, sun2020tracknetv2}, with the Multiple-Input Multiple-Output (MIMO) formulation from TrackNetV2~\cite{sun2020tracknetv2} being a key breakthrough.
This MIMO strategy has been adopted by subsequent works using different backbones~\cite{tarashima2023wasb, liu2022monotrack, chen2024tracknetv3, kienzle2024segformer, kienzle2026uplifting}.
Attention mechanisms have also been incorporated to enhance temporal feature fusion~\cite{hu2018squeeze, gossard2025blurball, raj2024tracknetv4}. \\[0.66ex]
\noindent Methods for \textbf{3D trajectory lifting} can be split into physics-based optimization and learned networks.
Optimization-based methods like TT3D~\cite{gossard2025tt3d}, LATTE-MV~\cite{etaat2025latte} and MonoTrack~\cite{liu2022monotrack} minimize reprojection error.
These non-learning methods are inherently limited to the ``Traditional Pipeline" and hence do not scale.
Their optimization is already unstable for single, shot segments; extending them to a ``3D-first" approach that jointly solves for the trajectory, all unknown bounce points, and all unknown hit-points is computationally infeasible.
This necessitates a learning-based approach to make the Lift-First Pipeline viable.
Recent learned approaches train a network to lift 2D tracks to 3D trajectories.
SynthNet~\cite{ertner2024synthnet} and \cite{ponglertnapakorn2025where} both tackle this for tennis.
Most relevant is the work of \citet{kienzle2025towards}, which proposes a transformer that lifts single shot segments from 2D to 3D with zero-shot generalization from synthetic to real data.
This was extended in~\cite{kienzle2026uplifting} and serves as the basis for our network.
However, all these methods still depend on the ``Traditional Pipeline" and its dependence on fragile 2D-based time segmentation.
We bypass this limitation and develop a network that can lift full rallies from 2D to 3D. \\[0.66ex]
\noindent\textbf{Spin estimation} methods include indirect inference from trajectories or direct logo tracking~\cite{tebbe2020spin}.
Direct tracking has been improved with custom dot patterns~\cite{gossard2023spindoe} and event cameras that mitigate motion blur~\cite{gossard2024event, nakabayashi2024event}.
These hardware-specific methods are complemented by works that classify spin from player stroke motion~\cite{kulkarni2021table, fujihara2025stroke}.
Most recently, \citet{kienzle2025towards, kienzle2026uplifting} showed that 2D-3D lifting transformers can also regress the initial 3D spin vector.
We adapt this to predict per-frame spin for an entire unsegmented rally. \\[0.66ex]
Specialized table-tennis \textbf{datasets} have only emerged recently.
Blurball~\cite{gossard2025blurball} provided blur-aware 2D annotations, while TT3D~\cite{gossard2025tt3d} offered precise 3D trajectories from a multi-camera setup.
Synthetic datasets of individual shot segments have been used for model training~\cite{kienzle2025towards, kienzle2026uplifting}, along with small-scale real-world sets with topspin/backspin annotations~\cite{kienzle2025towards, kienzle2026uplifting}.
TTNet~\cite{voeikov2020ttnet} and P2ANet~\cite{bian2024p2anet} primarily focus on event spotting and fine-grained action detection within broadcast or multi-task video contexts.
Most similar to our TT4D dataset in scale is LATTE-MV~\cite{etaat2025latte}, which reconstructed $26$ hours of gameplay using the ``Traditional Pipeline''.
Our TT4D dataset, enabled by our Lift-First Pipeline, surpasses this by an order of magnitude.
Beyond scale, TT4D provides higher-fidelity data, using an improved camera calibration method and providing realistic 3D trajectories, in contrast to LATTE-MV's simplified parabolic fits.
Crucially, it is the first to provide two key annotations at this scale: dense, per-frame 3D spin and robust 3D-derived time segmentation, which are reliable even when 2D occlusions break prior methods.

\begin{figure*}[t]
    \centering
    \includegraphics[width=0.96\linewidth]{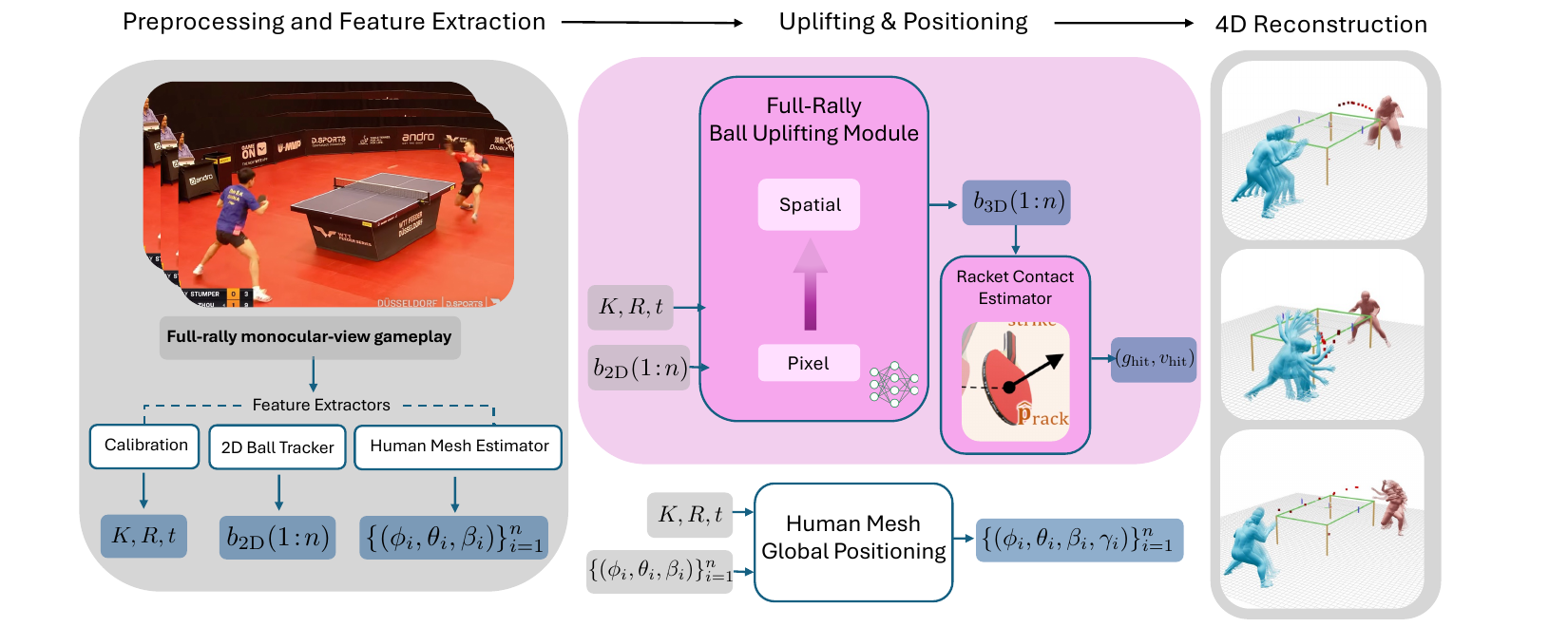}
    \Description{A block diagram illustrating the three stages of the Lift-First Pipeline: Preprocessing and Feature Extraction, Uplifting and Positioning, and 4D Reconstruction. In the first stage, a broadcast video of a table tennis match is processed by three feature extractors: Calibration (outputting $K, R, t$), 2D Ball Tracker (outputting $b_{2D}(1:n)$), and Human Mesh Estimator (outputting $\{(\phi_i, \theta_i, \beta_i)\}_{i=1}^n$). In the second stage, a Full-Rally Ball Uplifting Module takes the 2D ball track and calibration data through a neural network to produce a 3D trajectory $b_{3D}(1:n)$, which then feeds into a Racket Contact Estimator to estimate racket hit parameters. Concurrently, a Human Mesh Global Positioning module combines the mesh parameters and calibration data to output globally positioned meshes $\{(\phi_i, \theta_i, \beta_i, \gamma_i)\}_{i=1}^n$. The final stage displays three 3D rendered frames of the resulting 4D reconstruction, showing two human meshes interacting with the ball trajectory at a table tennis table.}
    \vspace{-0.2cm}
    \caption{
        A visual outline of our proposed Lift-First Pipeline.
        Instead of depending on challenging and noisy temporal segmentations of the video sequence before attempting the 3D uplifting of the trajectory, we invert this logic by lifting the entire sequence to 3D first, which makes subsequent temporal segmentation and refinement in the 3D domain a simple and robust task.
    }
    \vspace{-0.0cm}
    \label{fig:pipeline}
\end{figure*}

\section{Methodology: The Lift-First Pipeline}
\label{sec:methods}

\subsection{Terminology}
We first define terminology used throughout the paper:
\begin{itemize}[leftmargin=*,itemsep=0pt,parsep=0pt,topsep=0pt,partopsep=0pt]
    \item A \textsc{Segment} starts when one player hits the ball and ends when another player hits the ball.
    \item A \textsc{Point} starts with a serve and ends when the ball is first out of play. Segments partition the point into disjoint time intervals. 
    \item A \textsc{Game} is a sequence of points. It is complete when one team achieves the winning score. 
    \item \textsc{(2D/3D)-based Time Segmentation} partitions a point into segments using 2D or 3D information. 
    \item \textsc{4D Reconstruction} recovers camera parameters, 3D ball positions, ball spin, and 3D human meshes \textit{over time}.
\end{itemize}

\subsection{Pipeline Overview}
\label{subsec:overview}
Conventional table tennis reconstruction pipelines first perform 2D-based time segmentation of a point and then reconstruct each segment independently~\cite{etaat2025latte,gossard2025tt3d}.
However, this time segmentation scheme is highly sensitive to occlusions and missing detections, limiting scalability.\\[0.66ex]
\noindent We instead adopt a \textbf{Lift-First Pipeline} (Fig.~\ref{fig:pipeline}).
Rather than segmenting first, we lift the \emph{entire unsegmented point} to 3D and perform time segmentation and annotation directly in the 3D domain, where the ball trajectory signal is unambiguous. 
The pipeline consists of four stages:
\begin{enumerate}[leftmargin=*,itemsep=0pt,parsep=0pt,topsep=0pt,partopsep=0pt]
\item \textbf{Data Acquisition and Preprocessing} (\Cref{subsec:preprocessing}): clipping the points of a game from broadcast footage, calibrating cameras, extracting 2D ball tracks, and estimating 3D human meshes \cite{loper2015smpl}.
\item \textbf{Full 3D Lifting} (\Cref{subsec:uplifting}): a transformer-based network predicts dense 3D ball trajectories and per-frame spin for the full point from the 2D track.
\item \textbf{3D-Domain Annotation} (\Cref{subsec:3dannotation}): segments, hit points, bounces, and racket-contact parameters are computed from the reconstructed 3D trajectory.
\item \textbf{Filtering and Curation} (\Cref{subsec:filtering}): 2D and 3D consistency checks ensure that all retained trajectories are visually and physically plausible.
\end{enumerate}

\subsection{Stage 1: Data Acquisition and Preprocessing}
\label{subsec:preprocessing}
Our pipeline begins with raw, multi-hour uncut table tennis videos of full games.
We apply a two-stage clipping process.
The first stage splits the games into individual points by detecting scoreboard changes using YOLO~\cite{10533619} and PaddleOCR~\cite{cui2025paddleocr30technicalreport}.
The second stage identifies the approximate start and end of the point using 2D ball track oscillations using a ball tracker. \\[0.66ex]
\noindent We then process all clips to detect and remove duplicated frames, a common artifact in broadcast video that corrupts trajectory estimation.
Our method uses \ac{SSIM} \cite{wang2004ssim} to identify these frames. \\[0.66ex]
\noindent For each resulting valid clip, we extract the following information:
\begin{itemize}[leftmargin=*,itemsep=0pt,parsep=0pt,topsep=0pt,partopsep=0pt]
\item \textbf{Camera calibration:} We follow TT3D~\cite{gossard2025tt3d}, solving a Perspective-$n$-Point problem from table corners with unknown focal length, and improving robustness through enhanced table segmentation and temporal filtering.
\item \textbf{2D ball detections:} TrackNetV3~\cite{chen2024tracknetv3} is applied without its inpainting module.
\item \textbf{3D human meshes:} We use 4DHumans~\cite{goel2023humans4d} and align the meshes to the world frame.
\end{itemize}
\noindent Additional implementation and parameter details are provided in the supplementary material.

\subsection{Stage 2: Full-Sequence Lifting Network}
\label{subsec:uplifting}
The central component of our pipeline is a transformer-based Full-Sequence Lifting Network. 
For each point consisting of $N$ frames, it processes the sequence of 2D ball detections $\{ \vec{r}_{2D}(t_n) \}_{n=0}^{N-1}$, their corresponding timestamps $\{ t_n \}_{n=0}^{N-1}$, and a set of 2D table keypoints $\{ \vec{k}_i \}_{i=1}^{13}$ that are derived from the camera calibration.
The network infers the 3D trajectory $\{ \vec{r}_{\,3D}(t_n) \}_{n=0}^{N-1}$ and 3D spin vectors $\{ \vec{\omega}(t_n) \}_{n=0}^{N-1}$ for each frame. \\[0.66ex]
\noindent Our network is built upon the baseline lifting model from \citet{kienzle2026uplifting}, which demonstrated strong generalization to real videos despite being trained solely on synthetic data.
We retain its key innovations, such as using Rotary Positional Embeddings (RoPE) \cite{su2024rope} based on exact timestamps to handle varying frame rates and missing 2D ball detections~\cite{kienzle2026uplifting}. \\[0.66ex]
\noindent However, the baseline model is designed for a ``Traditional Pipeline'': it processes isolated, pre-segmented shots, predicts only a single initial spin vector per segment, and handles missing 2D ball detections by simply discarding them. This is insufficient for our ``Lift-First Pipeline,'' which must process unsegmented points of arbitrary length, generate dense spin estimates, and actively reconstruct occluded detections to enable precise 3D-based temporal segmentation.
Therefore, we introduce three key contributions to solve this: a massive-scale synthetic training dataset of full points, architectural extensions for modeling dense spin, and an interpolation token that enables predictions for missed detections. \\[0.66ex]
\noindent\textbf{Synthetic Dataset.}
To learn the dynamics of continuous play, we require training data that reflects the complexity of full, unsegmented points, not just isolated segments as is done in \cite{kienzle2025towards, kienzle2026uplifting}.
We therefore generate a new massive-scale synthetic dataset of 3 million points using the MuJoCo~\cite{todorov2012mujoco} physics simulation environment. \\[0.33ex]
We develop an iterative ``stitching'' algorithm. \noindent We first simulate a pre-serve ball toss; at its apex, we query a large data pool of initial conditions of serves from \cite{dambrosio2025achieving, kienzle2026uplifting} for the closest match in position. 
This serve is rolled out and from its terminal state we again query a large data pool of standard segments.
By iteratively matching and stitching these sampled trajectory segments, we produce continuous, physically plausible sequences that enable training our network on full unsegmented points. \\[0.66ex]
\noindent\textbf{Dense Spin Predictions.}
We adapt the baseline network architecture to exploit this new continuous data. The architecture is illustrated in the supplementary material. \\[0.33ex]
\noindent The baseline model uses a learnable ``spin token'' to aggregate information and predicts a single \textit{initial} spin vector $\vec{\omega}(t_0)$ for the input segment.
This is no longer feasible for processing full points, as the segments in the point are not known.
We therefore remove this specialized token entirely.
Instead, we modify the network to predict spin in a dense, per-frame manner by applying a small MLP head to \textit{every} output token of the transformer. \\[0.33ex]
\noindent To encourage the network to learn robust trajectory and spin dynamics for points of arbitrary length, we sample a subsequence (with a random length between $20$ and $250$ frames) from each full point in our synthetic dataset. \\[0.66ex]
\noindent\textbf{Interpolation Token.}
Ball detections are frequently missing due to occlusions in oblique views of gameplay.
While the baseline architecture~\cite{kienzle2026uplifting} simply ignores these detections, we treat the recovery of missing frames as a Masked Token Modeling (MTM) task. \\[0.33ex]
To prevent the loss of spatial camera context when a ball is missing, we introduce a \textit{Disentangled Context Embedding (DCE)}. 
The 2D ball position and the table keypoints are projected into a higher dimension vector via separate linear layers. 
For frames with missing ball detections, we replace only the ball vector with a learnable interpolation token, leaving the projected table keypoints intact to preserve the camera information.
Finally, we concatenate the ball vector and table keypoint vector and apply a linear layer to obtain the final embedding for each frame.
This is illustrated in the supplementary material. \\[0.33ex] 
To prevent polluting the valid information of successful ball detections with the interpolation tokens, we integrate Deferred Upsampling Token Attention (DUTA) \cite{einfalt2023uplift}. 
DUTA applies masking in the initial transformer layer to prevent context dilution.
Each token is only allowed to attend to tokens representing valid detections, ensuring that the tokens representing invalid detections can gather the necessary context without diluting the information of the valid tokens.
During training, we randomly mask valid 2D detections using empirically-derived occlusion probabilities (discussed in supplementary materials) and compute the dense 3D reconstruction loss over the entire trajectory, including the masked frames. 
This forces the network to internalize the underlying physical constraints of ball motion to accurately in-paint missing segments.

\subsection{Stage 3: 3D-Domain Annotation}
\label{subsec:3dannotation}
With an unambiguous 3D ball trajectory now available, we can perform time segmentation and annotation directly in the 3D domain. \\[0.66ex]
\noindent\textbf{Robust 3D-based Time Segmentation.}
Our Lift-First Pipeline transforms time segmentation from a complex, 2D image-level tracking problem into an unambiguous, 1D signal analysis task in world coordinate space.
We identify hit events as the peaks and troughs in the ball's 3D $x$-coordinates, using simple time and distance heuristics to filter local noise.
Similarly, we label table bounces as the local minima in the z-coordinates.
This provides the time segmentations that prior methods failed to reliably achieve. \\[0.66ex]
\noindent\textbf{Racket Stroke Estimation.}
Estimating a player’s 3D body pose provides useful cues for anticipating ball motion, but current human-pose estimators do not reliably capture hand orientation or wrist articulation.
This limitation is critical: the racket orientation at impact strongly determines the outgoing ball trajectory and spin.
Prior attempts to estimate racket pose directly from video~\cite{gao2019markerless, wang2013pose} remain insufficiently accurate or robust outside controlled laboratory settings.
Instead, we infer the racket state indirectly from the 3D ball trajectory.
When the ball’s flight time is short and its spin remains within a moderate range, the two-point boundary-value problem defined by the hit position and the subsequent table bounce admits a unique physically plausible ball trajectory~\cite{liu2012racket}.
This provides the required ball velocity and spin immediately after impact.
Implementation and validation details are provided in the supplementary material.
We use this procedure to augment our dataset with physically consistent racket-stroke parameters.

\subsection{Stage 4: Filtering and Curation}
\label{subsec:filtering}
\begin{figure}[t]
    \vspace{-0.0cm}
    \centering
    \begin{subfigure}{0.87\linewidth}
        \centering
        \includegraphics[width=\linewidth]{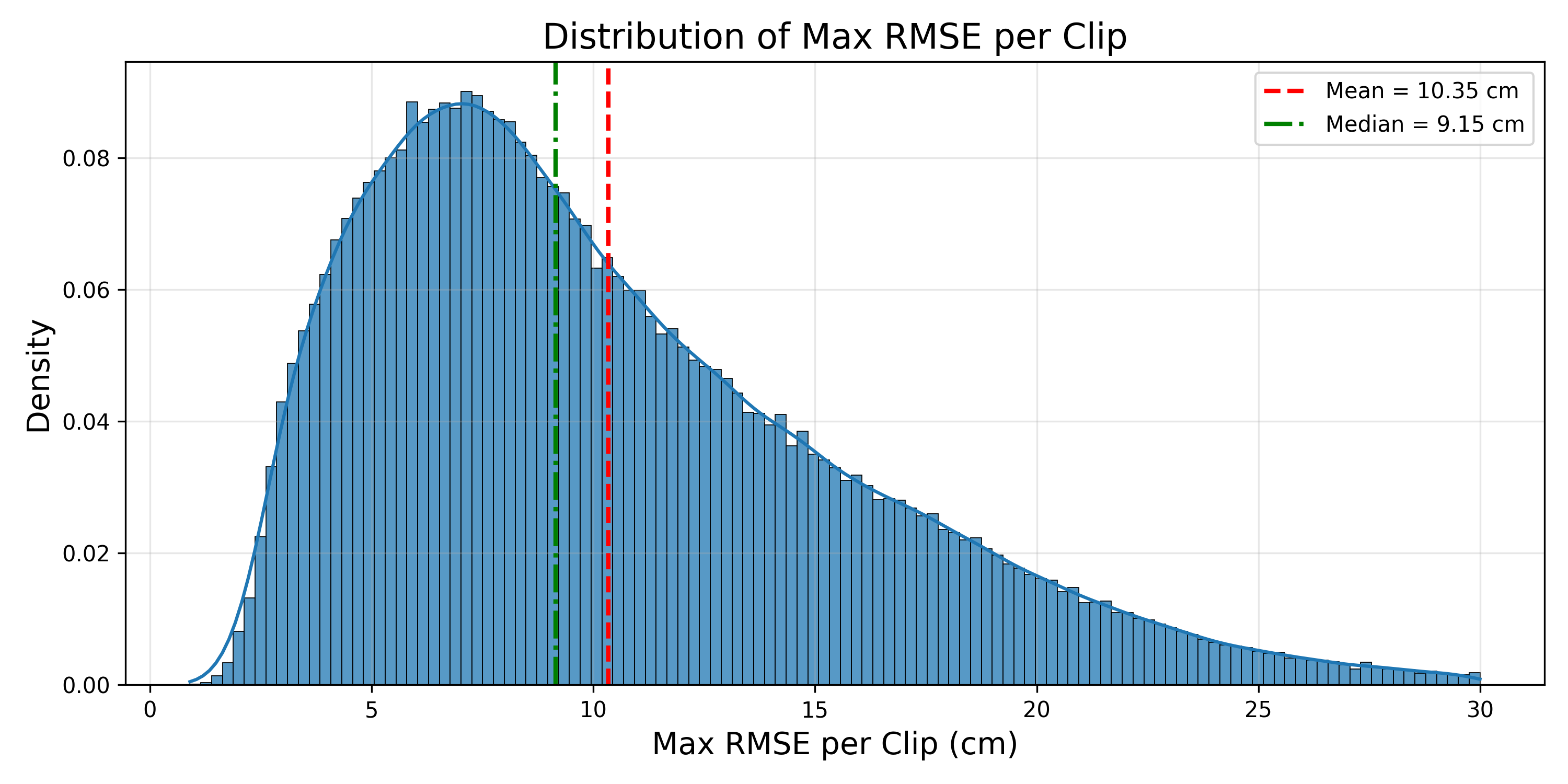}
        \label{fig:top}
    \end{subfigure}
    \vspace{-0.99cm}  
    \begin{subfigure}{0.87\linewidth}
        \centering
        \includegraphics[width=\linewidth]{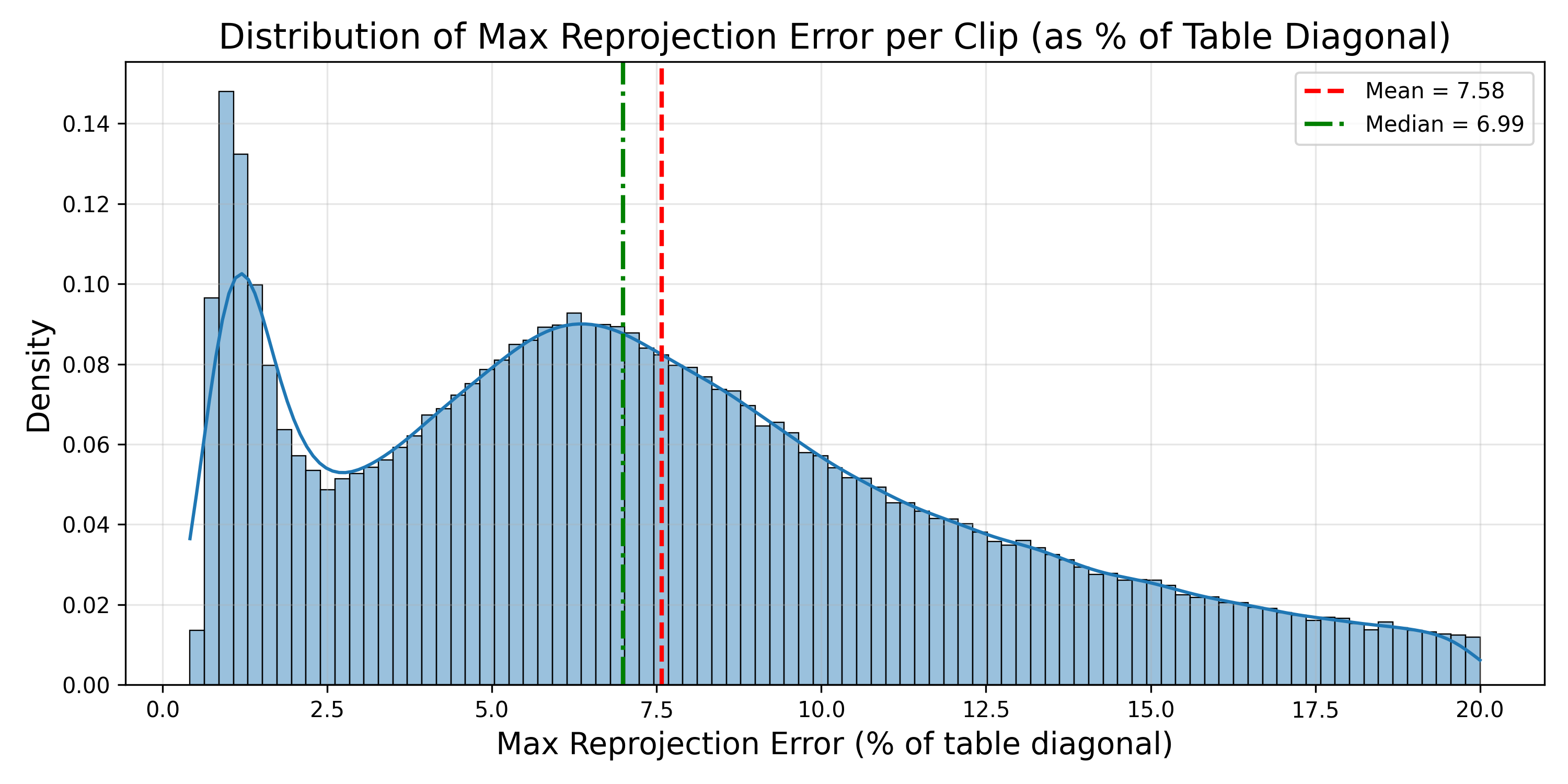}
        \label{fig:bottom}
    \end{subfigure}
    \vspace{0.3cm}
    \Description{Two vertically stacked histograms, both showing strongly right-skewed distributions with peaks near zero and long tails extending to the right. The top histogram plots the maximum Physics-Based ODE Fit error (RMSE) per clip in centimeters, featuring a vertical green dash-dot line marking the median at 8.30 cm and a vertical red dashed line marking the mean at 10.77 cm. The bottom histogram plots the maximum normalized 2D Reprojection Error per clip as a percentage of the table diagonal, featuring a vertical green dash-dot line marking the median at 8.12 and a vertical red dashed line marking the mean at 10.28.}
    \caption{Distributions of the two quality metrics used in our \textbf{Filtering} pipeline.
    \textbf{(Top)} Histogram of the maximum \textbf{Physics-Based ODE Fit} error (RMSE) per point.
    \textbf{(Bottom)} Histogram of the maximum \textbf{normalized 2D Reprojection Error} per point.
    The peaks in both distributions near zero demonstrate that our pipeline's outputs are overwhelmingly physically plausible and visually accurate.
    The tails of the distributions provide a clear margin for selecting reliable filtering thresholds..
    }
    \label{fig:two_vertical_subfigs}
    \vspace{-0.4cm}
\end{figure}

The final stage enforces visual and physical consistency across all reconstructed rallies in order to keep only high quality reconstruction for our TT4D dataset. We apply one 2D-based filter and two 3D-based filters. \\[0.66ex]
\noindent\textbf{2D Reprojection Check.}
We compare the reprojected 3D trajectory estimate to the original detections and normalize errors by the pixel length of the table diagonal. 
If the maximum normalized error exceeds a strict threshold (20\% of the table diagonal length), the point is rejected.
\Cref{fig:two_vertical_subfigs}(b) shows the distribution of this error. \\[0.66ex]
\noindent\textbf{Event Plausibility:} We enforce game logic by analyzing the 3D path, ensuring it contains a valid sequence of events: clearly identifiable hit points and a single table bounce per segment (or two for serves). \\[0.66ex]
\noindent\textbf{Physical Consistency:} We assess physical consistency by fitting the \ac{ODE} constrained ball trajectory~\cite{gossard2025tt3d} to our network's 3D output.
This step effectively projects our prediction onto the manifold of physically possible trajectories.
If the maximum Euclidean distance between the prediction and the ODE fit exceeds a 30 cm threshold, the point is discarded as physically implausible.
The error distribution is shown in \Cref{fig:two_vertical_subfigs}(a).

\section{The TT4D Dataset}
\label{sec:dataset}

\subsection{General Information}
The TT4D dataset is sourced from 45,946 broadcast table tennis games (best of five, first to three) from 2021-2024.
We support general stationary camera poses, singles and doubles gameplay, and video speeds of at least 25 FPS, ultimately resulting in 211,534 reconstructed points and 146 hours of gameplay. 
For comparison, the LATTE-MV dataset is based on a set of 1017 games, requires a particular camera pose, only handles singles gameplay, and ultimately results in 23,782 reconstructed points and 26 hours of gameplay.
Our scalability stems directly from the Lift-First Pipeline.
By lifting the full unsegmented point directly to 3D, hit points and bounces can be identified robustly in world coordinates, even under heavy occlusion. 
Moreover, because the 3D trajectory is reconstructed independently of human-pose tracking, the method naturally generalizes to doubles matches. 
A further advantage from a game-theoretic standpoint is that the final winning shot of each point — unrecoverable in 2D-based time segmentation pipelines — is reliably reconstructed.

\subsection{Filtering Effectiveness}
We display the amount of filtered data in \Cref{tab:rtt_pipeline_stages}. The first stage in the pipeline involves a two-part clipping process, which is outlined in \Cref{subsec:preprocessing}. Our simple, conservative heuristic succeeds on $56.8\%$ of the available points. Performance can be improved through refined scoreboard detection and gameplay-identification heuristics. A small fraction of points ($\sim 5\%$) are lost due to either (1) a detected change in the camera pose or (2) calibration algorithm failure. \\[0.66ex]
\noindent Our 3D-domain filters play a significant role in quality control. Discrepancies in the 3D ball trajectory prediction and 2D trajectory evidence $(4.3\%)$, measured as reprojection error, and physically implausible ball trajectories $(1.4 \%)$ and make for effective filters. A substantial number of points are rejected due to human mesh recovery (e.g. not enough potential athletes, missing detections or improperly positioned athletes). We also require a minimum of two segments to be reconstructed for each point, and we require an average ball visibility of at least 50\%. 
Additional data curation details are presented in the supplementary materials.

\begin{table}[t]
\centering
\setlength{\tabcolsep}{2pt}
\caption{Pipeline filtering statistics per stage. Broadcast videos are clipped into points at scoreboard changes. We further trim clip downs to the actual gameplay, eliminating rests before and after the point. Lastly, we calibrate the camera and apply consistency filters to the reconstructed ball trajectory.}
\label{tab:rtt_pipeline_stages}

\vspace{-0.2cm}
\footnotesize
\setlength{\tabcolsep}{4pt}
\renewcommand{\arraystretch}{1.15}
\begin{tabular}{@{}l r r r@{}}
\toprule
Stage & Count & \% vs prev & \% from start \\
\midrule

\textbf{Scoreboard clips} & & & \\
\hspace{2em}success & 714,664 &  & \textbf{100.0}\% \\

\textbf{Gameplay clips} & & & \\
\hspace{2em}success & 405,769 & 56.8\% & \textbf{56.8}\% \\
\hspace{2em}failed & 308,895 & 43.2\% & 43.2\% \\

\textbf{Calibrations} & & & \\
\hspace{2em}success & 371,733 & 91.6\% & \textbf{52.0}\% \\
\hspace{2em}moving camera & 33,137 & 8.2\% & 4.6\% \\
\hspace{2em}failed & 899 & 0.2\% & 0.1\% \\

\textbf{Reconstructed} & & & \\
\hspace{2em}success & 211,534 & 56.9\% & \textbf{29.6}\% \\
\hspace{2em}invalid human pos & 67,442 & 18.1\% & 9.4\% \\
\hspace{2em}not enough segments & 47,823 & 12.9\% & 6.7\% \\
\hspace{2em}high reproj errors & 30,937 & 8.3\% & 4.3\% \\
\hspace{2em}high ODE fit & 9,736 & 2.6\% & 1.4\% \\
\hspace{2em}low ball vis & 4,261 & 1.1\% & 0.6\% \\

\bottomrule
\end{tabular}
\end{table}

\subsection{Statistics and Basic Analysis}
\begin{figure}[t]
    \centering

    \begin{subfigure}{0.48\linewidth}
        \centering
        \includegraphics[width=\linewidth]{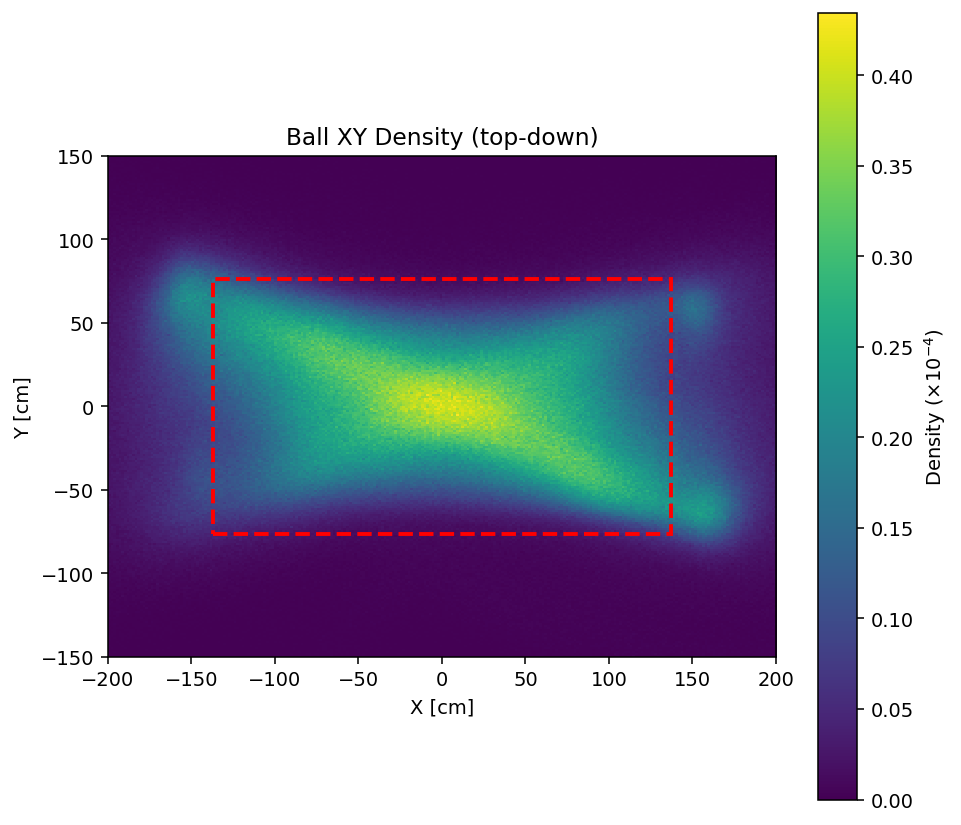}
        \label{fig:ball_xy_density}
    \end{subfigure}
    \hfill
    \begin{subfigure}{0.48\linewidth}
        \centering
        \includegraphics[width=\linewidth]{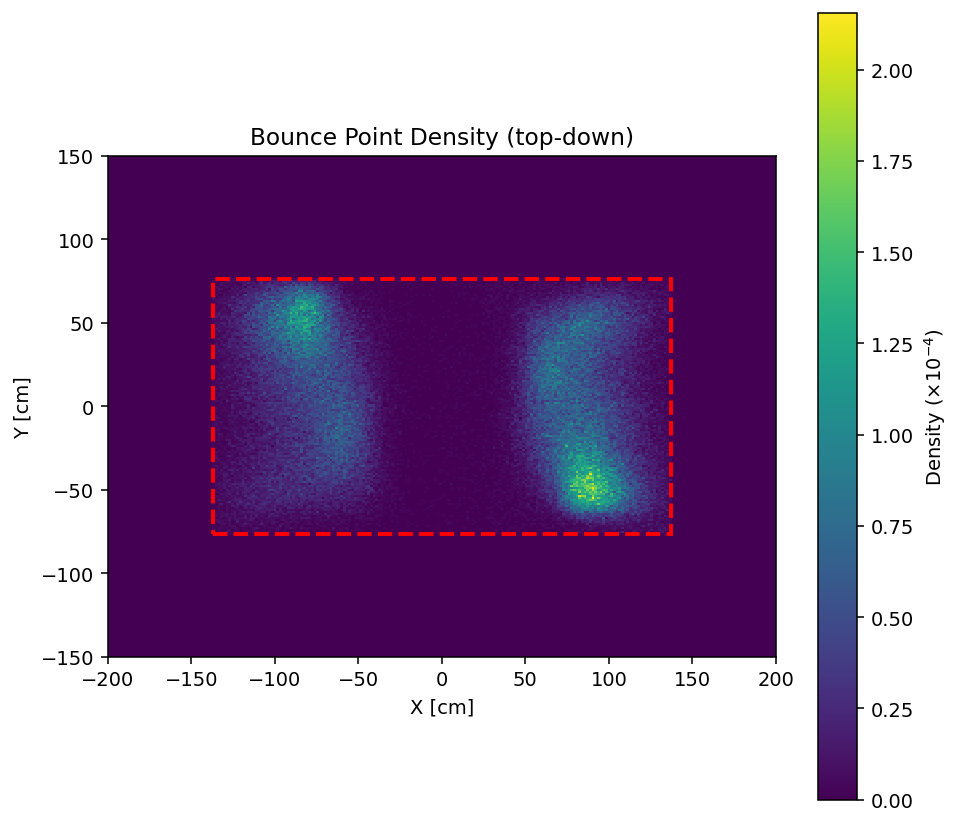}
        \label{fig:bounce_xy_density}
    \end{subfigure}

    \vspace{-0.2cm}

    \begin{subfigure}{0.70\linewidth}
        \centering
        \includegraphics[width=\linewidth]{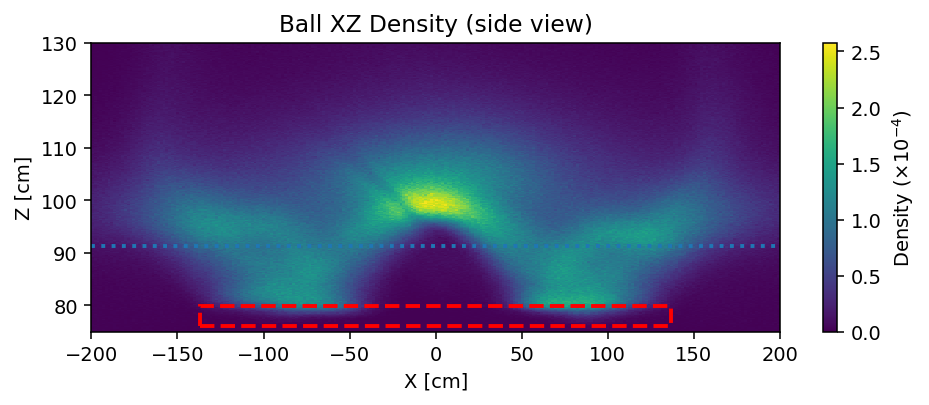}
        \label{fig:ball_xz_density}
    \end{subfigure}
    \vspace{-0.5cm}
    \Description{Three heatmaps displaying ball position and bounce densities for the TT4D dataset. A dashed red rectangle outlines the table boundaries in each plot. The top-left plot shows a top-down view of ball XY density, revealing an X-shaped high-density pattern that is concentrated in the center over the net and spreads towards the corners. The top-right plot shows top-down bounce point density, featuring concentrated clusters of bounces on the left and right halves of the table. The bottom plot presents a side view of ball XZ density, showing the ball's typical flight path arching closely over a dotted blue line that represents the net height, and dipping down to contact the table surface on both sides.}
    \caption{
    Ball position densities for the TT4D dataset. The table region is marked by the dashed red line, and the net's height is marked by the dotted blue line. 
    }
    \vspace{-0.2cm}
    \label{fig:ball_density_all}
\end{figure}
\begin{figure}
    \centering
    \includegraphics[width=0.9\linewidth]{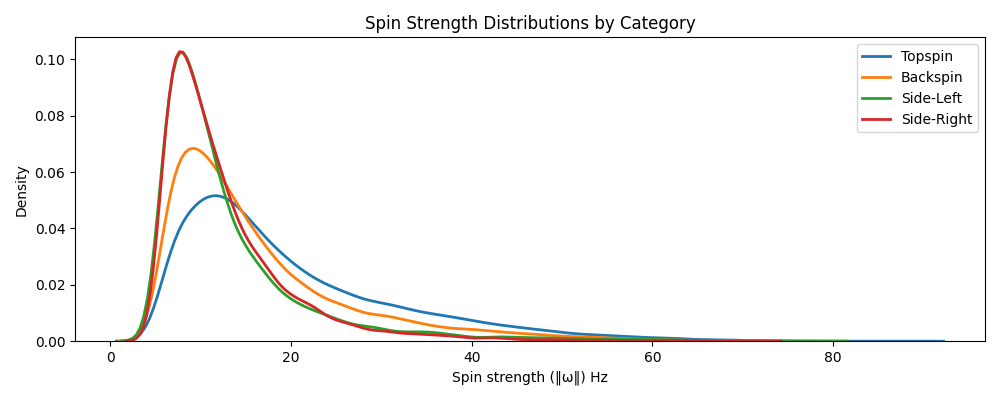}
    \vspace{-0.2cm}
    \Description{A density plot visualizing ball spin strength distributions categorized into Topspin (blue), Backspin (orange), Side-Left (green), and Side-Right (red). The horizontal axis measures Spin strength in Hz, and the vertical axis measures Density. The Side-Left and Side-Right curves are tightly aligned, displaying a single, sharp peak between 5 and 10 Hz before tapering off quickly. In contrast, the Topspin and Backspin curves are significantly flatter and exhibit heavy-tailed distributions; they feature an initial peak below 10 Hz, followed by a secondary, broader plateau that extends up to approximately 30 Hz.}
    \caption{Visualization of ball spin strength per spin category. 
    }
    \label{fig:spin_strength}
    \vspace{-0.5cm}
\end{figure}
The ball position densities in \Cref{fig:ball_density_all} demonstrate interesting aspects of competitive table tennis gameplay. 
The ball typically crosses between 5-15 cm above the net, as players keep the ball low to achieve high speed shots. 
Players often choose cross-court shots, which are safer to land at higher speeds, over down-the-line hits. 
The bounce point density also reveals insightful patterns. When players send the ball from the left to the right in a cross-court hit, they tend to hit a clear high-density region. However, when players hit the ball from the right to the left, the bounce-point density is far less concentrated. \\[0.66ex]
\noindent We further analyze gameplay patterns by examining the distribution of spin strengths in \Cref{fig:spin_strength}. 
For each segment, we extract the first predicted spin vector and assign it to one of five categories: ``topspin", ``backspin", ``sidespin-left", ``sidespin-right", and ``no spin". A spin vector $\omega$ is classified as ``no spin" when $||\omega||_\infty \leq 5 \text{ Hz}$; all remaining vectors are categorized following the conventions of ~\cite{kienzle2025towards, kienzle2026uplifting}. The resulting distributions reveal differences between spin types. While all spin types exhibit a unimodal distribution, the ``topspin" and ``backspin"  display slightly heavier-tailed distributions. These patterns highlight the broader variability and more extreme spin magnitudes characteristic of top- and backspin strokes in competitive rallies.

\section{Evaluation of Lifting Network}
\label{sec:experiments}
\begin{table}[t]
\centering
\caption{Evaluation of the Lifting Network on the synthetic and TTST~\cite{kienzle2025towards} datasets under augmentations. 
\ding{51}/\ding{53} indicates the presence/absence of augmentation.
The model shows strong resilience to lower frame rates and missing data.
}
\label{tab:augmentation_results}
\resizebox{0.9\columnwidth}{!}{%
\begin{tabular}{@{}cc|cc|cc@{}}
\toprule
\multicolumn{2}{c|}{\textbf{Augmentation}} & \multicolumn{2}{c|}{\textbf{Synthetic Dataset}} & \multicolumn{2}{c}{\textbf{TTST Dataset (Real)}} \\
\makecell{Half\\FPS} & \makecell{Missing\\Detections} & \makecell{$\Delta \vec{r}_{\text{3D}}$ (cm) $\downarrow$\\ \footnotesize{(3D Pos. Error)}} & \makecell{$\Delta \vec{\omega}$ (Hz) $\downarrow$\\ \footnotesize{(Spin Error)}} & \makecell{$\Delta \vec{r}_{\text{2D}}$ (px) $\downarrow$\\ \footnotesize{(2D Proj. Error)}} & \makecell{Macro F1 $\uparrow$\\ \footnotesize{(Spin Classif.)}} \\ \midrule
\ding{53} & \ding{53} & $2.60 \pm 1.15$ & $6.36 \pm 2.76$ & $2.52 \pm 0.86$ & $1.000$ \\
\ding{51} & \ding{53} & $3.28 \pm 1.45$ & $7.55 \pm 3.41$ & $2.34 \pm 0.76$ & $1.000$ \\
\ding{53} & \ding{51} & $2.69 \pm 1.17$ & $6.50 \pm 2.81$ & $2.78 \pm 1.00$ & $1.000$ \\
\ding{51} & \ding{51} & $3.55 \pm 1.63$ & $7.86 \pm 3.62$ & $3.49 \pm 2.44$ & $0.941$ \\ \bottomrule
\end{tabular}%
}
\end{table}
\begin{table}[t]
\centering
\caption{3D reconstruction error $\Delta \vec{r}_\text{3D}$ (cm) on TT4DBench. Processing the full point (our approach) consistently yields lower errors than processing individual segments (traditional approach), demonstrating the benefit of full context.}
\vspace{-0.2cm}
\renewcommand{\arraystretch}{1.15}
\resizebox{0.75\columnwidth}{!}{%
\begin{tabular}{c l cc cc}
\toprule
 &  & \multicolumn{2}{c}{\textbf{Full Point}} & \multicolumn{2}{c}{\textbf{Individual Segments}} \\
\cmidrule(lr){3-4} \cmidrule(lr){5-6}
\textbf{Noise} & \textbf{View} & \textbf{Mean} & \textbf{Std} & \textbf{Mean} & \textbf{Std} \\
\midrule
\multirow{4}{*}{\textbf{True}}
 & Back     & 21.93 & 11.34 & 32.42 & 40.98 \\
 & Side     & 13.11 &  8.28 & 15.04 & 16.04 \\
 & Oblique  & 18.71 &  6.86 & 21.43 & 16.56 \\
\rowcolor{gray!10}
 & \textbf{All Views}
           & \textbf{17.92} & \textbf{9.73}
           & \textbf{22.96} & \textbf{28.08} \\
\midrule
\multirow{4}{*}{\textbf{False}}
 & Back     & 20.68 & 10.88 & 30.14 & 39.10 \\
 & Side     & 12.81 &  8.07 & 14.65 & 16.41 \\
 & Oblique  & 18.28 &  6.96 & 20.84 & 16.56 \\
\rowcolor{gray!10}
 & \textbf{All Views}
           & \textbf{17.26} & \textbf{9.39}
           & \textbf{21.88} & \textbf{27.04} \\
\bottomrule
\end{tabular}
}
\label{tab:tt4d_benchmark}
\end{table}
\begin{table}[t]
\centering
\caption{Side-view 3D reconstruction error $\Delta \vec{r}_{\text{3D}}$ (cm) for individual segments from TT4DBench. Comparison to baseline methods with and without noisy 2D detections.
}
\vspace{-0.2cm}
\renewcommand{\arraystretch}{1.20}
\resizebox{0.6\columnwidth}{!}{%
\begin{tabular}{c l cc}
\toprule
 & & \multicolumn{2}{c}{\textbf{Individual Segments}} \\
\cmidrule(lr){3-4}
\textbf{Noise} & \textbf{Method} & \textbf{Mean} & \textbf{Std} \\
\midrule
\multirow{3}{*}{\textbf{True}}
 & TT3D      & 29.86 & 26.64 \\
 & LATTE--MV & 15.27 & 6.36  \\
 \rowcolor{gray!10}
 & \textbf{TT4D (Ours)} & \textbf{15.04} & \textbf{16.04} \\
\midrule
\multirow{3}{*}{\textbf{False}}
 & TT3D      & 29.91 & 27.79 \\
 & LATTE--MV & 15.78 & 6.37  \\
 \rowcolor{gray!10}
 & \textbf{TT4D (Ours)} & \textbf{14.65} & \textbf{16.41} \\
\bottomrule
\end{tabular}
}
\label{tab:side_view_comparison_v2}
\vspace{-0.3cm}
\end{table}
We compare our Full-Sequence Lifting Network to traditional regression pipelines, evaluate the model's robustness, show that the full-point setting is beneficial, and finally evaluate the physical plausibility of the predictions. \\[0.66ex]
\noindent\textbf{Metrics.}
For datasets with 3D ground truth (synthetic and TT3D), we report the \textbf{3D Trajectory Error} ($\Delta\vec{r}_{\text{3D}}$), which is the mean Euclidean distance between the predicted and ground truth 3D positions, and \textbf{F1} scores for table bounce and racket hit detection using a tolerance of $\pm 2$ frames ($\pm 0.08\,\mathrm{s}$).
On synthetic data, we additionally measure the \textbf{3D Spin Error} ($\Delta\vec{\omega}$), the mean Euclidean distance between predicted and ground truth spin vectors.
For 2D benchmarks (TTST \& TTHQ) that lack 3D ground truth, we report the \textbf{2D Reprojection Error} ($\Delta\vec{r}_{\text{2D}}$) in pixels and the \textbf{Macro F1} score for topspin/backspin classification as described in \cite{kienzle2026uplifting}.
Detailed mathematical definitions for all metrics are in the supplementary material. \\[0.66ex]
\noindent\textbf{Robustness Analysis.} \,
Our network must be robust to real-world video artifacts, such as low frame rates or dropped ball detections from occlusions, which are not present in ``perfect" evaluation datasets.
We therefore simulate ``in-the-wild'' conditions on the synthetic test set and the real-world TTST test set (following the evaluation protocol of \cite{kienzle2026uplifting}). 
We implement the augmentations: Half FPS (dropping every second frame) and Missing Detections (randomly removing 10\% of detections to simulate occlusions).
Results are shown in \Cref{tab:augmentation_results}. \\
The model yields good performance on both datasets. 
Applying the augmentations individually does not lead to significant performance degradation.
Even though combining both augmentations leads to a noticeable performance drop, the results are still strong.
In total, the results verify our model's robustness to the ``in-the-wild" conditions. \\[0.66ex]
\noindent\textbf{Interpolation Performance.}
\begin{figure}
    \centering
    \includegraphics[width=0.65\linewidth]{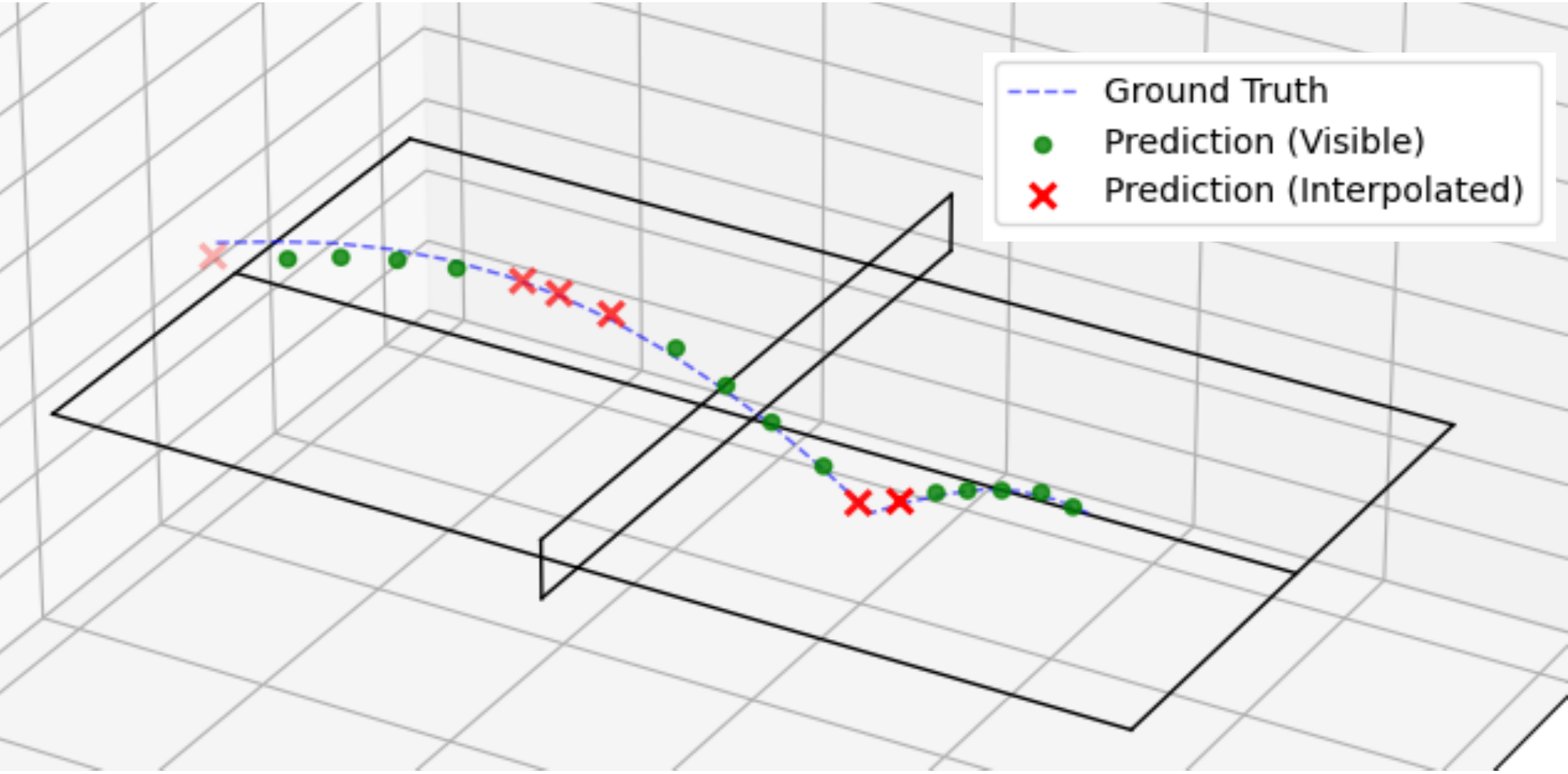}
    \vspace{-0.2cm}
    \Description{A 3D plot illustrating a reconstructed ball trajectory over a table tennis table. A legend indicates three distinct data representations: 'Ground Truth', 'Prediction (Visible)', and 'Prediction (Interpolated)'. The plot demonstrates how the interpolated predictions successfully bridge the gaps in the trajectory where visible 2D detections are missing, closely following the expected path.}
    \caption{Reconstructed 3D trajectory. Even when the 2D detection is missing, the model is able to compute reasonable predictions due to its interpolation capabilities.}
    \vspace{-0.1cm}
    \label{fig:interpolation}
\end{figure}
Notably, \Cref{tab:augmentation_results} shows that the model yields strong performance under the Missing Detections augmentation, which randomly sets 10\% of the ball detections to missed detections. 
This not only shows the models robustness to real-world input, but emphasizes the model's interpolation capabilities as the interpolated predictions only lead to minor performance degradations.
This is further illustrated by \Cref{fig:interpolation} which shows a reconstructed trajectory with successful interpolations. \\[0.66ex]
\noindent\textbf{Full-Point vs. Single-Segment.} \,
We next validate our core contribution that processing the full, unsegmented sequence is superior to the ``Traditional Pipeline'' approach of processing isolated segments.
To create a challenging benchmark, we adapt the TT3D dataset~\cite{gossard2025tt3d}, which provides ground-truth 3D trajectories.
While the original dataset is highly filtered and contains only single segments, we create a less-filtered stitched dataset (e.g. including throws) to obtain ``in-the-wild'' conditions. We denote this realistic multi-segment benchmark \textit{TT4DBench}.
We use this benchmark to compare our full-point approach against a single-segment approach. \\
In \Cref{tab:tt4d_benchmark} we compare our model's capability to process a full point at once vs processing each segment individually. 
Processing the Full Point consistently outperforms the Individual Segments baseline across all camera views, reducing the mean $\Delta\vec{r}_{\text{3D}}$ from $21.88$ cm to $17.26$ cm.
This shows that the network successfully uses the extra context from the full point to improve reconstruction accuracy. \\[0.66ex]
\noindent\textbf{Comparison with Traditional Pipelines.} We benchmark our Full-Sequence Lifting Network in \Cref{tab:side_view_comparison_v2} against the methods of LATTE-MV~\cite{etaat2025latte} and TT3D~\cite{gossard2025tt3d}, which estimate the trajectory at inference time instead of using a learning based approach. Since these methods are not able to reconstruct a full point, we restrict our evaluation to single segments. 
Since the LATTE-MV algorithm is limited to processing segments from a side-view perspective, we restrict its evaluation to side-view, single-segment examples.
Furthermore, LATTE-MV requires the precise 3D starting position and bounce point of the ball, information not available during in-the-wild inference.
To enable a comparison, we provide LATTE-MV with these values from the 3D ground truth, granting it a significant advantage with privileged 3D information. 
Despite this advantage, our method outperforms LATTE-MV and TT3D, achieving a lower mean reconstruction error. \\[0.66ex]
\noindent\textbf{Physical Consistency Check.} \,
Finally, we verify that our network's predictions are not just accurate but also physically consistent.
After segmenting the point sequence in 3D, we analyze each individual segment.
This per-segment validation is only possible because our Lift-First Pipeline provides robust time segmentations.
We fit a physics-based \ac{ODE} model~\cite{gossard2025tt3d} directly to our network's 3D output trajectories, acting as a proof of physical consistency.
Note this differs from the original TT3D method, since we directly fit the \ac{ODE} to our 3D predictions.
\Cref{fig:ode_fit_3d} provides a clear qualitative validation: the fitted physical path (solid line) closely follows the network's predictions (dots), demonstrating the physical plausibility of our predictions. \\
\noindent This figure also visualizes the exact failure case of the ``Traditional Pipeline": around the hit points, the 2D detections are missing (gaps in the dots) due to player occlusion.
Our network robustly predicts the 3D trajectory for the full sequence despite these gaps, yielding a continuous, physically plausible path where 2D-first methods would fail.
As detailed in \Cref{subsec:filtering}, this ODE fit is also used as a quantitative quality metric during dataset curation.\\[0.66ex]
\begin{figure}[t]
    \centering
    \includegraphics[width=0.88\linewidth]{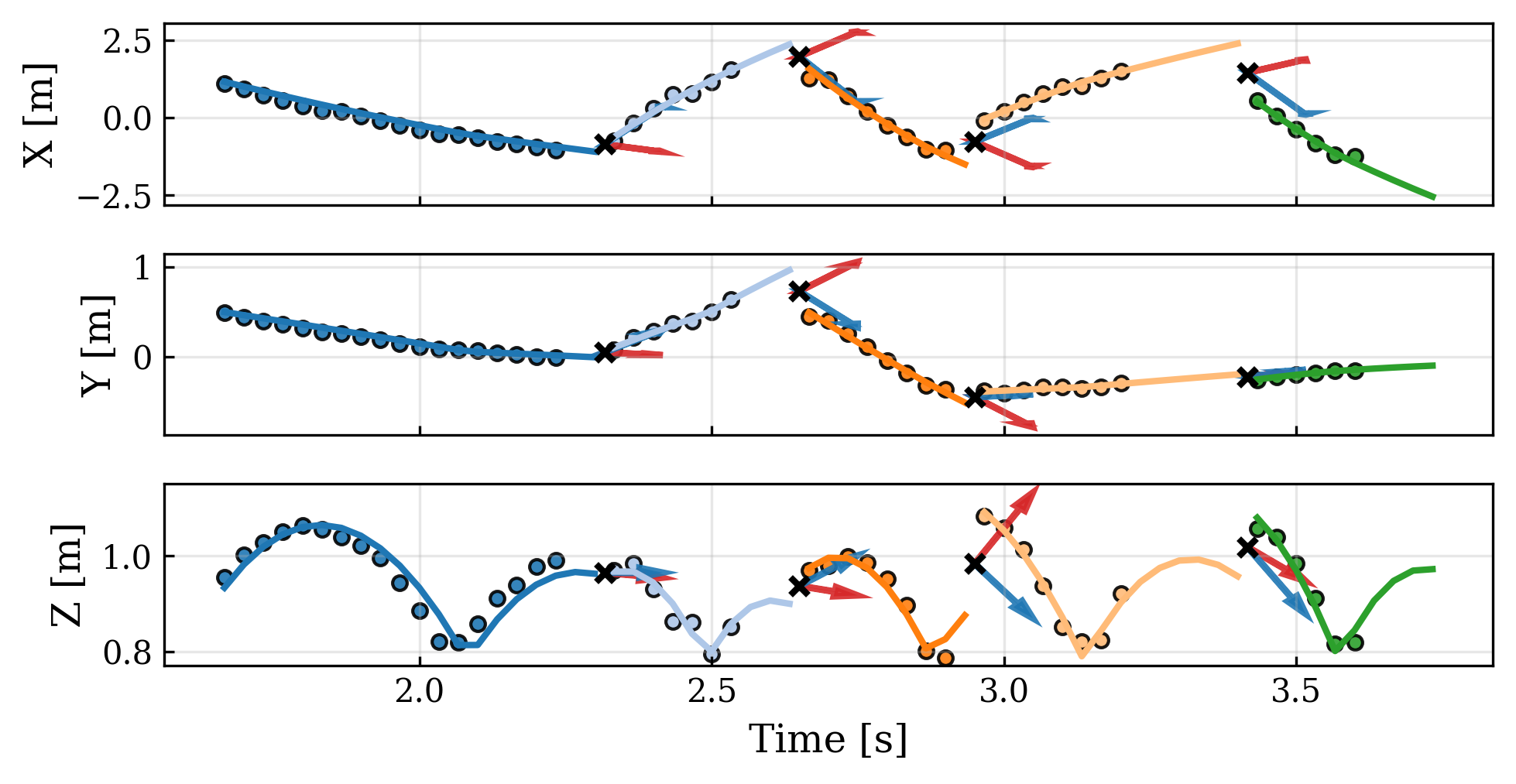}
    \vspace{-0.3cm}
    \Description{A chart with three vertically stacked subplots displaying the X, Y, and Z coordinates of a ball's trajectory in meters against time in seconds. In each subplot, discrete dots representing the network's predicted 3D points are closely fitted by a continuous solid line representing the physics-based ODE. Black crosses indicate estimated racket hit points along the trajectory. Originating from these crosses are red arrows, showing the ball's speed before the hit, and blue arrows, showing the ball's speed after the hit.}
    \caption{Physical consistency: A physics-based ODE (solid line) fits the network's predicted 3D points (dots) with high precision, confirming the physical consistency of our output. Crosses are estimated racket hit points and the corresponding red and blue arrow are respectively the ball speeds before and after, used for the racket stroke estimation.
    }
    \vspace{-0.4cm}
    \label{fig:ode_fit_3d}
\end{figure}
\noindent\textbf{Time Segmentation.} Using our detection method on the predicted 3D trajectories, we achieve F1-scores of 0.92 for table bounces and 0.83 for racket hits. The lower racket hit performance is mainly due to limited ball visibility near the players, making accurate annotation difficult.

\noindent\textbf{Inference Speed.} At inference time, the model processes more than $500$ points/s on a 10-year-old Titan X GPU, enabling reconstruction of thousands of games in minutes.
This contrasts with optimization-based methods such as LATTE-MV~\cite{etaat2025latte}, TT3D~\cite{gossard2025tt3d}, and MonoTrack~\cite{liu2022monotrack}, which require a separate fitting procedure for each segment during inference. We present quantitative speed comparisons in the supplementary materials. 
\section{Applications}
\subsection{Generative Model of Competitive Gameplay} 
As a key downstream application, we utilize the fidelity and diversity of our dataset for training a generative model of competitive gameplay.
High-quality, diverse generated samples serve as strong evidence that our \textbf{TT4D} dataset captures the underlying structure of the sport. \\[0.66ex]
\noindent\textbf{Generative Framework.} \,
We adopt the Conditional Flow Matching (CFM) framework~\cite{lipman2022flow, tong2024improving, tong2023simulation}.
Let $\mathcal{D}$ be our dataset of reconstructed trajectories, $p_0 = \mathcal{N}(0, I)$ be a base distribution, and $p_{\text{data}}$ be the empirical distribution over $\mathcal{D}$. Let $\tau$ denote a horizon of 20 observations, and $c$ denote a history of 10 observations. 
We learn a conditional vector field $v_\theta$ whose ODE solution $\frac{d\tau_t}{dt} = v_\theta(\tau_t, t \mid c)$ transports $p_0$ onto $p_{\text{data}}$.
We use the flow-matching loss $\mathcal{L}_{\text{CFM}} = \mathbb{E} \left[ \left\| v_\theta(\tau_t, t \mid c) - u_t \right\|_2^2 \right]$, where $\tau_t$ is a simple interpolation between samples from $p_0$ and $p_{\text{data}}$, and $u_t$ is the target velocity.
This conditional formulation enables the model to generate full 20-step future trajectories with high physical fidelity given past context.
Full details on the model architecture and training hyperparameters are in the supplementary material. \\[0.66ex]
\noindent\textbf{Evaluation.}
We generate 10{,}000 multi-segment rallies by autoregressively rolling out the model: starting from a 10-observation history from our dataset, we generate the next 20 observations and then repeatedly invoke the flow model on the 10 most recent generated observations. These rallies are clipped using the same 3D-based time segmentation scheme outlined in \Cref{subsec:3dannotation}. Only 6 out of the 10{,}000 generated rallies failed the time segmentation stage.
\noindent We evaluate the generated rallies using the same rigorous filtering and diversity metrics which we used to create the dataset.
As shown in \Cref{fig:flow_matching_eval}, the distribution of the Physics-Based ODE Fit error for our generated samples (Gen Mean: 8.72 cm) is remarkably similar to and even slightly better than that of the real data (Data Mean: 10.77 cm).
This confirms that our generative model, trained on TT4D, produces physically plausible trajectories.
We also demonstrate proper alignment in inter-hit time distribution. Though the generated gameplay distribution exhibits less spread, it nonetheless exhibits speeds at both ends of the real-world spectrum. Hence, we should expect a broad class of behaviors in our dataset. We provide qualitative visualizations in the supplementary material, which show coherent long-horizon behavior. 

\subsection{Racket Reconstruction}
\label{subsec:racket_reconstruction}
Our high-fidelity 3D ball trajectories and dense spin vectors provide a robust foundation for reconstructing racket motion. 
Since direct monocular tracking of the racket is often unreliable due to its high velocity, small size, and frequent occlusions, we leverage the reconstructed ball state instead.
We obtain the racket stroke parameters, defined as the racket's velocity and orientation at impact, from solving an inverse control problem.
By optimizing these stroke parameters such that the simulated post-impact ball trajectory accurately matches our observed 3D trajectory and bounce timing, we can recover the racket stroke that produced the observed ball trajectory. 
We evaluate our approach against motion capture ground truth using infrared markers on the racket, across 92 recorded strokes.
We obtain a mean orientation error of $26.4 \pm 4.4^\circ$ and a velocity error of $0.58 \pm 0.40$ m/s in the world frame, for an average impact speed of 3.72 m/s.
The residual error is mainly in the world-frame Z-velocity and rotation about the X-axis (racket open/closed angle), likely due to unknown racket bounce properties.

\subsection{Other Applications}
\label{sec:otherapplications}
\noindent We believe this dataset unlocks significant potential for \textbf{sports analytics}, enabling advanced match visualization, human behavior analysis ~\cite{muelling2014learning}, and new training paradigms.
For example, a ball launcher~\cite{dittrich2023aimy} could be conditioned to reproduce an opponent's specific playing style, offering individualized preparation for professional players.
The dataset also opens new frontiers in \textbf{robot learning}, e.g. providing the foundation for imitation learning, where a humanoid robot could learn to emulate professional-level strokes and rallies \cite{su2025hitter, zhang2026learning}. We demonstrate this by training a motion tracking policy \cite{liao2025beyondmimic} on a retargeted motion \cite{araujo2025retargeting} from our dataset. Implementation details and results may be found in the supplementary material. 
Furthermore, TT4D provides the necessary data for training behavior-aware models, such as anticipatory robotic policies, allowing agents to reason about the strategic decisions made by other players.

\begin{figure}[t]
    \vspace{-0.0cm}
    \centering

    \begin{subfigure}{0.87\linewidth}
        \centering
        \includegraphics[width=\linewidth]{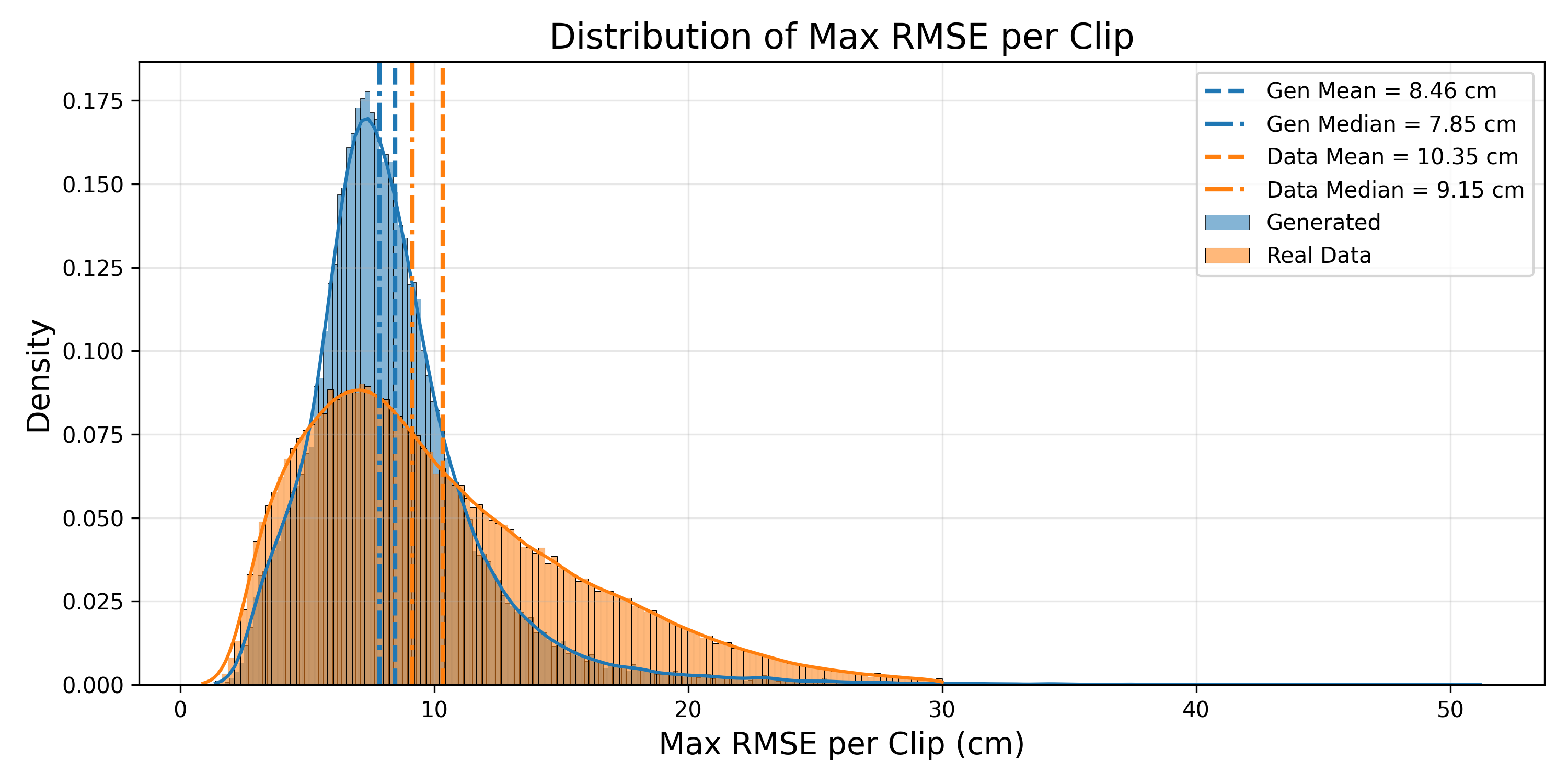}
        \label{fig:max_rmse_comp_top}
    \end{subfigure}

    \vspace{-0.4cm}  

    \begin{subfigure}{0.87\linewidth}
        \centering
        \includegraphics[width=\linewidth]{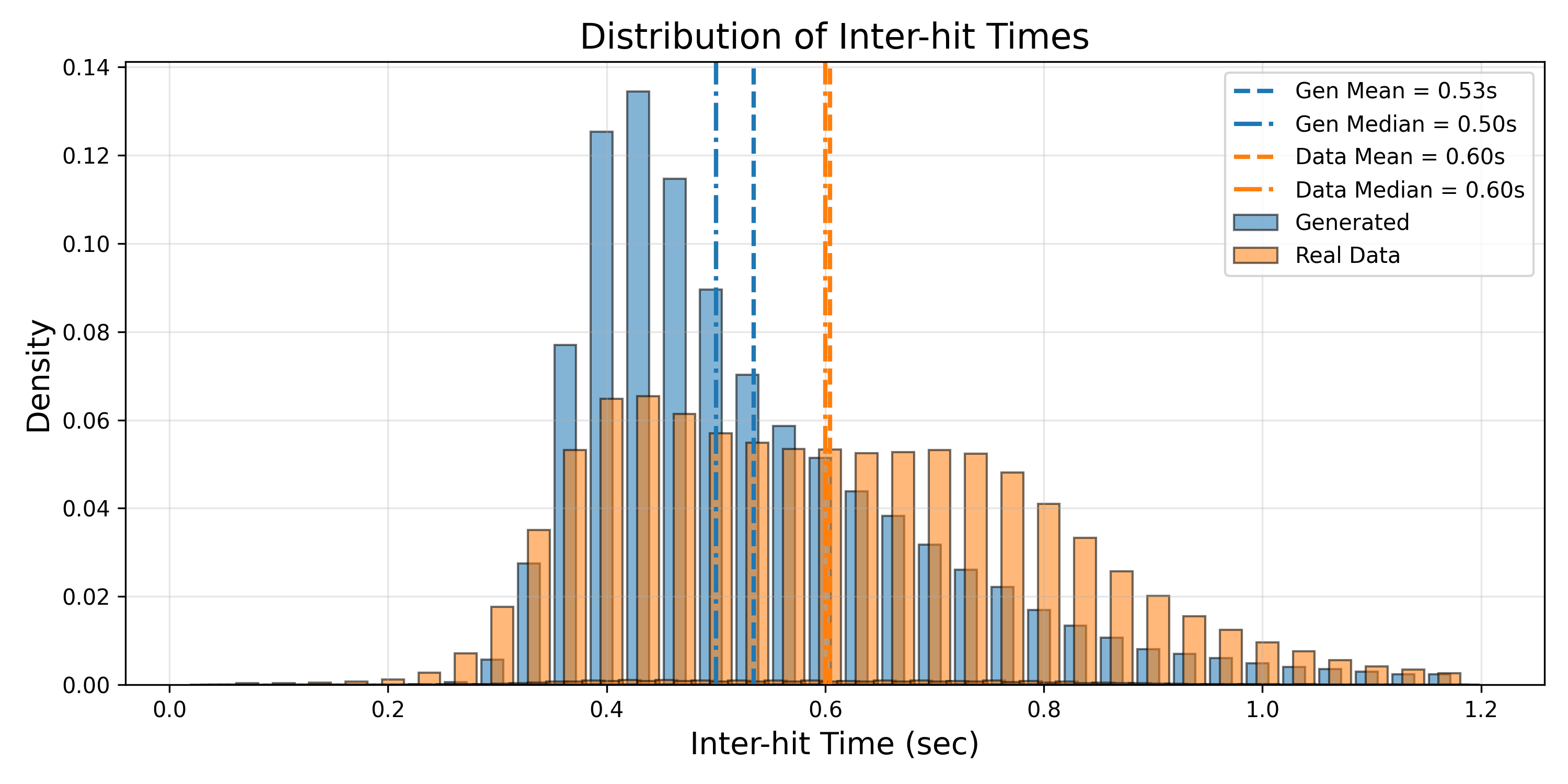}
        \label{fig:inter_hit_times_comp_bottom}
    \end{subfigure}

    \vspace{-0.5cm}
    \Description{Two vertically stacked histograms comparing generated data in blue and real data in orange. The top histogram displays the distribution of maximum ODE fit errors (RMSE), showing both distributions heavily overlapping and peaking near the lower end of the axis. The bottom histogram displays the distribution of inter-hit times in seconds, showing both distributions overlapping with a primary peak between 0.4 and 0.6 seconds and sharing a similar right-skewed tail.}
    \caption{
    \textbf{(Top)} Physical plausibility: The distribution of maximum ODE fit errors (RMSE) for generated rallies (blue) closely matches the real gameplay distribution (orange), indicating that our generative model produces physically valid 3D trajectories.  
    \textbf{(Bottom)} Temporal realism: The inter-hit time distribution of generated rallies (blue) aligns closely with real gameplay (orange), demonstrating the model captures tempo and exchange dynamics.
    }
    \label{fig:flow_matching_eval}
    \vspace{-0.3cm}
\end{figure}

\section{Conclusion}
\label{sec:conclusion}
We presented the \textbf{Lift-First Pipeline} for multimodal sports reconstruction from monocular videos, which lifts entire unsegmented rallies to 3D before performing any time segmentation. 
By working in the 3D domain, our approach avoids the fragility of 2D-based time segmentation under occlusions and missing detections. 
This is enabled by a Full-Sequence Lifting Network trained on a large-scale synthetic dataset of $3$M full points that also models the pre-serve toss.
While demonstrated on table tennis, our general pipeline can be readily extended to other sports such as tennis or badminton. \\[0.66ex]
\noindent The result of this pipeline is the \textbf{TT4D dataset}, a \gameplaytime{} hour high-fidelity multimodal dataset with 3D ball trajectories, 3D human meshes, and, for the first time at this scale, 3D spin vectors and precise 3D-derived time segmentations.
We demonstrated the fidelity and utility of TT4D through physics-based validation and downstream applications, including racket-contact estimation and a generative model of competitive rallies.

\clearpage

\begin{acks}
This research was supported in part by DAF Air Force Research Laboratory project Provably Correct Design of Adaptive Hybrid Neuro-Symbolic Cyber Physical Systems under Award FA8750-23-C-0080. 
\end{acks}

\bibliographystyle{ACM-Reference-Format}
\bibliography{main}

\appendix
\clearpage

\setcounter{figure}{0}
\setcounter{table}{0}
\setcounter{equation}{0}
\renewcommand{\thefigure}{SM\arabic{figure}}
\renewcommand{\thetable}{SM\arabic{table}}
\renewcommand{\theequation}{SM\arabic{equation}}

\begin{strip}
\begin{center}
    {\Huge \textbf{TT4D: A Pipeline and Dataset for Table Tennis 4D Reconstruction From Monocular Videos}} \\[3ex]
    {\huge \textbf{Supplementary Material}} \\[3ex]
\end{center}
\end{strip}

\noindent We provide additional details on our dataset generation and evaluation pipeline in this supplementary material. 
Moreover, additional experiments are evaluated and some qualitative examples are analyzed.
In addition, we also illustrate one example rally in the dataset (Figure \ref{fig:supp_qualitative_reconstruction_example}) as well as one example rally generated by our generative algorithm (Figure \ref{fig:supp_qualitative_flow}).
We publish our dataset upon acceptance.
Our gameplay visualizations are powered by Viser \cite{yi2025viser}.

\section{Data Preprocessing Details}
This section provides implementation details for the \textbf{Data Acquisition and Preprocessing} stage of our pipeline.

\subsection{Video Clipping and Deduplication}
\noindent\textbf{Video Clipping}
Our reconstruction pipeline starts with a set of multi-hour, uncut table tennis competition videos.
This is processed into a set of reconstructed points by first progressing through two stages of clipping.
The first stage clips the broadcast video at frames when the scoreboard advances.
We detect the change in score by first identifying the scoreboard using YOLO ~\scite{10533619} and then recovering its text using PaddleOCR \scite{cui2025paddleocr30technicalreport}. The text associated with the player names also allows us to discern singles from doubles gameplay.

\noindent This scoreboard-based clip contains a significant number of frames preceding and following the actual gameplay.
To trim the clips down further, the second stage identifies the approximate start and end of the point by using the 2D ball trajectory, $b_{\text{2D}}(n)$.
For this stage, we use the LATTE-MV ball tracker \scite{etaat2025latte} to obtain $b_{\text{2D}}(n)$. This ball tracker uses the full TracknetV3 model, which includes the inpainting module. Note that the inpainting module is removed when we compute $b_{\text{2D}}(n)$ for reconstruction. 
The key idea of this stage is that while the oscillations of $b_{\text{2D}}(n)$ \textit{cannot} reliably solve hit-point identification, it \textit{can} be used to determine the approximate start and end of the point. 
This procedure is outlined in Algorithm \ref{alg:second-clipping-stage-all}.
Note that this stage may fail and produce clips that do not exhibit any gameplay. This is not a problem, however, since these clips are removed in the filtering stage.
\begin{algorithm*}[t]
\caption{Second Clipping Stage and Helper Functions}
\label{alg:second-clipping-stage-all}
\small

\textbf{Helper Functions.}

\textsc{ComputeConstantIntervals}$(Signal,\ strength\_th,\ time\_th)$:  
\hspace*{1em}\textit{Inputs:} A 1D signal; a magnitude threshold defining ``near-zero'' values; a minimum interval duration.  
\hspace*{1em}\textit{Returns:} All contiguous intervals where the signal remains below the threshold for longer than the required duration. \\[0.75ex]

\textsc{FindOverlappingIntervals}$(Intervals_1,\ Intervals_2)$:  
\hspace*{1em}\textit{Inputs:} Two sorted lists of disjoint intervals.  
\hspace*{1em}\textit{Returns:} All sub-intervals formed by intersections of the two lists. \\[0.75ex]

\textsc{FindComplementaryIntervals}$(Signal,\ Intervals,\ min\_length)$:  
\hspace*{1em}\textit{Inputs:} A signal; a list of covered intervals; a minimum acceptable gap length.  
\hspace*{1em}\textit{Returns:} All uncovered intervals (gaps) of at least the minimum length. \\[0.75ex]

\textsc{BestPointEstimate}$(corr\_x,\ Intervals,\ min\_length,\ peak\_th)$:  
\hspace*{1em}\textit{Inputs:} A correlation signal; candidate gameplay intervals; a minimum interval length; a peak magnitude threshold.  
\hspace*{1em}\textit{Returns:} The index of the interval with the highest number of above-threshold peaks (or $-1$ if none qualify). \\[2ex]

\textbf{Main Procedure: Second Clipping Stage}

{\small
\begin{algorithmic}[1]

\Function{FindApproximateStartEnd}{$x,\ y$}

\Comment{1. Kernels and thresholds}
\State $horizontal\_kernel \gets [1\!\times\!15,\ 0\!\times\!6,\ -1\!\times\!15]$
\State $vertical\_kernel \gets -[1\!\times\!20,\ -1\!\times\!20]$

\Comment{2. Zero-valued intervals in both $x$ and $y$}
\State $I_x \gets \textsc{ComputeConstantIntervals}(x, 1, 20)$
\State $I_y \gets \textsc{ComputeConstantIntervals}(y, 1, 20)$
\State $Zero \gets \textsc{FindOverlappingIntervals}(I_x, I_y)$

\Comment{3. Candidate gameplay intervals}
\State $Possible \gets \textsc{FindComplementaryIntervals}(x,\ Zero,\ 80)$

\State $\delta x[t] \gets x[t+1] - x[t]$
\State $\delta y[t] \gets y[t+1] - y[t]$
\State $corr\_x \gets \mathrm{Convolve}(\delta x,\ \mathrm{reverse}(horizontal\_kernel))$
\State $corr\_y \gets \mathrm{Convolve}(\delta y,\ \mathrm{reverse}(vertical\_kernel))$

\If{$|corr\_x| \ne |corr\_y|$} \State \textbf{return} $(-1, -1)$ \EndIf

\Comment{5. Mask correlations outside candidate intervals}
\State Construct Boolean mask over $[0, |corr\_x|)$ based on $Possible$
\State Zero out entries of $corr\_x$ and $corr\_y$ where mask is False

\Comment{6. Extract correlation-domain intervals}
\State $Signal \gets \textsc{FindComplementaryIntervals}(corr\_x,\ Zero,\ 0)$

\For{each $(s,e)$ in $Signal$}
    \If{$|(e-s)-50| < 20$}
        \State $dot \gets \sum_{t=s}^{e} corr\_x[t]\, corr\_y[t]$
        \If{$|dot| \ge 30000$} \State Zero out $corr\_x,\ corr\_y$ on $[s,e]$ \EndIf
    \EndIf
\EndFor

\Comment{7. Choose best gameplay interval}
\State $idx \gets \textsc{BestPointEstimate}(corr\_x,\ Signal,\ 60, 50)$
\If{$idx = -1$} \State \textbf{return} $(-1, -1)$ \EndIf
\State $(start, end) \gets Signal[idx]$

\Comment{8. Reject clips where the ball remains still too long}
\State $dx \gets \Delta x_s,\ dy \gets \Delta y_s$
\If{$|\{t : dx[t]=0,\ x[t]\neq 0,\ dy[t]=0,\ y[t]\neq 0\}| > 30$}
    \State \textbf{return} $(-1, -1)$
\EndIf

\State \textbf{return} $(start, end)$

\EndFunction

\end{algorithmic}
}

\end{algorithm*} \\[0.66ex]
\noindent\textbf{Duplicated Frame Removal}
While processing online table-tennis footage, we observed that certain frames were duplicated within the video stream.
This phenomenon typically arises when the frame rate used for camera capture differs from that used during rendering or re-encoding.
In such cases, the renderer/encoder fills missing timestamps by repeating the temporally nearest frames, thereby matching the target output frame rate.
We identified three cases: (i) periodic duplicated frames, characterized by a first duplicated index $s$ and a duplication period $f$, (ii) aperiodic duplicated frames, and (iii) no duplicated frames.
The periodic case was the most common in the processed videos.
Although imperceptible to the human eye, this frame-duplication artifact causes the tracked ball to appear at the same position for two consecutive frames.
As a consequence, the reconstruction algorithm falsely interprets these repetitions as sudden deceleration or temporary stops, which deteriorates trajectory estimation accuracy.
A direct pixel-wise absolute difference is not a reliable measure of frame similarity.
Even when two frames are visually identical, their difference image resembles approximately white noise due to encoder and compression artifacts.
The aggregated absolute difference over all pixels in such cases can be comparable to that between genuinely different consecutive frames.
This motivates the use of \ac{SSIM}, which evaluates perceptual similarity by comparing luminance, contrast, and structural content.
Using a high \ac{SSIM} threshold allows us to detect and discard duplicated frames, though occasional false positives and false negatives may still occur.
To robustly detect periodic duplication, let $\{u_i\}_{i=1}^{N}$ denote the sorted indices of duplicated frames.
We consider the inter-duplicate spacings \(\Delta_i = u_{i+1} - u_i,\text{for} \; i = 1,\dots,N-1,\) and ignore trivial repetitions with $\Delta_i = 1$.
Throughout this paper, we use $|\mathcal{A}|$ to denote the cardinality of a set $\mathcal{A}$.
We estimate the duplication period as the modal spacing
\begin{equation}
    \hat{f}
    = \arg\max_{d \in \mathbb{N}}
      \left|
 \left\{ i \mid \Delta_i = d,\, \Delta_i > 1 \right\} \right|.
\end{equation}
Assuming periodic duplicated frames follow the model $u_k = s + k f$, the start offset is obtained as the most frequent residue modulo $\hat{f}$:
\begin{equation}
\hat{s}
= \arg\max_{r \in \{0,\dots,\hat{f}-1\}}
\sum_{i=1}^{N} \mathbf{1}\!\left( u_i \equiv r \; (\mathrm{mod}\;\hat{f}) \right),
\end{equation}
where $\mathbf{1}(\cdot)$ denotes the indicator function.
If fewer than three duplicated indices exist, or if no non-trivial spacings are observed, periodicity cannot be estimated reliably and the case is treated as aperiodic.
Finally, the video is re-encoded with the corrected frame rate after removing all detected duplicated frames.

\subsection{Camera Calibration}
The camera parameters are not provided for online videos, so the calibration must be performed without a dedicated calibration pattern.
We therefore adopt the TT3D approach~\scite{gossard2025tt3d}, which uses the table itself as the calibration object.
The world frame is defined by the standard table geometry, and the four table corners serve as known 3D reference points.
A segmentation model first extracts the table mask, after which table edges are detected using a Hough line transform and intersected to obtain the corner locations.
With these correspondences, the camera extrinsics, rotation $\mathbf{R}$ and translation $\mathbf{T}$ relative to the table frame, are estimated jointly with the focal length $f$.
The problem is formulated as a Perspective-$n$-Point task with unknown focal length and solved by iteratively minimizing the reprojection error over $(\mathbf{R}, \mathbf{T}, f)$, enabling robust calibration even under partial occlusions and in unconstrained broadcast footage. \\[0.66ex]
In contrast to the original algorithm in \scite{gossard2025tt3d}, our segmentation model is based on the UNet++ architecture~\scite{zhou2018unetplusplus} with the bigger Efficientnet-B0 backbone~\scite{tan2019efficientnet}.
For training, we use a combination of Binary Cross Entropy and DICE Loss.
With these modifications, we are able to obtain high quality camera calibrations for our dataset.

\subsection{Player Reconstruction}
\label{sup:player-reconstruction}
We use 4DHumans \scite{goel2023humans4d} to track and reconstruct the human players.
The human body is represented using the SMPL model \scite{loper2015smpl}, and the parameters are given in the camera frame.
We orient the SMPL mesh in the world frame using the camera rotation.
To position the SMPL mesh in the world frame, we first use the camera extrinsics to compute the homography transform from the image plane to the ground plane.
Next, we assume that the rotated 3D keypoint closest to the ground \textit{is} on the ground plane, and we use the homography matrix to map the corresponding 2D keypoint to its 3D location on the ground plane. \\[0.66ex]
\noindent We verify global 3D consistency by checking that the average location of all reconstructed 3D human meshes is plausibly near the 3D table.
This step is particularly important to filter non-players like referees or the crowd, common misdetections that are especially present for doubles gameplay.

\section{Lifting Network Details}
\label{sup:liftingnetworkdetails}

\begin{algorithm*}[t]
\caption{Stitching Algorithm}
\label{alg:stitching}
{
\small
\begin{algorithmic}[1]
\State \textbf{Input:} Data pools $P_{\text{throw}}, P_{\text{serve}}, P_{\text{return}}$
\State \textbf{Output:} A list of stitched trajectory segments $Point$ \\

\Function{BuildStitchedPoint}{}
\State $Point \gets \text{EmptyList}()$ \\

\State $initial\_conditions\_throw \gets P_{\text{throw}}.\text{Sample}()$ 
\Comment{1. Simulate the initial ball toss}
\State $throw\_seg \gets \text{Simulate}(initial\_conditions\_throw)$
\State $r_{\text{start}} \gets throw\_seg.\text{end\_positon}$ 
\State $Point.\text{Add}(throw\_seg)$ \\

\State $serve\_seg \gets \text{StitchNextSegment}(r_{\text{start}}, P_{\text{serve}})$
\Comment{2. Stitch the serve segment}
    \If{$serve\_seg$ is \textbf{null}} 
        \State \textbf{return} \text{FailedPoint}
    \EndIf
\State $Point.\text{Add}(serve\_seg)$
\State $r_{\text{start}} \gets serve\_seg.\text{end\_position}$

\\
\For{$i \gets 1 \text{ to MAX\_Point\_LENGTH}$}
\Comment{3. Recursively stitch return segments}
     \State $return\_seg \gets \text{StitchNextSegment}(r_{\text{start}}, P_{\text{return}})$
         \If{$return\_seg$ is \textbf{null}} 
                \State \textbf{return} \text{FailedPoint}
        \EndIf
        \State $Point.\text{Add}(return\_seg)$
        \State $r_{\text{start}} \gets return\_seg.\text{end\_position}$
    \EndFor \\
    
    \State \textbf{return} $Point$
\EndFunction \\

\Function{StitchNextSegment}{$r_{\text{start}}, \text{Pool}$}
    \Comment{Multiple tries to find the next segment}
    \For{$i \gets 1 \text{ to MAX\_ATTEMPTS}$} \\

        \State $prior \gets Pool.\text{FindClosestPrior}(r_{\text{start}})$
        \Comment{Use initial velocity and spin of close match}
        \State $v_{\text{new}} \gets prior.\text{velocity}$
        \State $\omega_{\text{new}} \gets prior.\text{spin}$

        \State $trajectory \gets \text{Simulate}(r_{\text{start}}, v_{\text{new}}, \omega_{\text{new}})$ \\

        \If{(\text{IsValidTrajectory}(trajectory)) }
        \Comment{Check if we simulated a valid table tennis trajectory}
            \State \textbf{return} $trajectory$
        \EndIf
    \EndFor \\
    
    \State \textbf{return} \textbf{null} 
\EndFunction
\end{algorithmic}
}
\end{algorithm*}

\subsection{Network Architecture}
\begin{figure}[ht]
    \centering
    \includegraphics[width=0.99\linewidth]{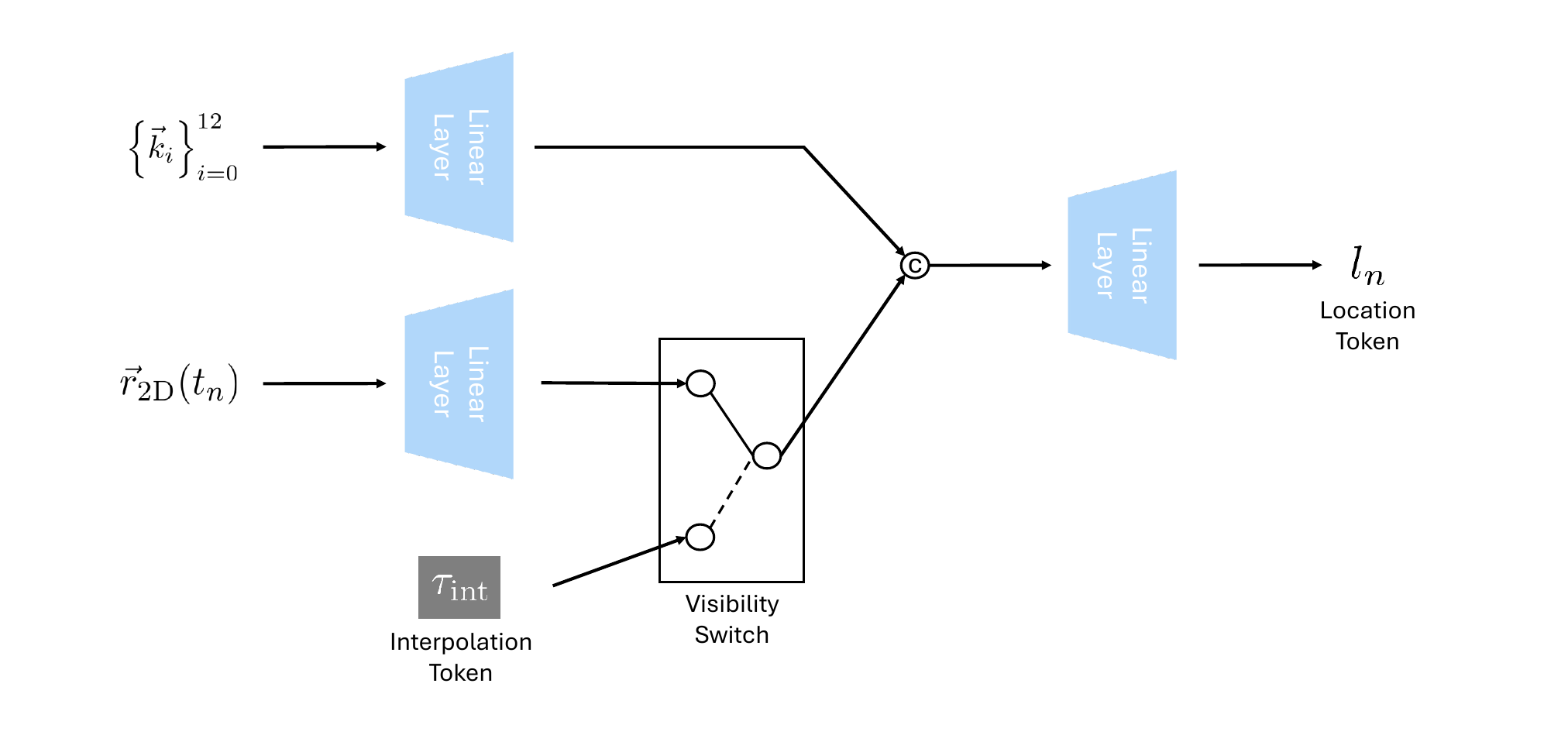}
    \Description{A block diagram illustrating the Disentangled Context Embedding process. Two parallel input paths for Table Keypoints and detected 2D ball coordinates are shown being projected into higher dimensional vectors. A conditional routing step shows that a learnable interpolation token is substituted for the ball vector if the ball is not correctly detected. The diagram then shows these two resulting vectors being concatenated together to compute the final output, designated as the location token.}
    \caption{Schematic illustration of the Disentangled Context Embedding (DCE). Table Keypoints $\{ \vec{k}_i \} _{i=0}^{12}$ and detected 2D ball coordinates $\vec{r}_\text{2D}$ are projected to higher dimensional vectors. If a ball is not correctly detected, a learnable interpolation token $\tau_\text{int}$ is used instead of the ball vector. Finally, both vectors are concatenated and the final location token $l_n$ is computed.}
    \label{fig:embedding_supp}
\end{figure}

\begin{figure}[t]
    \centering
    \includegraphics[width=0.99\linewidth]{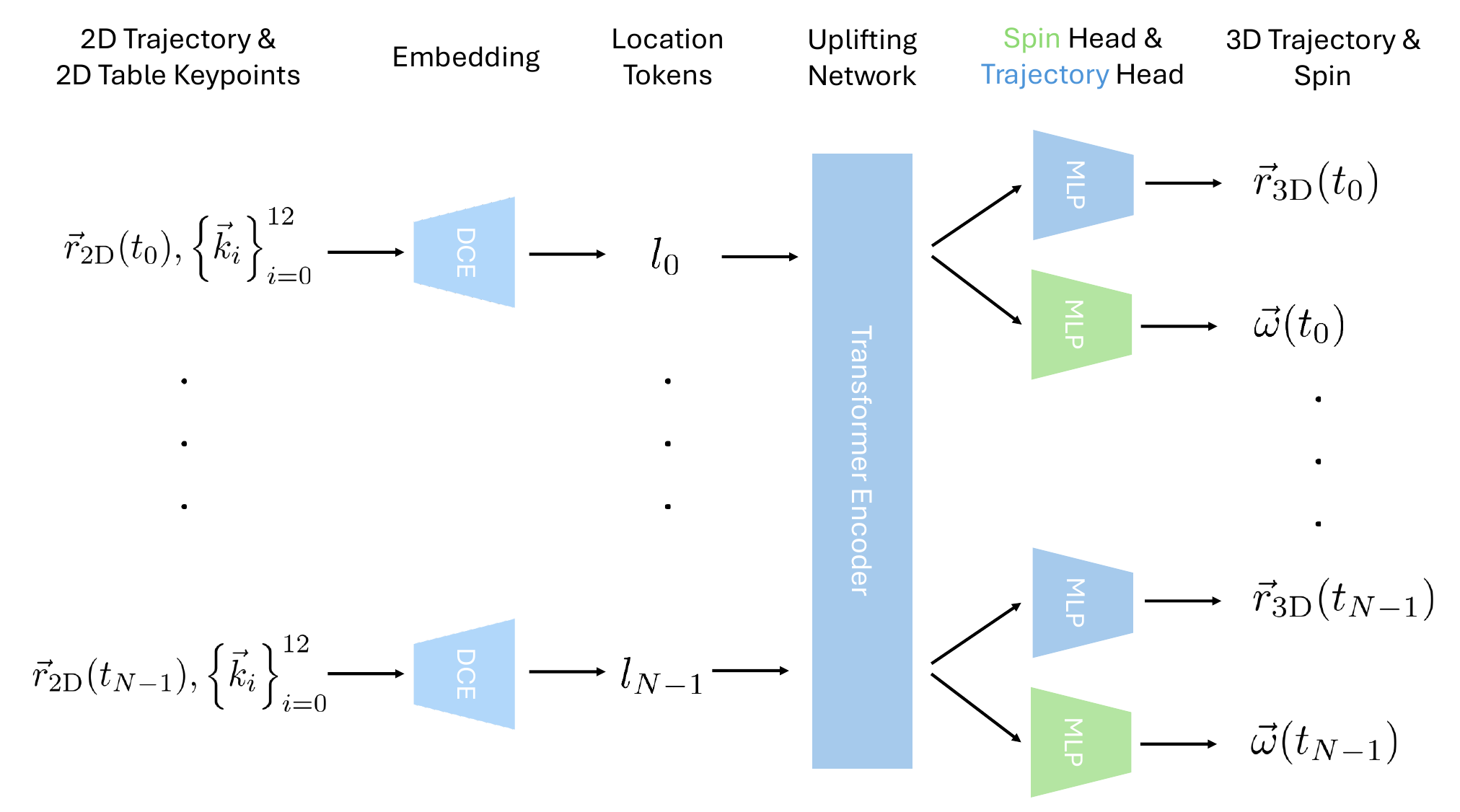}
    \Description{A high-level architectural diagram of the Full-Sequence Lifting Network. The process starts on the left with inputs of 2D trajectories and table keypoints. These are passed through a Disentangled Context Embedding (DCE) block to produce a sequence of location tokens ($l_0$ to $l_{N-1}$). These tokens enter a central "Uplifting Network" block, identified as a Transformer Encoder. The output tokens from this encoder are then split into parallel heads: a Trajectory Head and a Spin Head, both represented as MLP blocks. The final outputs on the right are the 3D ball positions $r_{3D}(t_n)$ and 3D spin vectors $\omega(t_n)$ for each corresponding timestep.}
    \caption{Schematic illustration of the Lifting Network.
First, the detected ball positions $\vec{r}_{2D}(t_n)$ and table keypoints $\{ \vec{k}_i \} _{i=0}^{12}$ are embedded to obtain the sequence of location tokens $\{ l_n \} _{n=0}^{N-1}$.
The lift network transforms these tokens and each output is further processed by a Spin Head to obtain the spin vector $\vec{\omega}(t_n)$ and Trajectory Head to obtain the 3D position $ \vec{r}_\text{3D}(t_n)$ for each timestep $t_n$. We refer to \cite{kienzle2026uplifting} for further information on the transformer architecture.
}
    \label{fig:uplifting_network_supp}
\end{figure}

Our lifting network builds upon the 2D-to-3D lifting transformer introduced by \citet{kienzle2026uplifting}.
Its core design takes a sequence of $N$ 2D ball detections $\{ \vec{r}_{2D}(t_n) \} _{n=0}^{N-1} \in \mathbb{R}^{N \times 2}$, the corresponding exact times $\{ t_n \}_{n=0}^{N-1} \in \mathbb{R}^{N}$ that can be derived from the videos framerate, and a set of 13 predefined 2D table keypoints $\{ \vec{k}_i \} _{i=0}^{12} \in \mathbb{R}^{13 \times 2}$ as input.
These 13 keypoints implicitly provide the network with the full camera calibration information needed for trajectory lifting. \\[0.66ex]
\noindent The ball detections and 13 table keypoints are first embedded into sequence of $N$ "location tokens" $ \{ l_n \} _{n=0}^{N-1} \in \mathbb{R}^{d_{\text{model}}}$ with embedding dimension $d_{\text{model}}$.
We implement a custom Disentangled Context Embedding (DCE), that projects the ball positions and table keypoints separately into higher dimension vectors, keeping the ball information disentangled from the camera calibration information. 
This now allows replacing a ball vector with a learnable interpolation token in case the ball was not correctly detected.
Finally, ball vector and table keypoints vector are concatenated and projected to compute the location token.
This is visualized in \cref{fig:embedding_supp}. 
 \\[0.66ex]
\noindent To effectively integrate these interpolation tokens into the transformer without degrading the feature quality of observed frames, we adopt the Deferred Upsampling Token Attention (DUTA) mechanism \cite{einfalt2023uplift}. 
Since the learnable interpolation tokens initially contain no specific ball information beyond their temporal embedding, allowing valid location tokens to attend to them in the first layers can introduce significant noise and corrupt the high-fidelity representations of detected ball positions. 
DUTA addresses this by restricting the self-attention mask in the initial transformer layers: 
valid location tokens are prohibited from attending to interpolated location tokens, while interpolated location tokens are permitted to attend to valid location tokens. 
This ensures that the interpolated location tokens can effectively aggregate context from their neighbors to estimate missing positions without deteriorating the features of the rest of the sequence. \\[0.66ex]
\noindent Moreover, a key architectural extension is the removal of the learnable "spin token" $\mathbf{s}$ used in the baseline, as this cannot be extended for processing full points.
Instead of predicting a single initial spin vector, we modify the network to predict spin in a dense, per-frame manner by applying a small Spin Head, consisting of a small 3-layer MLP, to \textit{every} output location token $l_n$.
Consequently, the network outputs a full sequence of dense spin vectors $\{ \vec{\omega}(t_n) \}_{n=0}^{N-1} \in \mathbb{R}^{N \times 3}$ alongside the 3D trajectory sequence $\{ \vec{r}_\text{3D}(t_n) \} _{n=0}^{N-1} \in \mathbb{R}^{N \times 3}$.
This dense prediction provides a much richer output than prior methods and resolves the ambiguity of defining an "initial" spin for an unsegmented sequence.
A schematic overview of our modified lifting network is presented in \Cref{fig:uplifting_network_supp}.

\subsection{Synthetic Dataset Generation}
A network designed to process full points requires training data that reflects this new, more complex structure.
In contrast, previous methods~\scite{kienzle2025towards,kienzle2026uplifting} are only trained on smaller datasets of only 50k-140k individual segments, which can only be obtained from unreliable 2D-based time segmentation during inference.
To power our lifting network, we generate a new massive-scale synthetic dataset of approximately \textbf{3 million points}, with each point consisting of multiple stitched segments.
This dataset utilizes the MuJoCo~\scite{todorov2012mujoco} physics simulation environment, ensuring all trajectories are physically correct and include perfect, per-frame ground truth for 3D position, and spin. \\[0.66ex]
\noindent Simulating 3 million complete end-to-end points from purely random initial conditions is computationally infeasible.
To overcome this, we present a "stitching" pipeline, detailed in \Cref{alg:stitching}, that assembles points from a large pool of pre-validated, physically-correct segments.
The authors of~\scite{dambrosio2025achieving} and~\cite{kienzle2026uplifting} provide a large collection of initial conditions for serves and standard shots, which we utilize as a starting point for our pool.
Simulating a full point then becomes a recursive, hybrid process:
First, we simulate an initial segment (e.g., a serve) and find its final 3D position $\vec{r}_{N-1}^{\,3D}$.
This becomes the initial position for the next segment.
To determine the segments initial spin and velocity, we find the segment in our data pool whose initial position is the closest match to our current state.
We then simulate the trajectory and check if it is a valid table tennis shot segment (i.e., it clears the net and bounces on the opponent's side).
If valid, we "stitch" this new segment to our point and repeat the process until the point is complete. \\[0.66ex]
\noindent Furthermore, we identify that a major failure point for generalization to real-world data is the pre-serve \textit{ball toss}, a phase that is difficult to clip before lifting.
To make our network robust to this, we explicitly simulate an entirely new pool of synthetic "throw" segments.
Our final training points are assembled by first simulating a "throw" segment, stitching a "serve" segment from our pool to its apex, and then recursively stitching the subsequent return segments.
This novel, training data is the key that enables our network to learn the complex transitions between segments and handle unsegmented real-world videos.

\subsection{Ball Occlusion Model}
\label{subsec:ball-occlusion-model}

\begin{figure}
    \centering
    \includegraphics[width=1\linewidth]{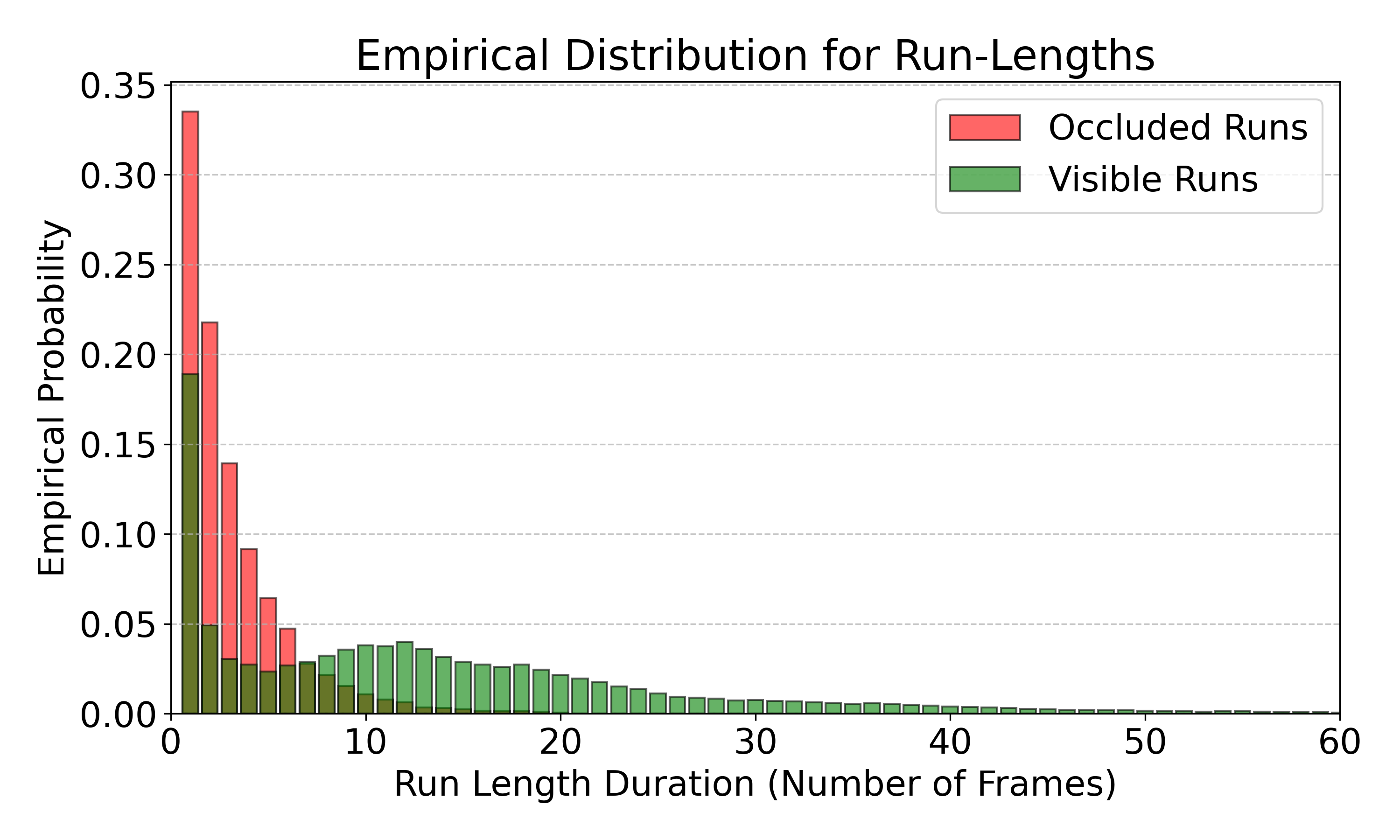}
    \caption{Empirical distribution of contiguous occluded/non-occluded segment lengths.}
    \label{fig:occlusion_model}
\end{figure}

During training, we randomly mask valid 2D detections using an empirically-derived occlusion model instead of a naive occlusion pattern (e.g. uniform random drop). We record the frequency of consecutive visible and occluded frames seen in the broadcast videos (illustrated in Figure \ref{fig:occlusion_model}). To generate a random mask, we alternate between the two distributions and sample sequences of visible and occluded segment lengths. 

\section{3D Hit-Point Identification Heuristics}
As described in the main paper, we identify hit-points and bounces by finding extrema in the 3D trajectory.
This 1D signal analysis is simple and robust.
To avoid identifying incorrect local extrema (e.g., from minor network noise) as valid events, we apply simple heuristics.
A 3D x-trajectory extremum is only considered a valid \textbf{hit point} if:
\begin{itemize}
    \item The time between it and the previous hit point of the same type (e.g., peak-to-peak) is at least 0.2 seconds.
    \item The absolute value of its x-coordinate (distance from the center net) is at least 0.3 meters, ensuring it is a full shot segment and not a small jitter.
\end{itemize}
A similar heuristic is applied to the z-trajectory for \textbf{bounce} detection. 
These simple filters are highly effective at isolating the true game events.

\section{Table Tennis Physics}
\label{sec:table_tennis_physics}

The physics of table tennis has been accurately modeled and can be used to predict or simulate ball trajectories.
In this section, we go over the different physical models used in the paper.

\subsection{Aerodynamics}
The ball's trajectory is defined by the following \ac{ODE}:

\begin{equation}
    m \bm{a} = \underbrace{-k_d ||\bm{v}|| \bm{v}}_\text{Drag} + \underbrace{k_m \bm{\omega}\times\bm{v}}_\text{Magnus force} + \bm{g}
    \label{eq:ode_aero}
\end{equation}
where $k_d=3.8\times10^{-4}$ is the drag coefficient and $k_m=3\times10^{-6}$ is the Magnus coefficient and $\bm{g}$ is the gravity vector.

\subsection{Table bounce}
The table bounce is defined by the following equation:
\begin{equation}
\begin{split}
\bm{v^{+}} & = \bm{A} \bm{v^{-}} + \bm{B} \bm{\omega^{-}}\\
\bm{\omega^{+}} & = \bm{C} \bm{v^{-}} + \bm{D} \bm{\omega^{-}}
\end{split}
\label{eq:linear_bounce_model}
\end{equation}
The matrices $\bm{A}, \bm{B}, \bm{C}, \bm{D}$ encode restitution and friction parameters. 
It distinguishes two cases: the ball can either have a rolling contact or a sliding contact.
The nature of the contact is determined by the coefficient:
\begin{equation}
\begin{aligned}
    \alpha &= \frac{\mu (1+\mathrm{COR})\, |v_z^-|}{v_s}, \\[4pt]
    v_s &= \sqrt{\, (v_x^- + \omega_y^- r)^2 + (v_y^- + \omega_x^- r)^2 \, }
\end{aligned}
\label{eq:alpha}
\end{equation}
where $COR$ is the coefficient of restitution, $\mu$ is the friction coefficient between the ball and the table and $v_s$ is the tangential velocity of the ball at the point of contact with the table surface.
If $\alpha \geqslant 0.4$, then the velocity of the ball's contact point is 0 and the ball is rolling: 

\begin{equation}
\resizebox{0.91\linewidth}{!}{%
$
\begin{aligned}
\bm{A}=&\left[\begin{array}{ccc}
1-\alpha & 0 & 0 \\
0 & 1-\alpha & 0 \\
0 & 0 & -COR
\end{array}\right] \quad &\bm{B}&=\left[\begin{array}{ccc}
0 & \alpha r & 0 \\
-\alpha r & 0 & 0 \\
0 & 0 & 0
\end{array}\right] \\
\bm{C}=&\left[\begin{array}{ccc}
0 & -\frac{3 \alpha}{2 r} & 0 \\
\frac{3 \alpha}{2 r} & 0 & 0 \\
0 & 0 & 0
\end{array}\right] \quad &\bm{D}&=\left[\begin{array}{ccc}
1-\frac{3}{2} \alpha & 0 & 0 \\
0 & 1-\frac{3}{2} \alpha & 0 \\
0 & 0 & 1
\end{array}\right]
\end{aligned}
$
}
\label{eq:sliding_mat}
\end{equation}

If $\alpha < 0.4$,  then the velocity of the ball's contact point is not 0 and the ball is sliding:

\begin{equation}
\resizebox{0.91\linewidth}{!}{%
$
\begin{aligned}
&\bm{A}=\left[\begin{array}{ccc}
0.6 & 0 & 0 \\
0 & 0.6 & 0 \\
0 & 0 & -COR
\end{array}\right] \quad \bm{B}=\left[\begin{array}{ccc}
0 & 0.4 r & 0 \\
-0.4 r & 0 & 0 \\
0 & 0 & 0
\end{array}\right] \\
& \bm{C}=\left[\begin{array}{ccc}
0 & -0.6 / r & 0 \\
0.6 / r & 0 & 0 \\
0 & 0 & 0
\end{array}\right] \quad \bm{D}=\left[\begin{array}{ccc}
0.4 & 0 & 0 \\
0 & 0.4 & 0 \\
0 & 0 & 1
\end{array}\right]
\end{aligned}
$
}
\label{eq:rolling_mat}
\end{equation}

\section{ODE Fit}
\label{sec:ode_fit}

Given the reconstructed 3D ball trajectory $\{ \vec{r}_\text{3D}(t_n) \}_{n=0}^{N-1}$ produced by our lifting network, we recover a physically consistent trajectory by estimating the initial ball state that best explains the observations under the physics model from\Cref{sec:table_tennis_physics}.  
We simulate ball flight using the aerodynamic \ac{ODE} from \Cref{eq:ode_aero} together with the table–bounce model in \Cref{eq:linear_bounce_model,eq:alpha}, integrating with a fixed-step \ac{RK4} scheme.
Table impacts are handled explicitly: at each step, we detect when the height crosses the table plane $z = h_{\mathrm{table}} = 0.78\,\mathrm{m}$ with downward velocity, interpolate to the exact impact time, snap the ball to the surface, and apply the corresponding sliding or rolling update from 
\Cref{eq:sliding_mat,eq:rolling_mat}.
The simulation is limited to at most two bounces, which covers all trajectories encountered in practice. \\[0.66ex]
\noindent Let $\vec{x}_0 = [\,\vec{p}_0^{\top},\,\vec{v}_0^{\top},\,\vec{\omega}_0^{\top}\,]^{\top}$ denote the unknown initial position, velocity, and spin at time $t_0$.
For any candidate $\vec{x}_0$, the physics model generates a simulated 3D trajectory
\[
\hat{\vec{r}}_\text{3D}(t_n;\vec{x}_0),
\]
which we align to the network predictions $\vec{r}_\text{3D}(t_n)$ through a robust nonlinear least-squares objective:
\begin{equation}
\resizebox{0.9\linewidth}{!}{%
$
\begin{aligned}
\min_{\vec{p}_0,\,
      \vec{v}_0,\,\vec{\omega}_0}
\quad
& \sum_{n \in \mathcal{I}}
  \rho\!\left(
    \big\|
      \hat{\vec{r}}_\text{3D}(t_n;\vec{p}_0,\vec{v}_0,\vec{\omega}_0)
      - \vec{r}_\text{3D}(t_n)
    \big\|_2
  \right)
\\[0.3em]
\text{s.t.}\qquad
& \vec{p}_{\min} \;\le\; \vec{p}_0 \;\le\; \vec{p}_{\max}, \\
& \|\vec{v}_0\|_{\infty} \;\le\; v_{\max}, \\
& \|\vec{\omega}_0\|_{\infty} \;\le\; \omega_{\max}.
\end{aligned}
$
}
\end{equation}
Here, $\mathcal{I}$ is the set of valid (non-missing) observations and $\rho(\cdot)$ is a Huber loss that downweights outliers. 
The bounds ensure that the recovered initial state remains physically plausible.  \\[0.66ex]
\noindent Solving this problem yields the optimal state
\[
\vec{x}_0^{\star} = 
  [\,\vec{p}_0^{\star},\vec{v}_0^{\star},\vec{\omega}_0^{\star}\,],
\]
together with a fully simulated, physically consistent trajectory $\hat{\vec{r}}_\text{3D}(t_n;\vec{x}_0^{\star})$.
For each segment, we report the root-mean-square error between simulated and reconstructed positions, as well as the number of predicted bounces, which allows us to distinguish single- and multi-bounce trajectories in downstream analyses.

\begin{figure}
    \centering
    \includegraphics[width=0.95\linewidth]{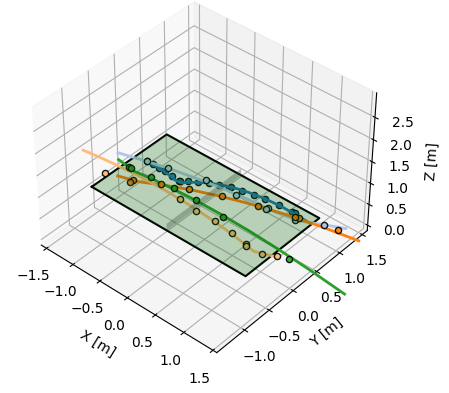}
    \Description{A 3D visualization showing multiple ball trajectory segments over a green table tennis table. The plot displays discrete circular markers representing the 3D points inferred by the Lifting Network, which are closely fitted by continuous, color-coded solid lines representing the physics-based ODE constrained trajectories. The trajectories illustrate different types of shots, including those that clear the net and bounce on the table surface.}
    \caption{Fitting the \ac{ODE} constrained ball trajectory to the 3D points inferred by the Lifting Network.}
    \label{fig:3d_ode_fit}
\end{figure}

\section{Racket Strike Reconstruction}
\label{sec:racket_inv_pb}
\noindent Given the pre- and post-impact ball velocity and spin, the impulse delivered by the racket is fully determined.
This allows us to recover the racket’s orientation $\mathbf{R}^{\text{w}}_{\text{r}}$ and velocity $\mathbf{V}^{\text{w}}_{\text{r}}$ at contact.
Although multiple solutions may exist in principle~\cite{liu2012racket}, any recovered pair $(\mathbf{R}^{\text{w}}_{\text{r}}, \mathbf{V}^{\text{w}}_{\text{r}})$ exactly reproduces the observed ball trajectory and is therefore consistent with the recorded impact.

\noindent To compute $(\mathbf{R}^{\text{w}}_{\text{r}}, \mathbf{V}^{\text{w}}_{\text{r}})$, we formulate an \ac{OCP} that minimizes the L2 distance between the predicted and observed bounce locations.
We use a single-shooting formulation with an $\mathrm{RK4}$ integrator, enabling us to propagate the full ball-flight ODE, including the Magnus effect, unlike the simplified models used in~\cite{liu2012racket}.
Racket–ball contact is modeled using the linear impact formulation of 
\scite{nakashima2010modeling}, expressed in the racket reference frame,
\begin{equation}
\begin{aligned}
    \bm{v}_{\mathrm{r}}^{+} &= \bm{A}\,\bm{v}_{\mathrm{r}}^{-} 
        + \bm{B}\,\bm{\omega}_{\mathrm{r}}^{-}, \\
    \bm{\omega}_{\mathrm{r}}^{+} &= \bm{C}\,\bm{v}_{\mathrm{r}}^{-} 
        + \bm{D}\,\bm{\omega}_{\mathrm{r}}^{-},
\end{aligned}
\label{eq:linear_bounce_model_rewrite}
\end{equation}
where $\bm{v}_{\mathrm{r}}^{-}$ and $\bm{\omega}_{\mathrm{r}}^{-}$ denote the incoming 
linear and angular ball velocities in the racket frame, and 
$\bm{v}_{\mathrm{r}}^{+}$, $\bm{\omega}_{\mathrm{r}}^{+}$ are the outgoing quantities. 
The model was originally developed for inverted-rubber rackets, 
which dominate competitive play; we therefore use a single fixed parameter set in all experiments.\\[0.66ex]
The matrices used in the model are defined as follows:
\begin{equation}
\resizebox{0.91\linewidth}{!}{%
$
\begin{aligned}
\bm{A} &= 
\begin{bmatrix}
1 - \frac{k_p}{m} & 0 & 0 \\
0 & 1 - \frac{k_p}{m} & 0 \\
0 & 0 & -COR
\end{bmatrix}, 
\quad
\bm{B} = \frac{k_p}{m}
\begin{bmatrix}
0 & r & 0 \\
-r & 0 & 0 \\
0 & 0 & 0
\end{bmatrix}, \\[1em]
\bm{C} &= \frac{k_p}{I}
\begin{bmatrix}
0 & -r & 0 \\
r & 0 & 0 \\
0 & 0 & 0
\end{bmatrix}, 
\quad
\bm{D} = 
\begin{bmatrix}
1 - \frac{k_p}{I} r^2 & 0 & 0 \\
0 & 1 - \frac{k_p}{I} r^2 & 0 \\
0 & 0 & 1
\end{bmatrix}
\end{aligned}
$
}
\label{eq:elastic_bounce_mat}
\end{equation}
where $I= 2/3r^2m$ is the inertia of a hollow sphere, \(m=2.7\mathrm{g}\) is the ball mass, \(r=0.02\mathrm{m}\) is the ball radius, \(COR=0.75\) is the Coefficient Of Restitution of the racket, and \(k_p=0.002\) is the friction coefficient of the racket.

\noindent The racket state is represented by its position $\bm{p}_{\mathrm{r}}$, world-frame 
orientation $R_{\mathrm{r}}^{\mathrm{w}}$, and linear velocity $\bm{V}_{\mathrm{r}}^{\mathrm{w}}$. 
Transforming ball velocities from the world frame to the racket frame yields
\begin{equation}
\left\{
\begin{aligned}
    \bm{v}_{\mathrm{r}} &= 
        \left(R_{\mathrm{r}}^{\mathrm{w}}\right)^{\!\top}
        \left( \bm{v}_{\mathrm{b}}^{\mathrm{w}} 
              - \bm{V}_{\mathrm{r}}^{\mathrm{w}} \right), \\
    \bm{\omega}_{\mathrm{r}} &= 
        \left(R_{\mathrm{r}}^{\mathrm{w}}\right)^{\!\top}
        \bm{\omega}_{\mathrm{b}}^{\mathrm{w}} ,
\end{aligned}
\right.
\end{equation}
In contrast to \scite{liu2012racket}, who estimate racket parameters using a simplified flight model that neglects the Magnus effect and the vertical drag component, we solve the boundary-value problem using the full aerodynamic flight dynamics.
We simulate the outgoing trajectory with a \ac{RK4} integrator, combining the racket–ball interaction model with the aerodynamic ODEs.\\[0.66ex]
\noindent We formulate the reconstruction of racket parameters as a nonlinear \ac{OCP} with single shooting:
\begin{align}
\min_{t_{\mathrm{net}},\,
      \bm{q}_{\mathrm{r}},\,\bm{V}_{\mathrm{r}}^{\mathrm{w}}}
\quad
& \alpha \big\| \bm{\omega}_{\mathrm{b}}^{+} - \bm{\omega}_{\mathrm{tgt}} \big\|^2
 + \beta \big\| \bm{p}_{\mathrm{b}}(t_{\mathrm{flight}}) - \bm{p}_{\mathrm{tgt}} \big\|^2
\nonumber \\
\text{s.t.}\qquad
& \bm{e}_z^{\!\top}\bm{p}_{\mathrm{b}}(t_{\mathrm{flight}}) = 0 , \label{eq:landing_constraint}\\
& \|\bm{q}_{\mathrm{r}}\|^2 = 1 , \label{eq:racket_ball_aligned_constraint}\\
& \left(R(\bm{q}_{\mathrm{r}})\bm{e}_z\right)^{\!\top} 
       \bm{V}_{\mathrm{r}}^{\mathrm{w}} \ge 0 , \nonumber \\
& \bm{e}_y^{\!\top} R(\bm{q}_{\mathrm{r}})\bm{e}_z \ge 0 . \label{eq:racket_front_constraint}
\end{align}%
Here, $t_{\mathrm{net}}$ is the net-crossing time, $t_{\mathrm{flight}}$ the bounce time on the opponent’s side, $\bm{q}_{\mathrm{r}}$ the racket orientation (unit quaternion), $\bm{\omega}_{\mathrm{b}}^{+}$ the outgoing spin after impact, $\bm{\omega}_{\mathrm{tgt}}$ the target spin, and $\bm{p}_{\mathrm{tgt}}$ the desired bounce location.
The weights $\alpha$ and $\beta$ balance spin accuracy and landing precision, and $\varepsilon$ enforces a safety clearance above the net.\\[0.66ex]
\noindent The resulting nonlinear OCP is solved using CasADi~\scite{andersson2019casadi} together with IPOPT~\scite{wachter2006on}, using a single-shooting transcription of the dynamics.
We run it with a fixed number of nodes and enforce that the last node be the bouncing position with \Cref{eq:landing_constraint}.
We also constrain the racket to always be facing the table with \Cref{eq:racket_front_constraint} and the racket's norm to always be in the same direction as the ball's outgoing velocity \Cref{eq:racket_ball_aligned_constraint}.
\noindent We validate our solver with a Monte-Carlo evaluation in which we randomly sample hit positions, bounce locations, incoming velocities, spins, and flight times over realistic ranges.
Across $10{,}000$ trials, the method converges to an exact solution (sub-millimeter bounce error) in \textbf{97.22\%} of cases.

\section{Filtering Details}
\label{sec:filtering-details}
\begin{table}[h]
\centering
\caption{Quality criteria used across reconstruction and filtering stages.}
\label{tab:quality_thresholds}
\resizebox{0.8\columnwidth}{!}{%
\begin{tabular}{l l}
\toprule
\textbf{Stage} & \textbf{Criterion / Threshold} \\
\midrule


\textbf{2D Reprojection Check} & \\
\hspace{1.75em}-- Max normalized error
& $0.2$ \\[0.5ex]

\textbf{3D-Domain Filtering} & \\
\hspace{1.75em}-- Max ODE fit RMSE & $0.3$ m \\[0.5ex]
\hspace{1.75em}-- Human average transl bounds: \\
\hspace{3.5em} $|\bar{h_x}|$ & $\in [0.5,\; 8.22]\,\text{m}$ \\
\hspace{3.5em} $|\bar{h_y}| $ & $\in [0.5,\; 1.525]\,\text{m}$ \\[0.75ex]

\textbf{Hit-Point Identification} & \\
\hspace{1.75em}-- Min peak-to-peak time & $0.2$ s \\
\hspace{1.75em}-- Min extrema $x$-position & $0.3$ m \\
\hspace{1.75em}-- Min number of hits & $2$ \\[0.5ex]

\bottomrule
\end{tabular}
}
\end{table}
We discuss the filtering of our dataset in Section \ref{subsec:filtering} of the main paper. 
We provide the individual threshold used in each filtering step in Table \ref{tab:quality_thresholds}.

\subsection{Filtering Bias}
There is no obvious pattern to the \textit{type} of rejected clips at the ``Gameplay clip" stage. Among the filtered clips, $204,720$ are singles, $86,247$ are doubles clips, and the distribution of the camera view is no different from the curated dataset (see Figures \ref{fig:focals}, \ref{fig:cameraposes}, \ref{fig:camera_pose_examples}). 

\noindent One common reason for clip rejection stems from the previous ``Scoreboard clips" stage: the scoreboard updates are not always a reliable indicator of starts and ends of gameplay (i.e. the scoreboard may be updated during the next point).
Our heuristic in the ``Gameplay clip" stage is rather conservative, focusing on reducing false positives. 

\begin{figure}
    \centering
    \includegraphics[width=\linewidth]{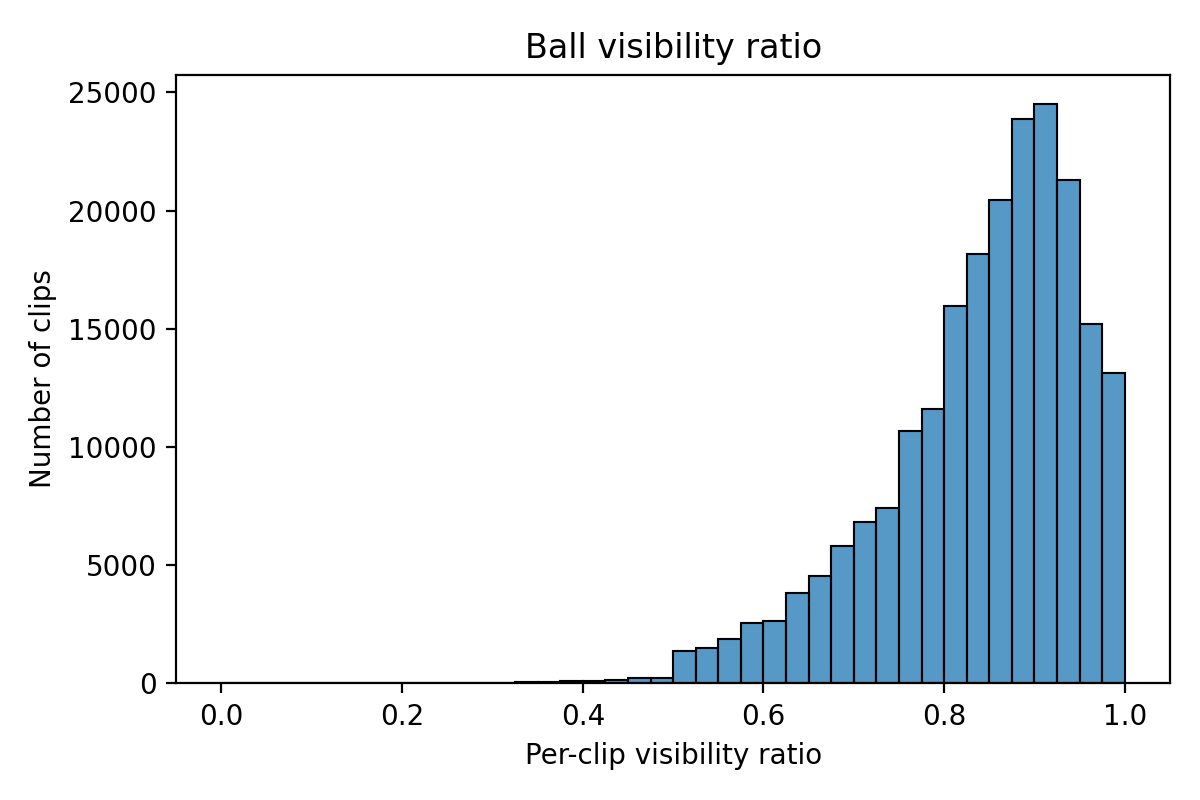}
    \caption{Distribution of the fraction of frames where the ball is visible.}
    \label{fig:visibility_ratio}
\end{figure}

\noindent Importantly, the final dataset is \textbf{not} skewed toward clean trajectories. 
In Figure 8, the inter-hit ball times indicate highly varied gameplay speeds, and the dataset also includes clips with high occlusion rates (\Cref{fig:visibility_ratio}). \\[0.33ex]

\section{Additional Dataset Statistics}
We provide further insights into the \textbf{TT4D} dataset by analyzing the distribution of the estimated camera parameters and the spatial statistics of the trajectories generated by our Flow Matching model.
\begin{figure}
    \centering
    \includegraphics[width=1\linewidth]{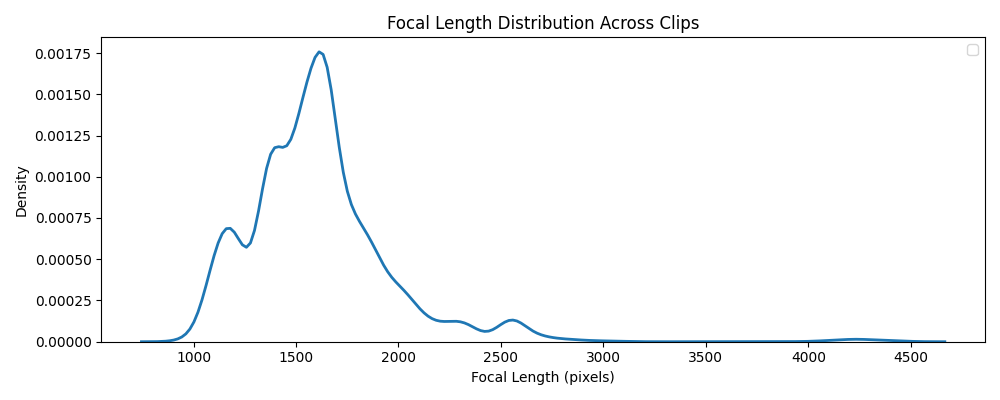}
    \Description{A density plot showing the distribution of camera focal lengths. The horizontal axis represents focal length in millimeters, ranging from under 1000 to over 4500. The vertical axis represents density. The curve is right-skewed with a primary peak around 1600 mm, a secondary smaller peak near 1200 mm, and a long tail extending to the right.}
    \caption{Distribution of camera focal lengths in the TT4D dataset.}
    \label{fig:focals}
\end{figure}
\begin{figure}
    \centering
    \includegraphics[width=1\linewidth]{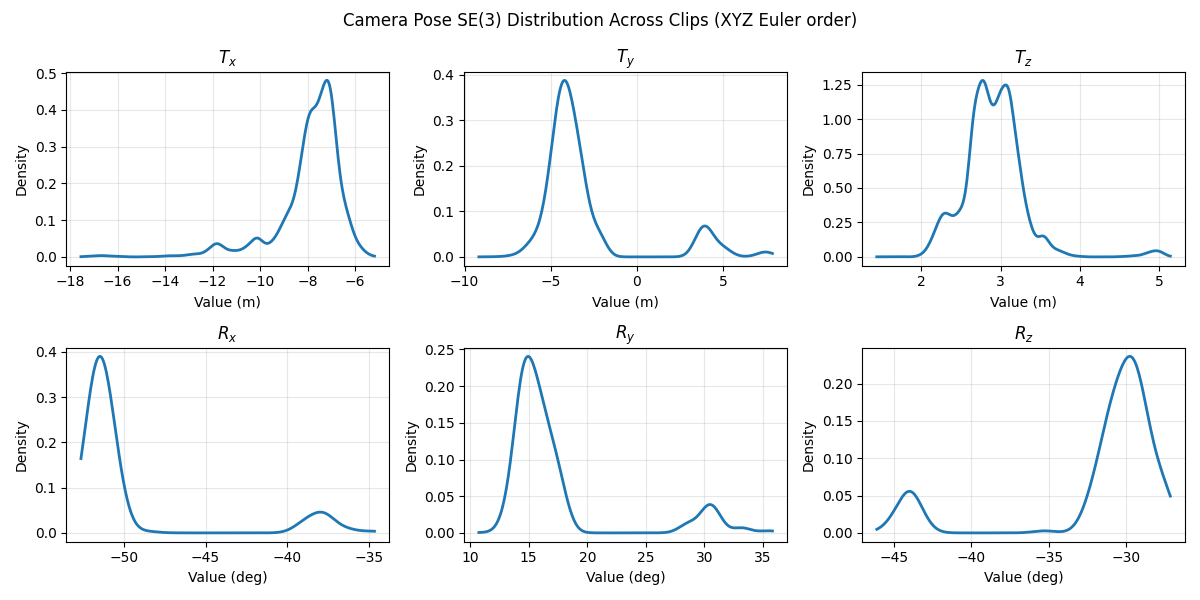}
    \Description{A 2 by 3 grid of density plots illustrating the distribution of camera poses. The top row plots translation parameters ($T_x$, $T_y$, $T_z$) in meters, while the bottom row plots rotation parameters ($R_x$, $R_y$, $R_z$) in degrees. Most of the individual subplots display a clear bimodal distribution featuring one prominent major peak and one significantly smaller secondary peak.}
    \caption{The distribution of camera poses $SE(3)$ in the TT4D dataset reveals two most common camera configurations, where one is roughly 5x more common than the other. }
    \label{fig:cameraposes}
\end{figure}
\begin{figure}[t]
    \centering
    \begin{subfigure}{0.70\linewidth}
        \centering
        \includegraphics[width=\linewidth]{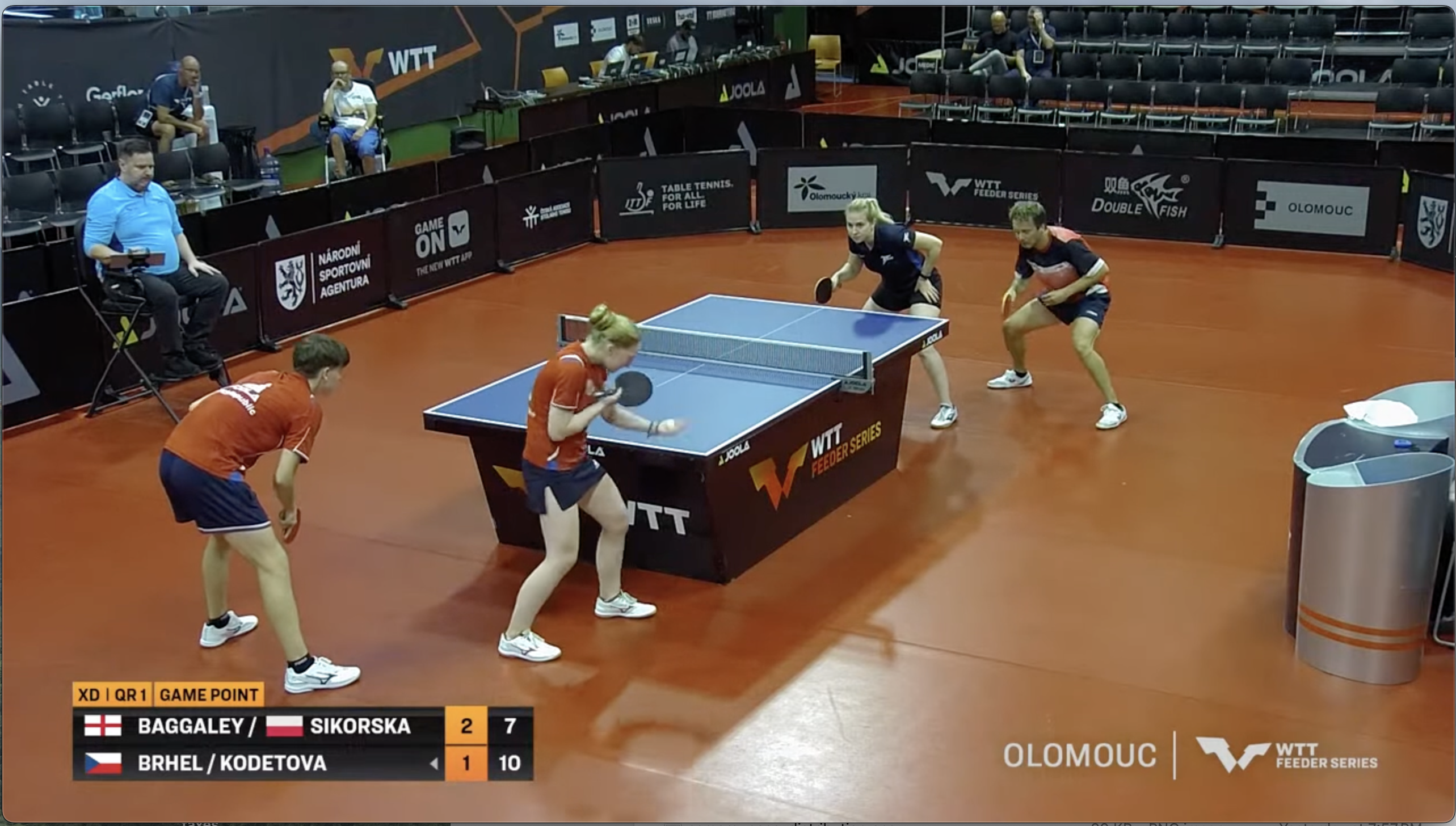}
        \label{fig:camera_pose_examples_1}
    \end{subfigure}
    \vspace{-0.2cm}
    \begin{subfigure}{0.70\linewidth}
        \centering
        \includegraphics[width=\linewidth]{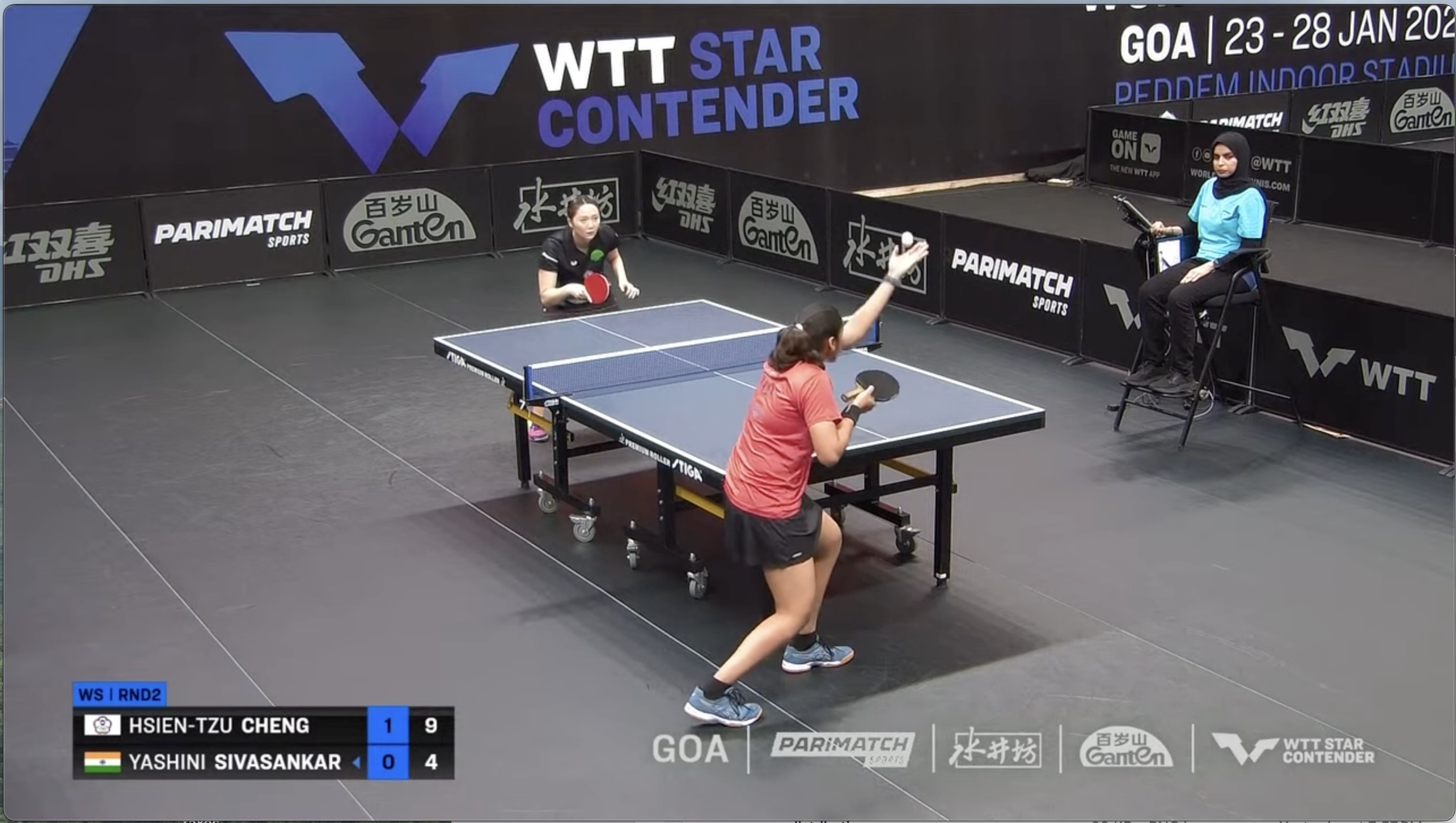}
        \label{fig:camera_pose_examples_2}
    \end{subfigure}
    \vspace{-0.3cm}
    \Description{Two photographs of table tennis matches demonstrating different camera angles. The top image shows a doubles match viewed from a higher, downward-angled perspective, corresponding to Euler angles of -55, 15, and -30 degrees. Four players are visible around a blue table on a red floor. The bottom image shows a singles match from a slightly lower, more offset perspective, corresponding to Euler angles of -37, 32, and -42 degrees. Two players and a seated umpire are visible around a blue table on a gray floor.}
    \caption{
    Two common camera poses in the TT4D dataset. The top and bottom figures corresponds to Euler angle modes of $(-55 \degree, 15\degree, -30\degree)$ and $(-37\degree, 32\degree, -42\degree)$, respectively.
    }
    \vspace{-0.2cm}
    \label{fig:camera_pose_examples}
\end{figure}
\noindent Figure \ref{fig:focals} shows a diverse distribution of different focal lengths in our dataset.
The distribution of the extrinsic camera parameters (translation $\mathbf{T}$ and rotation $\bm{\theta}$ in Euler angles) is visualized in Figure \ref{fig:cameraposes}.
While a wide variety of poses is present in the dataset, a bimodal distribution is visible in some parameters, indicating that some camera poses are especially frequent in the dataset.
We show two dominant views in Figure \ref{fig:camera_pose_examples}

\section{Generative Model Details}
\label{sec:supp_generative_model}

\subsection{Data and Representation}
Let $\mathcal{D}$ denote our dataset of reconstructed trajectories.
Each trajectory $\tau = (o_1,\ldots,o_N) = o_{1:N} \in \mathcal{D}$ consists of observations $o_t = (b_t, h_{1,t}, h_{2,t})$, where the ball state is $b_t \in \mathbb{R}^3$ and the skeletons $h_{1,t}, h_{2,t} \in \mathbb{R}^{21 \times 3}$.
All trajectories are resampled to 30\,Hz.
In total, the dataset contains 237{,}054 reconstructed points ($\sim$\,151 hours).

\subsection{Model Architecture and Training}
We model the time-dependent vector field $v_\theta(\tau_t, t \mid c)$ using a DiT-style architecture with 6 attention heads, 6 layers, MLP ratio of 4, dropout of 0.1, model embedding size of 384, conditioning embedding size of 512, and time embedding size of 128.
We predict a horizon of 20 observations $\tau = o_{1:20}$ from an observation history of $c = o_{-9:0}$ ~\scite{tong2023simulation, tong2024improving}. 
We train this model for 600{,}000 iterations with a batch size of 512 on a single NVIDIA RTX~4090 GPU.
Optimization uses AdamW~\scite{loshchilov2017decoupled} with a learning rate of $2\times 10^{-4}$, a cosine decay schedule, and 500 warmup steps. 

\subsection{Generation Quality}
The model weights used at inference are computed by passing an exponential moving average (EMA) filter over the sequence of model weights from training .
During evaluation, we generate future trajectories by numerically integrating the learned ODE forward from Gaussian noise using $n_{\text{sample}}=5$ uniform steps.\\[0.66ex]
\noindent We assess two categories of metrics: \\
\noindent\textbf{Physical Plausibility:}
We run the generated trajectories through the same \textbf{Physics-Based ODE Fit} filter from our pipeline's Stage 4 (\Cref{subsec:filtering}).
As shown in \Cref{fig:flow_matching_eval} in the main paper, the distribution of this error closely matches the real data.
We also measure the smoothness of ball accelerations, continuity of human joint velocities, and any violations of kinematic limits across the full predicted rallies. \\
\noindent\textbf{Gameplay Realism and Diversity:}
We compute distributional statistics over sequence duration, inter-hit timing, ball height profiles, and stroke kinematics.
We compare these distributions from the generated rallies to those from a held-out test set of real data.
Qualitative visualizations of sampled rallies (see \Cref{fig:supp_qualitative_flow}) further demonstrate that the learned flow produces coherent long-horizon behavior, including consistent hitting mechanics and realistic ball-table interactions.

\section{Evaluation Metrics}
\label{subsec:supp_metrics}
We use several metrics to evaluate our \textbf{Lifting Network}, which are defined below.
Let $M$ be the total number of trajectories in the test set, and $N_m$ be the number of frames with valid 2D ball detections in the $m$-th trajectory.
Let $\vec{r}(t_n)$ and $\vec{\omega}(t_n)$ be the ground truth 3D position and 3D spin at timestep $t_n$, and let $\hat{\vec{r}}(t_n)$ and $\hat{\vec{\omega}}(t_n)$ be the predicted values.

\subsection{3D Trajectory Error}
The $\Delta\vec{r}_{\text{3D}}$ metric computes the mean Euclidean distance between the predicted and ground truth 3D positions, measured in centimeters.
It is our primary metric for 3D accuracy on the synthetic and TT4DBench datasets.
\begin{equation} 
    \Delta\vec{r}_\text{3D} = \frac{1}{M} \sum_{m=0}^{M-1} \frac{1}{N_m} \sum_{n=0}^{N_m-1} || \vec{r}_{\text{3D}}(t_n) - \hat{\vec{r}}_{\text{3D}}(t_n) ||_2
\end{equation}

\subsection{3D Spin Error}
The $\Delta\vec{\omega}$ metric computes the mean Euclidean distance between the predicted and ground truth 3D spin vectors, measured in Hz.
It is only used on our synthetic dataset, as real-world datasets do not provide ground truth 3D spin vectors.
\begin{equation} 
    \Delta \vec{\omega} = \frac{1}{M} \sum_{m=0}^{M-1} \frac{1}{N_m} \sum_{n=0}^{N_m-1} || \vec{\omega}(t_n) - \hat{\vec{\omega}}(t_n) ||_2
\end{equation}

\subsection{2D Reprojection Error}
For real-world datasets like TTST that lack 3D ground truth, we evaluate the 2D reprojection error $\Delta\vec{r}_{\text{2D}}$.
The predicted 3D trajectory $\hat{\vec{r}}(t_n)$ is projected back into the 2D image plane using the provided camera projection matrix $\mathcal{P}$, and compared against the 2D ground truth annotations $\vec{r}_{\text{2D}}(t_n)$.
The error is measured in pixels.
\begin{equation}
    \resizebox{.89\linewidth}{!}{
    $
    \Delta\vec{r}_\text{2D} = \frac{1}{M} \sum_{m=0}^{M-1} \frac{1}{N_m} \sum_{n=0}^{N_m-1} || \mathcal{P}(\hat{\vec{r}}_{\text{3D}}(t_n)) - \vec{r}_{\text{2D}}(t_n) ||_2
    $
    }
\end{equation}

\subsection{Macro F1 Score}
For real-world datasets with binary spin labels (topspin/backspin), we compute the Macro F1 score.
To obtain a binary class from our network's continuous 3D spin prediction $\hat{\vec{\omega}}(t_0)$, we first transform the spin vector from the world coordinate system into the ball coordinate system defined in \scite{kienzle2025towards}.
In this local frame, the $\tilde{y}$-axis is orthogonal to the ball's velocity and parallel to the table plane, such that the spin component $\hat{\omega}_{\tilde{y}}$ directly corresponds to the topspin/backspin magnitude.
We classify the segment as \textbf{topspin} if $\hat{\omega}_{\tilde{y}} > 0$ and as \textbf{backspin} if $\hat{\omega}_{\tilde{y}} \leq 0$. \\[0.66ex]
\noindent We then calculate the F1 score for each class $c \in \{\text{Topspin}, \text{Backspin}\}$:
\begin{equation}
    \text{F1}_c = \frac{2 \cdot \text{TP}_c}{2 \cdot \text{TP}_c + \text{FP}_c + \text{FN}_c}
\end{equation}
where $\text{TP}_c, \text{FP}_c, \text{FN}_c$ are the true positives, false positives, and false negatives for class $c$.
The final Macro F1 score is the unweighted mean:
\begin{equation}
    \text{Macro F1} = \frac{\text{F1}_{\text{Topspin}} + \text{F1}_{\text{Backspin}}}{2}
\end{equation}

\section{Additional Experiments and Visualizations}
\subsection{Lifting Network Training}
The network, consisting of $1.6$ million parameters, is trained solely on the train split of our synthetic dataset ($2.6$ million rallies). 
It is trained for $17$ epochs on a single NVIDIA H100 GPU and we use the ADAM optimizer \cite{diederik2015adam} with a learning rate of $10^{-4}$. 
We track an exponential moving average of the model weights \cite{tarvainen2017meanteachers} and select the best model based on its validation performance on the TTST dataset~\cite{kienzle2026uplifting}. \\[0.66ex]

\subsection{Lifting Network Inference Speed}
\label{subsec:lifting-inference-speed}
Unlike optimization-based approaches~\scite{etaat2025latte,gossard2025tt3d,liu2022monotrack}, which require a computationally expensive fitting process for every individual segment of a point, learning-based methods~\scite{kienzle2025towards,kienzle2026uplifting} perform lifting via a single, efficient forward pass.
While the initial training cost is non-negligible (approximately 2 days on a single NVIDIA H100 GPU), the resulting model offers exceptional efficiency during inference.
Our proposed network further enhances this efficiency by processing entire points consisting of multiple segments in a single forward pass, rather than lifting each segment individually. \\[0.66ex]
\noindent We evaluate the inference speed of our Lifting Network across three generations of GPU hardware: NVIDIA H100, V100, and Titan X (Pascal).
We distinguish between two operating regimes: an \textit{online} mode (batch size 1), representing latency-critical real-time applications, and an \textit{offline} mode (batch size 128), representing high-throughput dataset generation.
The results are summarized in \Cref{tab:efficiency}.
\begin{table}[ht]
\centering
\caption{Inference performance of the Lifting Network. The network's high throughput (measured in Rallies per Second) allows for extremely efficient large-scale dataset processing. \ding{51}/\ding{53} indicate the presence/absence of batching (batch size 128).}
\label{tab:efficiency}
\resizebox{0.95\columnwidth}{!}{%
\begin{tabular}{@{}cccc@{}}
\toprule
\textbf{Batched} & \textbf{Inference Speed (Rallies/s)} & \textbf{GPU VRAM} & \textbf{GPU} \\ \midrule
\ding{53}        & 120                                 & 650 MB            & \multirow{2}{*}{H100} \\
\ding{51}        & 3011                                & 1070 MB            &           \\             
\midrule
\ding{53}        & 75                                  & 410 MB            & \multirow{2}{*}{V100} \\
\ding{51}        & 1310                                & 810 MB            &               \\
\midrule
\ding{53}        & 25                                  & 280 MB            & \multirow{2}{*}{\begin{tabular}[c]{@{}c@{}}Titan X \\ (Pascal)\end{tabular}} \\
\ding{51}        & 543                                 & 670 MB            &    \\
\bottomrule
\end{tabular}%
}
\end{table} \\[0.66ex]
\noindent The results demonstrate that our model achieves real-time performance even on legacy hardware.
On a Titan X (Pascal), the network processes 25 points per second in online mode.
Given that a typical table tennis point lasts several seconds, this inference speed is orders of magnitude faster than real-time, enabling low-latency applications such as live broadcasting analysis or robotic anticipation on consumer-grade hardware.
In the batched offline setting, the throughput scales dramatically, reaching over 3000 rallies per second on an H100.
This extreme efficiency was a critical enabler for the creation of the TT4D dataset, allowing us to lift hundreds of hours of gameplay in minutes, shifting the computational bottleneck entirely to the preliminary steps.

\subsection{Racket Strike Reconstruction Visualization}
\begin{figure}[ht]
    \centering
    \Description{Two 3D plots, an isometric view and a side view, illustrating a table tennis shot. A legend in the isometric view identifies a blue line as the ball trajectory, a green dot as the hit, a red dot as the target, and red, green, blue, and purple lines representing the racket X-axis, Y-axis, Z-axis, and velocity, respectively. Both plots show the blue ball trajectory arching from the hit point on one side of a green table plane to the target point on the other, with the racket orientation axes and velocity vectors originating from the hit point.}
    \begin{subfigure}{\linewidth}
        \centering
        \includegraphics[width=0.7\linewidth]{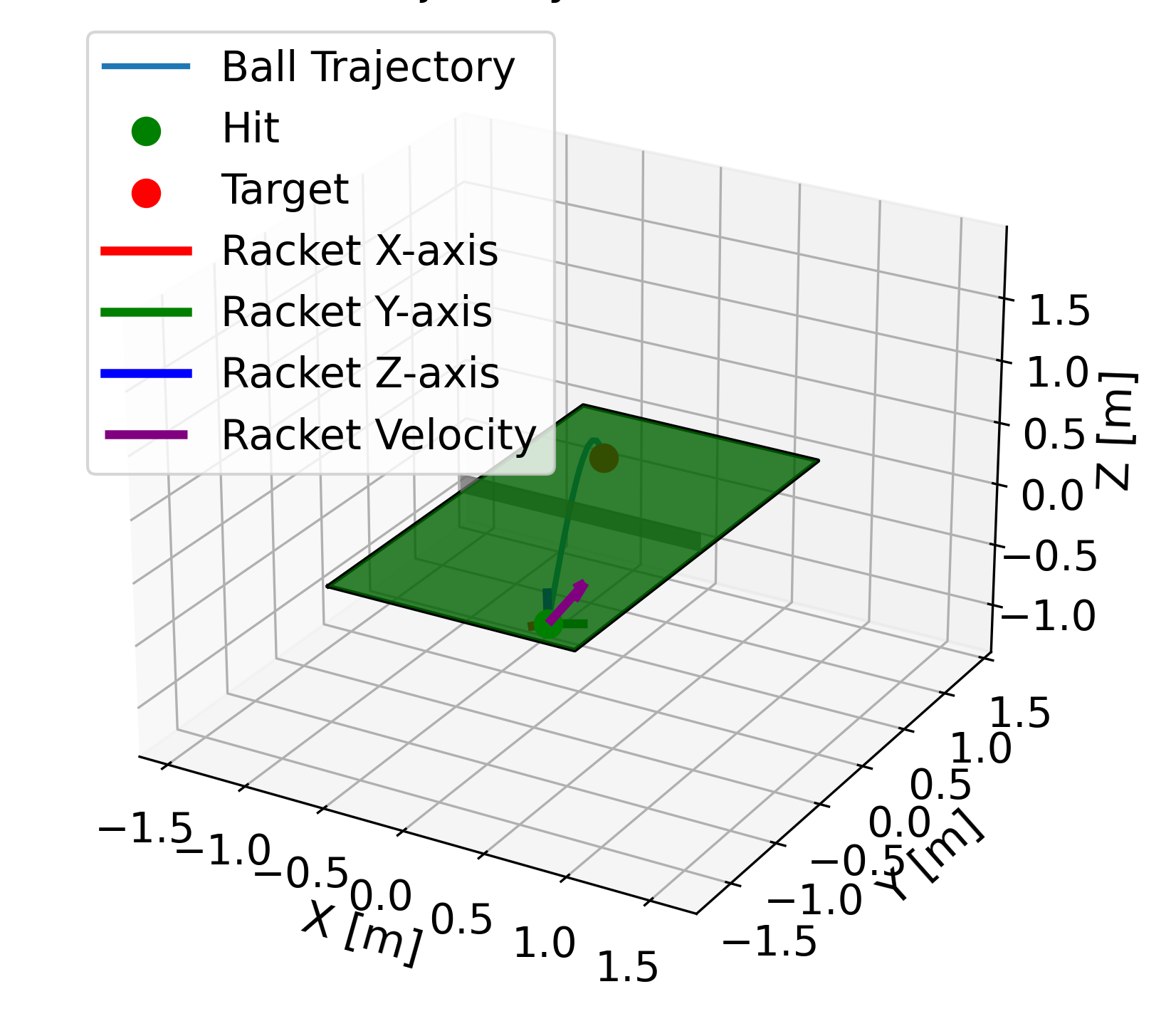}
        \caption{Isometric view}
    \end{subfigure}
    \begin{subfigure}{\linewidth}
        \centering
        \includegraphics[width=0.8\linewidth]{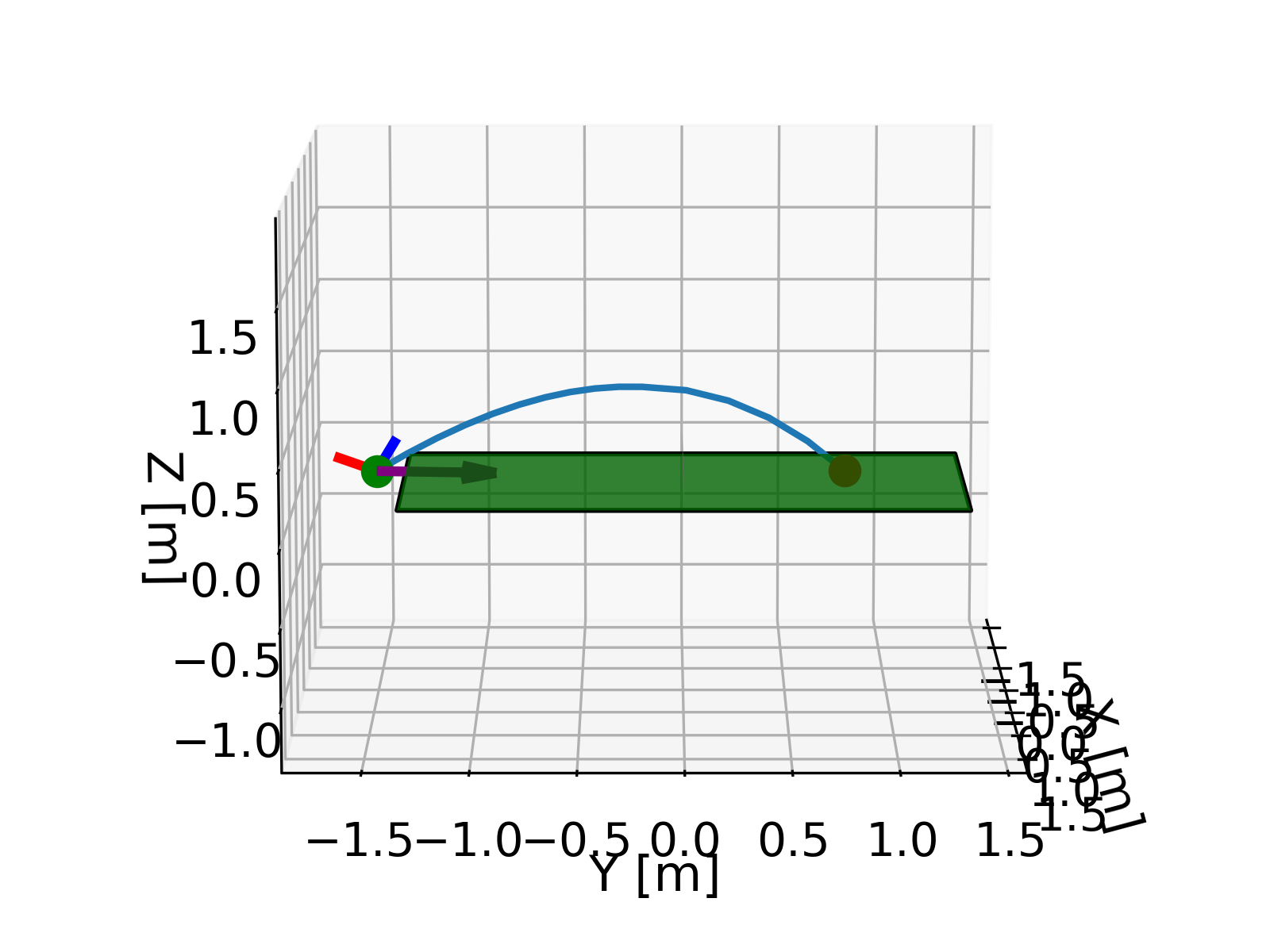}
        \caption{Side view}
    \end{subfigure}
    \caption{Example of the estimated racket orientation and velocity for a given ball time of flight, ball incoming velocity and spin and bounce target. This was solved using our \ac{OCP}.}
    \label{fig:racket_ocp_views}
\end{figure}
To further validate our racket reconstructions, we provide a qualitative visualization of the reconstruction in Figure \ref{fig:racket_ocp_views}.
The figure shows that the racket reconstructions look realistic.

\subsection{Evaluation of Generative Data}
\begin{figure}[t]
    \centering

    \begin{subfigure}{0.48\linewidth}
        \centering
        \includegraphics[width=\linewidth]{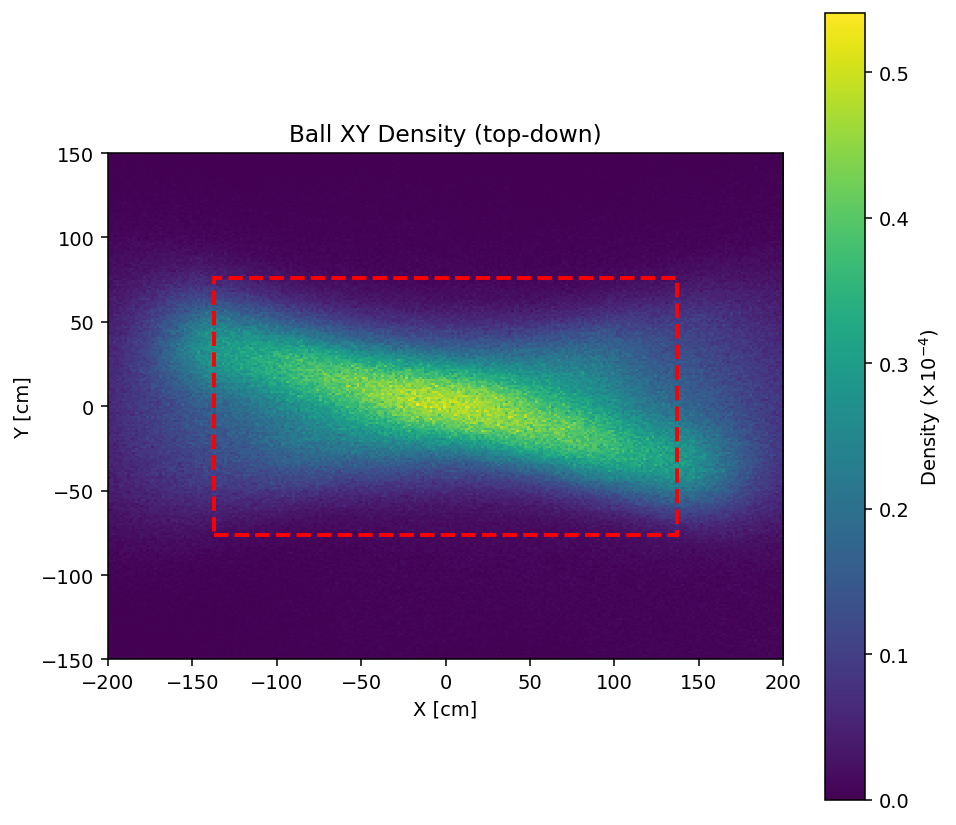}
        \label{fig:ball_xy_density_gen}
    \end{subfigure}
    \hfill
    \begin{subfigure}{0.48\linewidth}
        \centering
        \includegraphics[width=\linewidth]{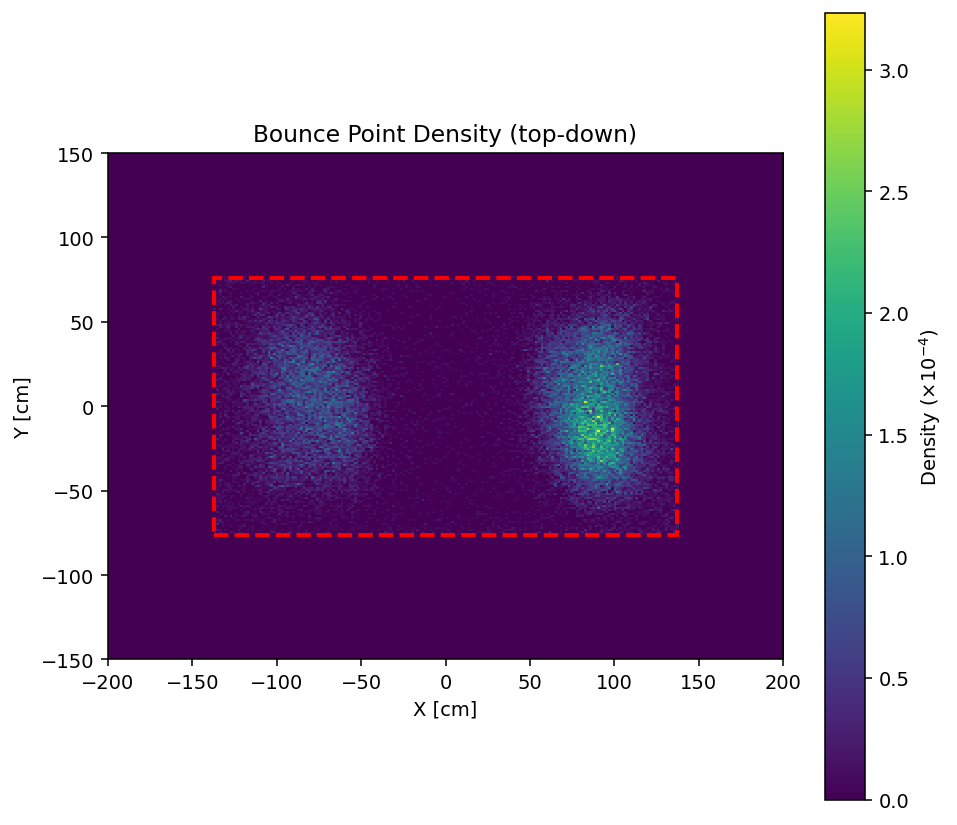}
        \label{fig:bounce_xy_density_gen}
    \end{subfigure}

    \vspace{-0.2cm}

    \begin{subfigure}{0.70\linewidth}
        \centering
        \includegraphics[width=\linewidth]{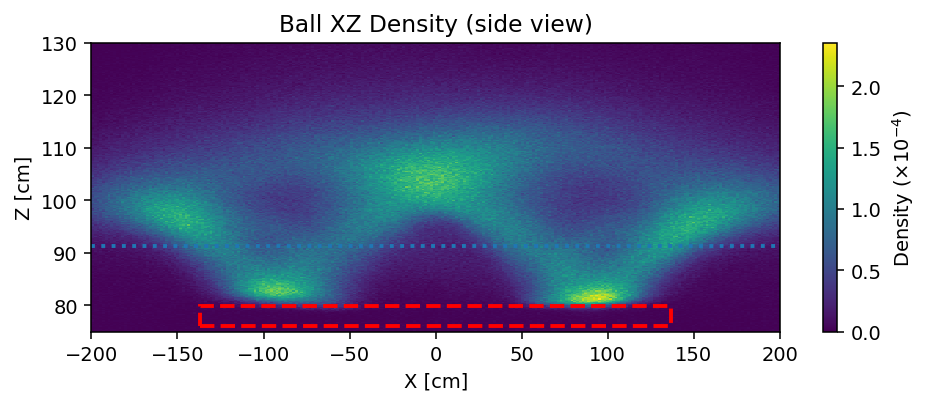}
        \label{fig:ball_xz_density_gen}
    \end{subfigure}
    \vspace{-0.3cm}
    \Description{Three heatmaps illustrating ball position and bounce densities for 10,000 generated samples. A dashed red rectangle marks the table boundaries in all plots. The top-left plot displays a top-down view of ball XY density, revealing a prominent diagonal band indicative of frequent cross-court trajectories. The top-right plot shows top-down bounce point density, featuring two distinct high-density clusters on opposite halves of the table. The bottom plot presents a side view of ball XZ density, showing the ball's trajectory arching low over a dotted blue line representing the net, with density concentrating near the table surface on both sides indicating bounces.}
    \caption{
    Ball position densities for the 10{,}000 generated samples. The table region is marked by the dashed red line, and the net's height is marked by the dotted blue line. 
    }
    \vspace{-0.2cm}
    \label{fig:ball_density_all_gen}
\end{figure}
We provide a qualitative example of a generated sequence in Figure \ref{fig:supp_qualitative_flow}. 
Moreover, we illustrate the distribution of the ball locations in Figure \ref{fig:ball_density_all_gen}.

\subsection{Humanoid Motion Tracking of Table Tennis Motions}
We validate the fidelity of our Player Reconstruction algorithm (\ref{sup:player-reconstruction}) by replaying the motion on a Unitree G1 robot. First, we retarget the SMPL motion to the Unitree G1 using GMR \cite{araujo2025retargeting}. Next, to facilitate smooth hardware deployment, we use the motion in-betweening capabilities of GEM \cite{genmo2025} to ease the start and end of each motion. Specifically, we generate half a second of motion to smoothly transition from an A-pose to the initial motion pose and from the final motion pose back to an A-pose. Finally, we train a motion tracking policy on this smoothed motion using BeyondMimic \cite{liao2025beyondmimic} for 30,000 iterations on an NVIDIA GeForce RTX 5090 GPU. One hardware deployment video along with the original motion is included in the supplementary zip file. 

\clearpage

\begin{figure*}[p]
    \centering
    \setlength{\tabcolsep}{2pt}

\begin{tabular}{ccc}
    \includegraphics[width=0.32\textwidth]{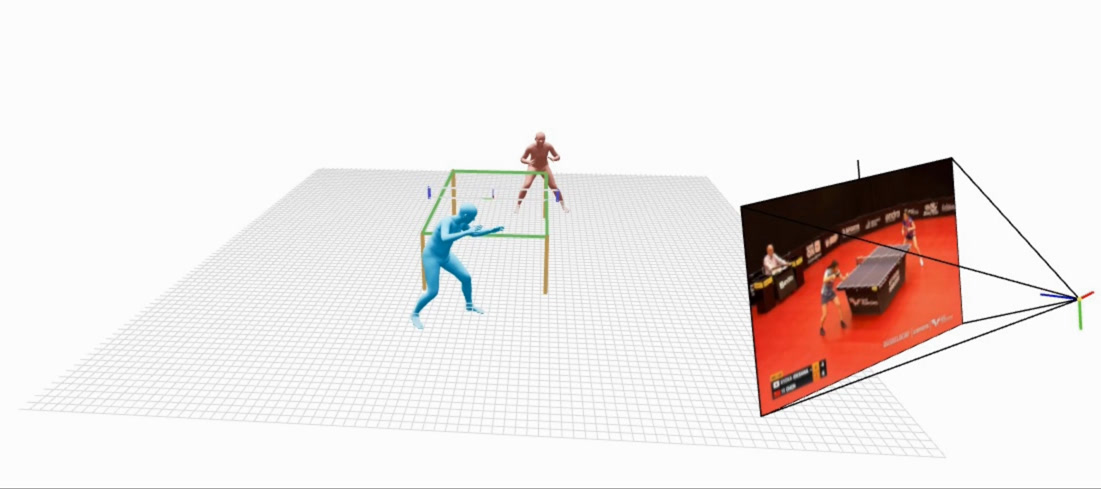} &
    \includegraphics[width=0.32\textwidth]{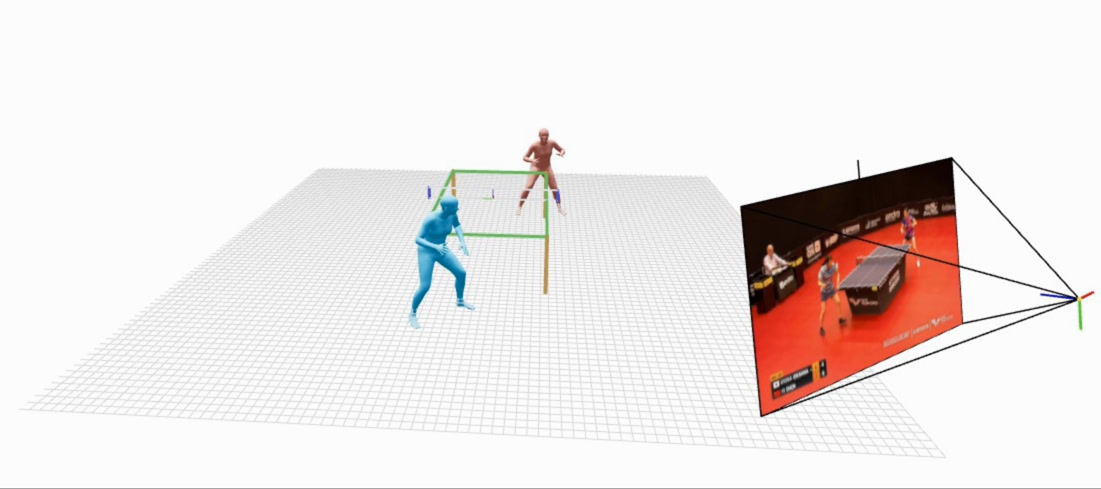} &
    \includegraphics[width=0.32\textwidth]{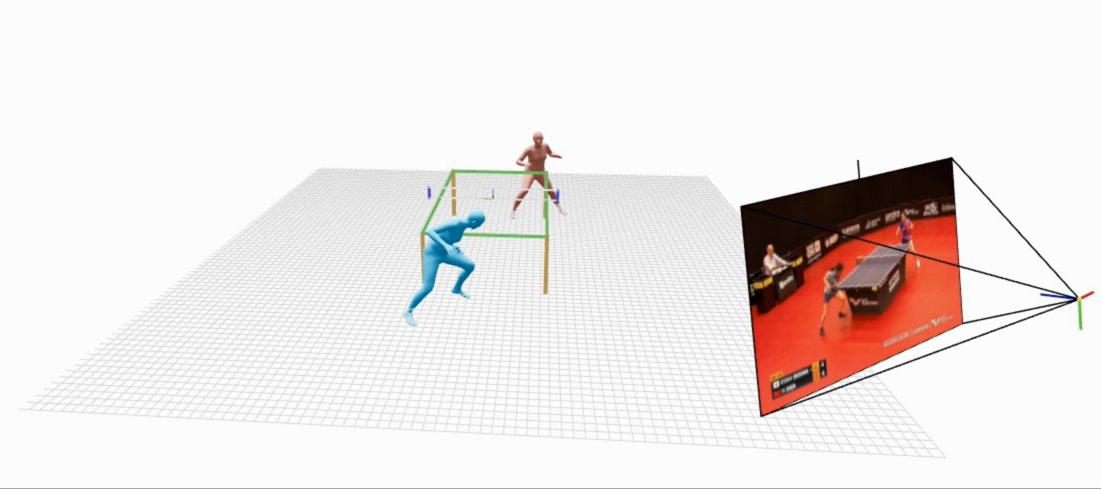} \\

    \includegraphics[width=0.32\textwidth]{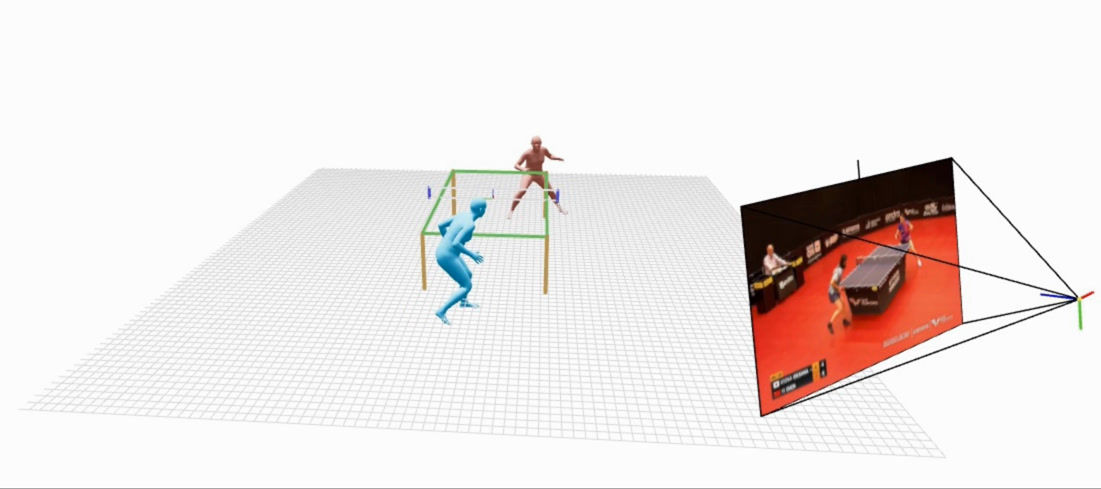} &
    \includegraphics[width=0.32\textwidth]{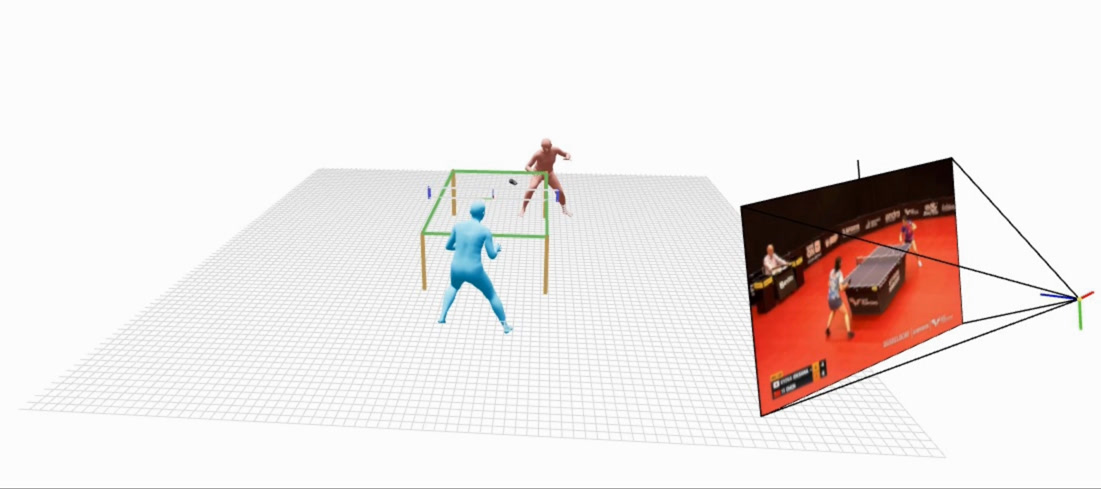} &
    \includegraphics[width=0.32\textwidth]{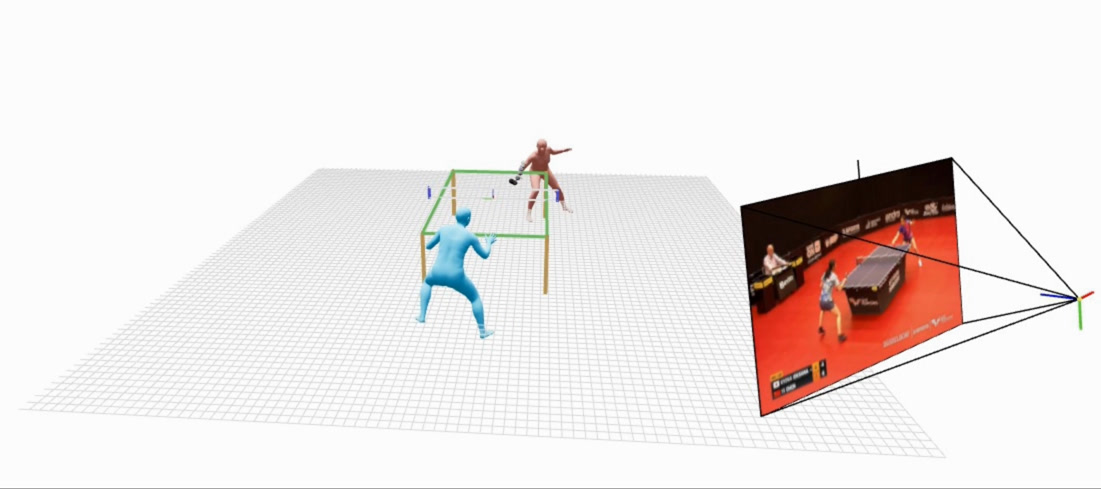} \\

    \includegraphics[width=0.32\textwidth]{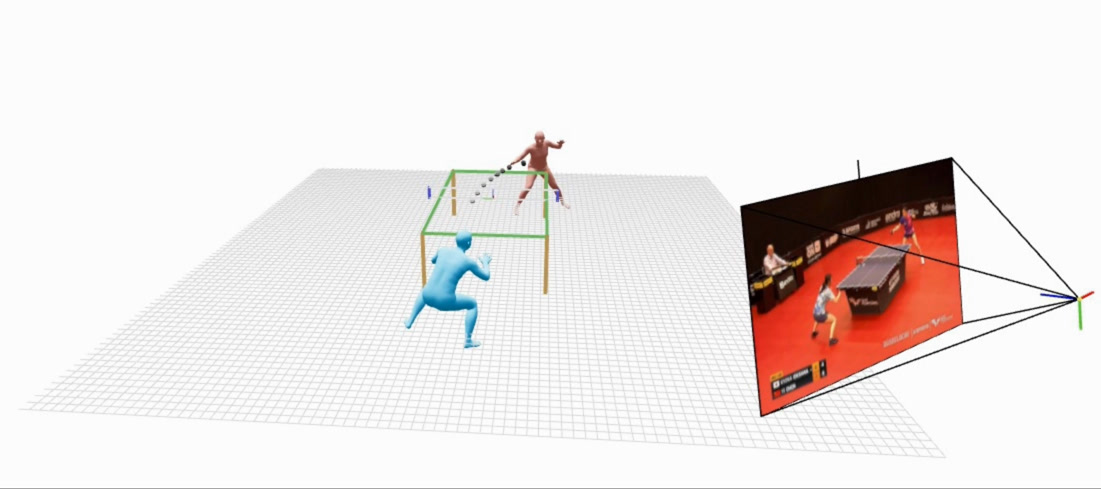} &
    \includegraphics[width=0.32\textwidth]{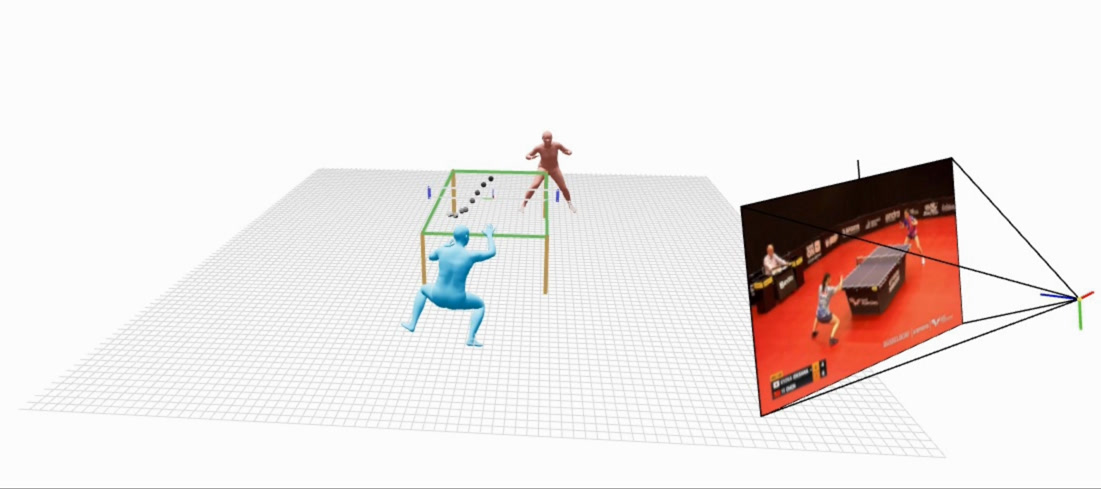} &
    \includegraphics[width=0.32\textwidth]{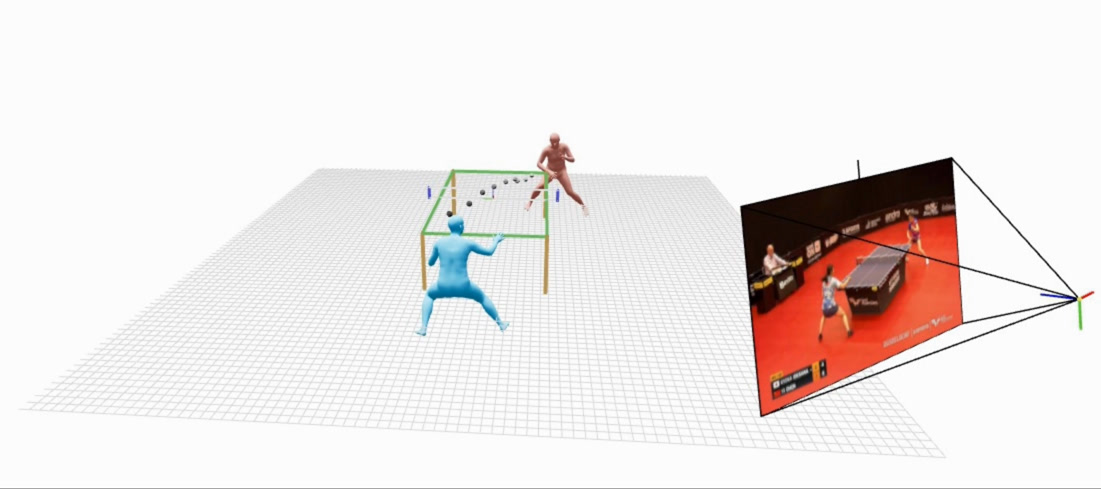} \\

    \includegraphics[width=0.32\textwidth]{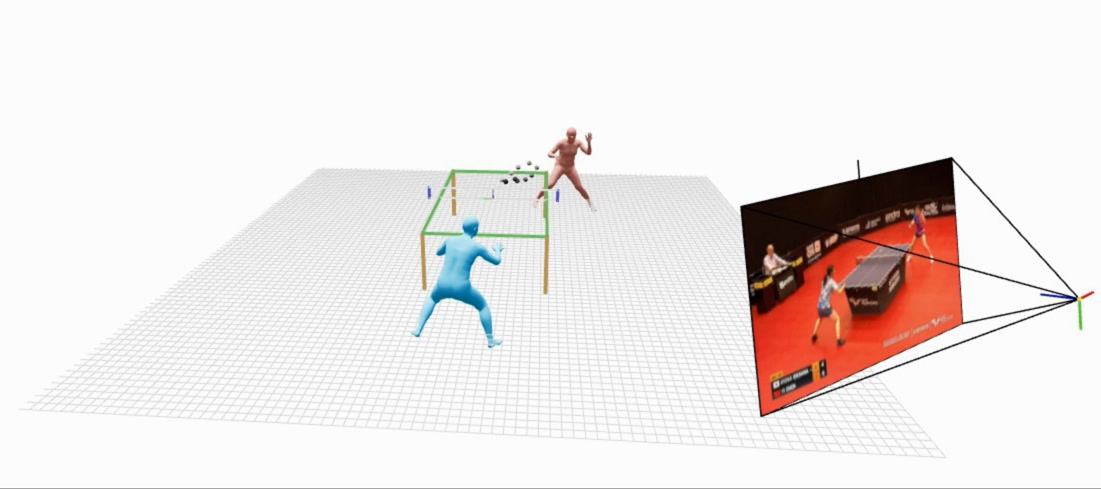} &
    \includegraphics[width=0.32\textwidth]{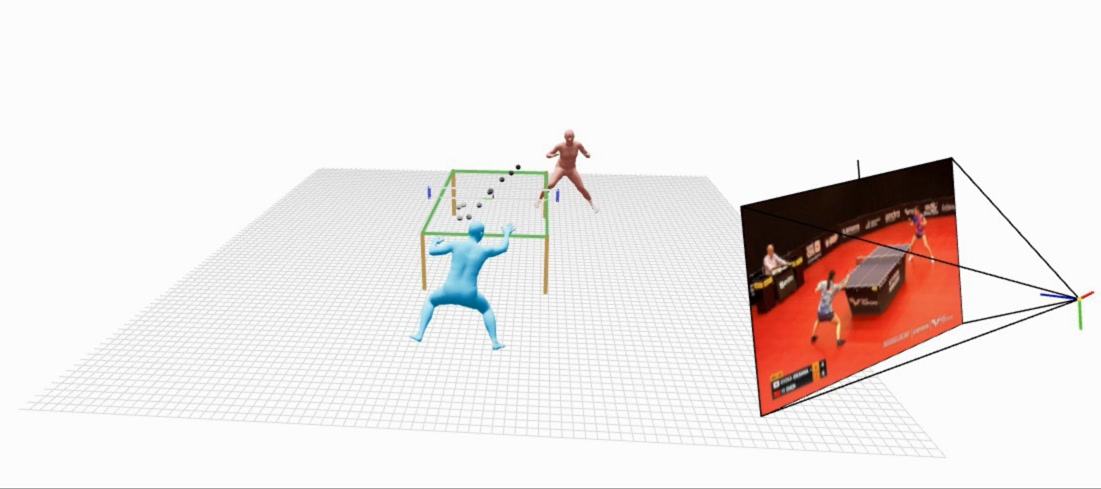} &
    \includegraphics[width=0.32\textwidth]{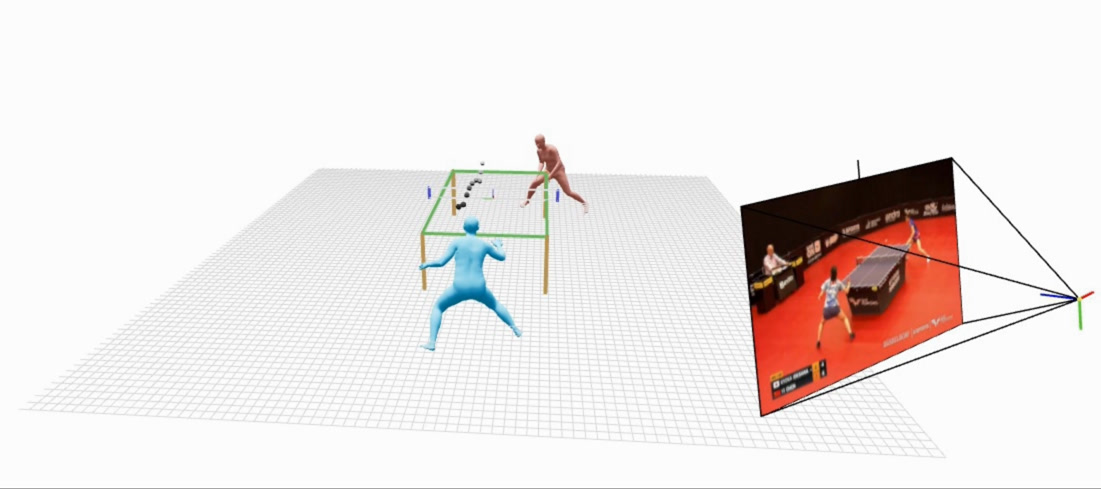} \\

    \includegraphics[width=0.32\textwidth]{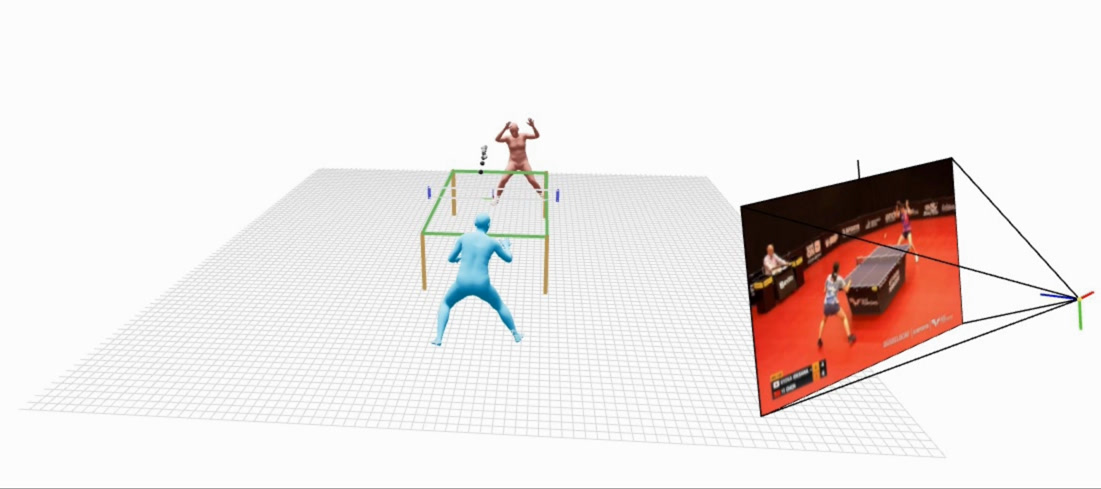} &
    \includegraphics[width=0.32\textwidth]{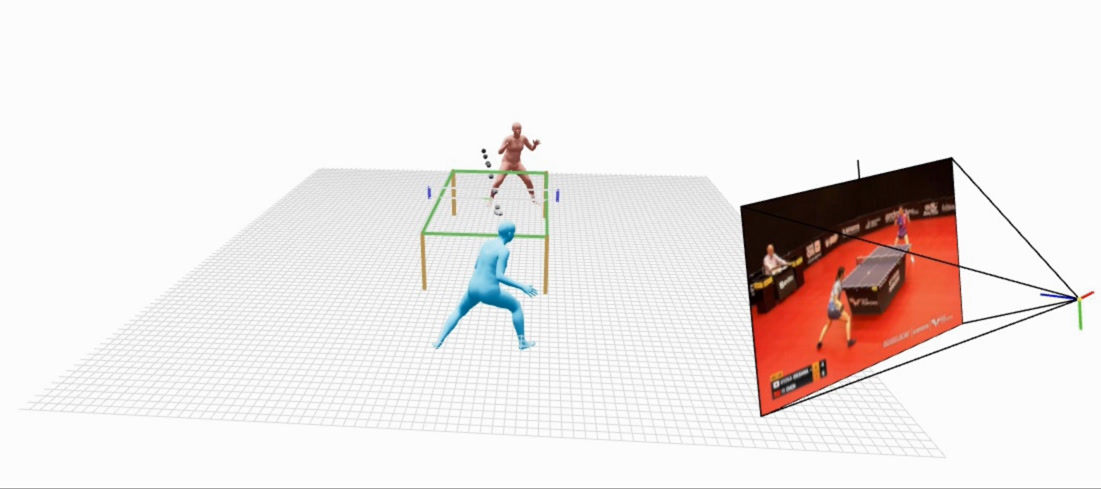} &
    \includegraphics[width=0.32\textwidth]{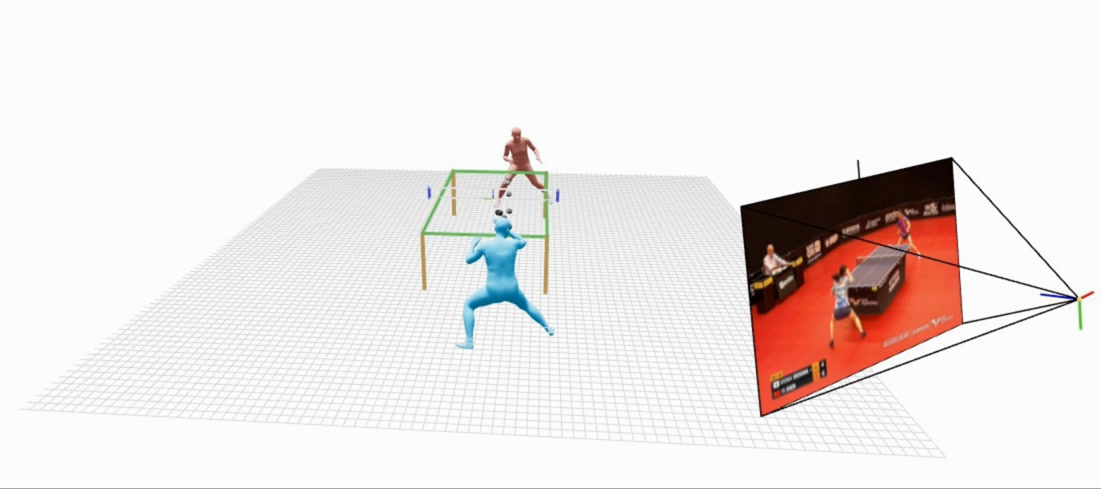} \\

    \includegraphics[width=0.32\textwidth]{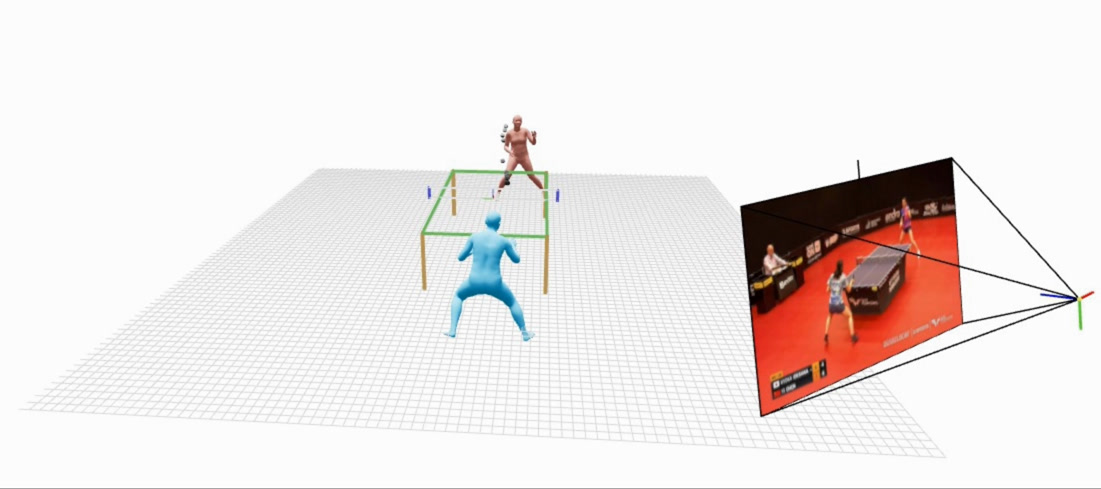} &
    \includegraphics[width=0.32\textwidth]{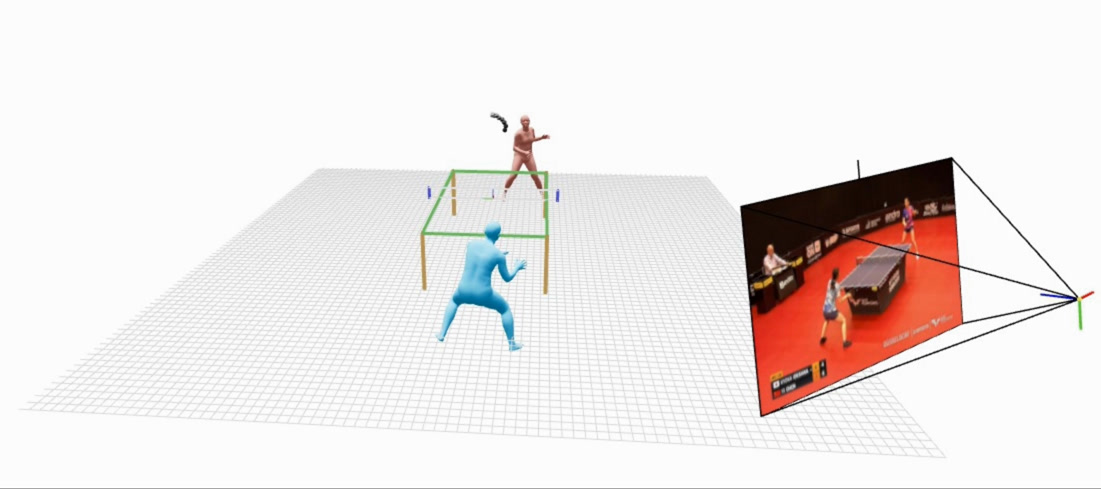} &
    \includegraphics[width=0.32\textwidth]{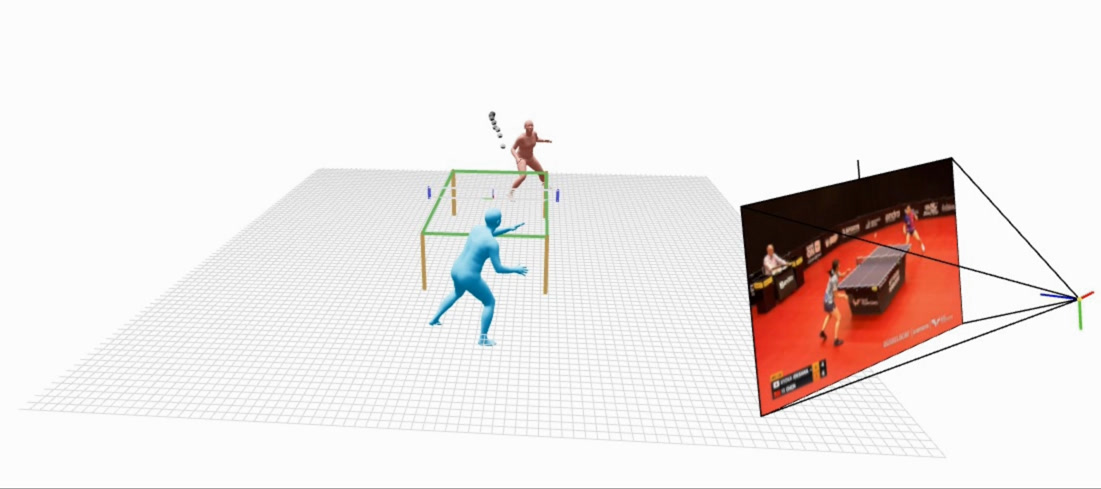} \\

    \includegraphics[width=0.32\textwidth]{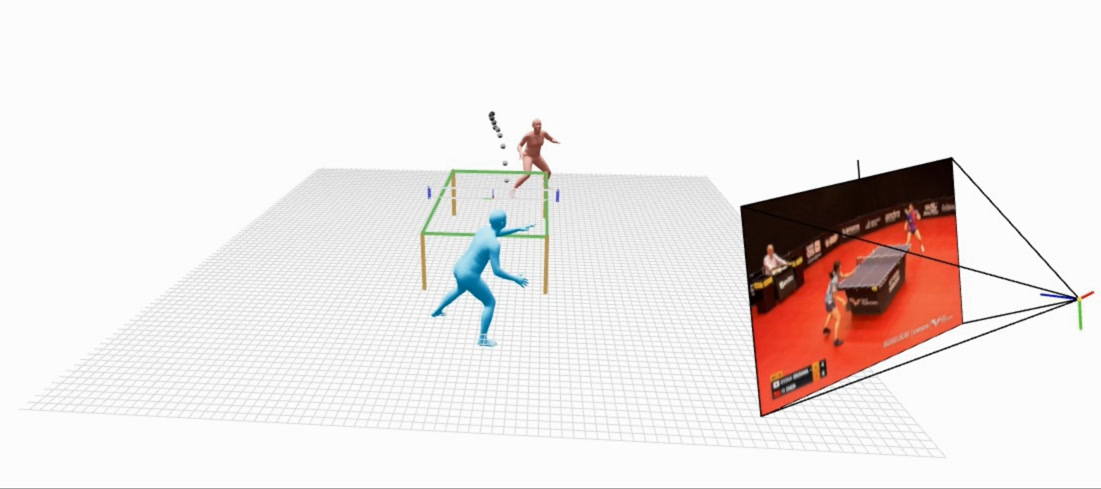} &
    \includegraphics[width=0.32\textwidth]{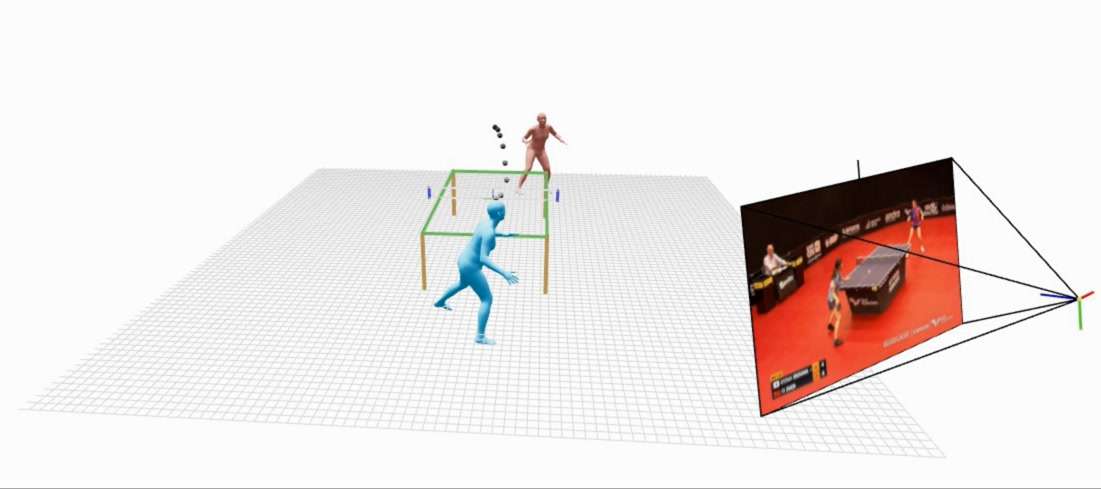} &
    \includegraphics[width=0.32\textwidth]{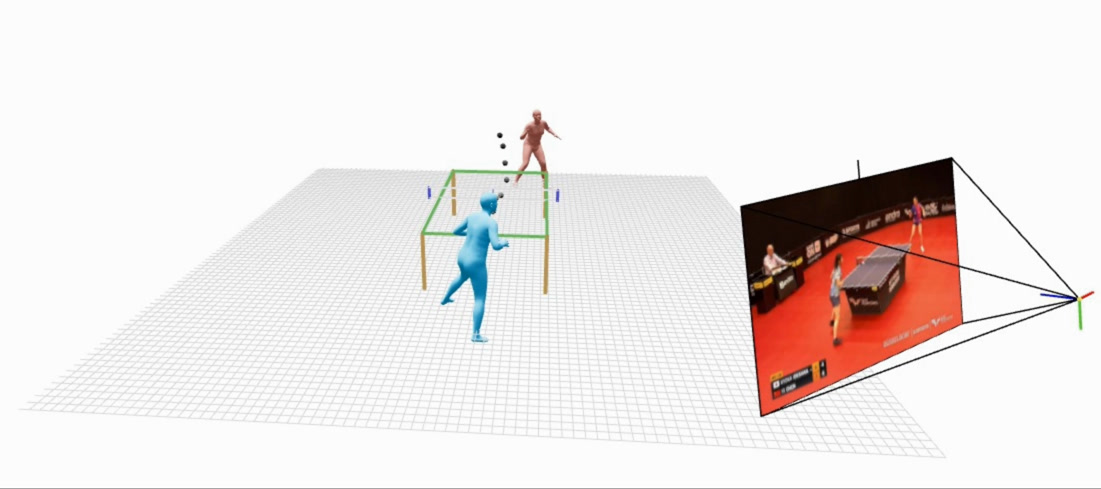} \\
\end{tabular}

    \Description{A 7 by 3 grid of frames showing a chronological sequence of a reconstructed table tennis rally. Each frame displays a 3D scene with two human meshes (one light blue in the foreground, one red in the background) playing at a green wireframe table, with the ball's trajectory tracked by a series of dots. On the right side of each frame, a 2D inset of the original broadcast video is projected from a camera frustum, illustrating the estimated camera viewpoint. The sequence progresses from left to right, top to bottom, showing the continuous player movements and ball exchanges.}
    \caption{One example of a \textbf{reconstructed sequence} from our TT4D dataset at 30 FPS. We display every 16th frame. The diagram should be read from left to right.}
    \label{fig:supp_qualitative_reconstruction_example}
\end{figure*}

\begin{figure*}[p]
    \centering
    \setlength{\tabcolsep}{2pt}

    \begin{tabular}{cccc}
        \includegraphics[width=0.24\textwidth]{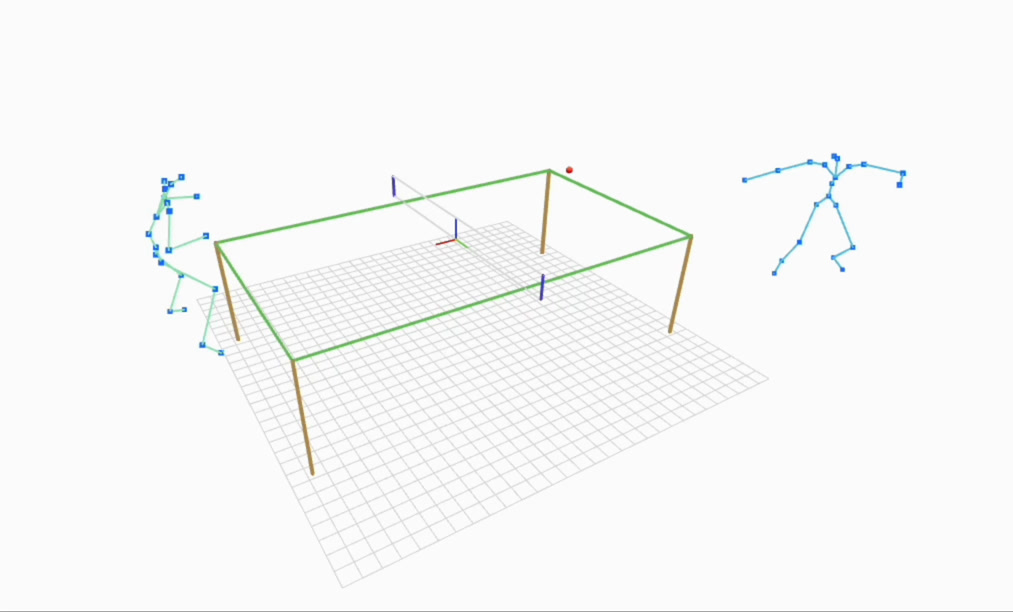} &
        \includegraphics[width=0.24\textwidth]{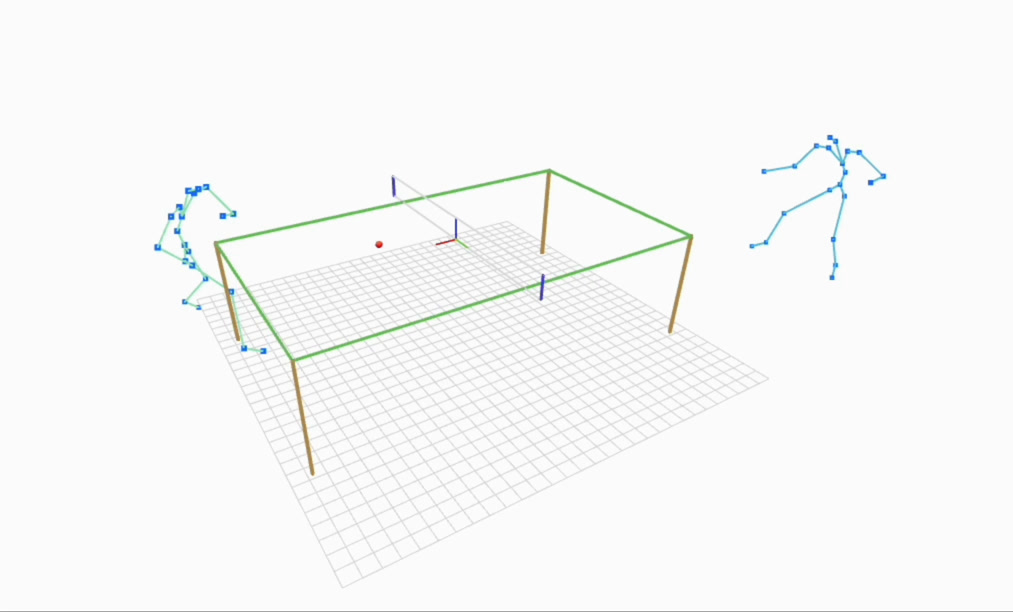} &
        \includegraphics[width=0.24\textwidth]{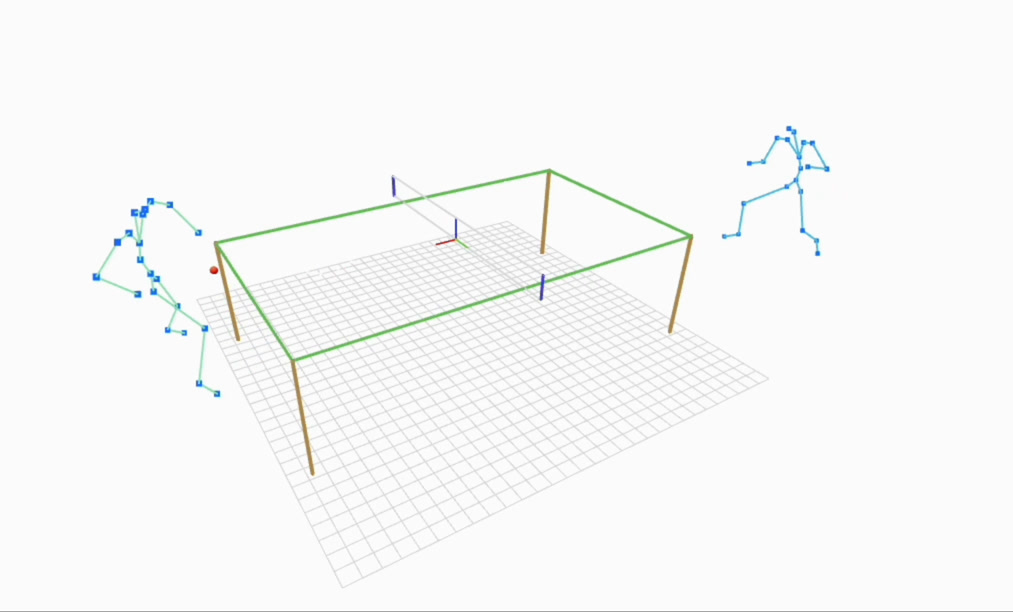} &
        \includegraphics[width=0.24\textwidth]{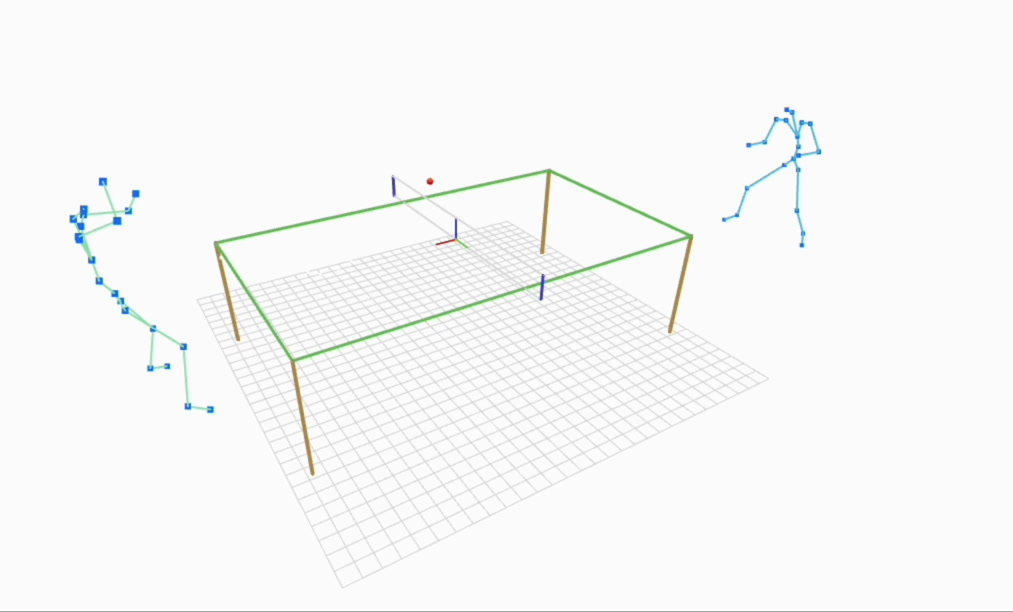} \\

        \includegraphics[width=0.24\textwidth]{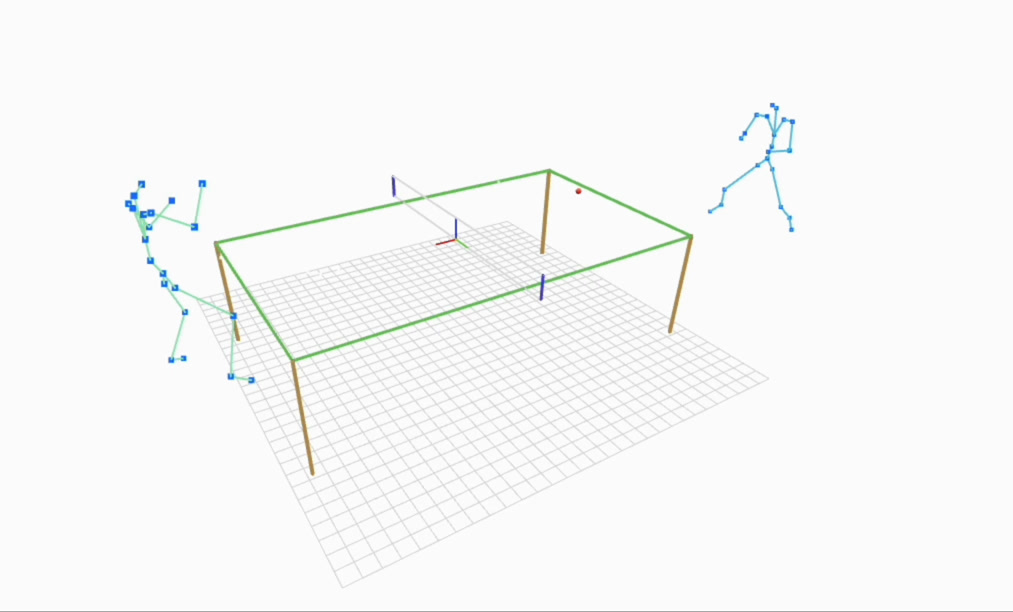} &
        \includegraphics[width=0.24\textwidth]{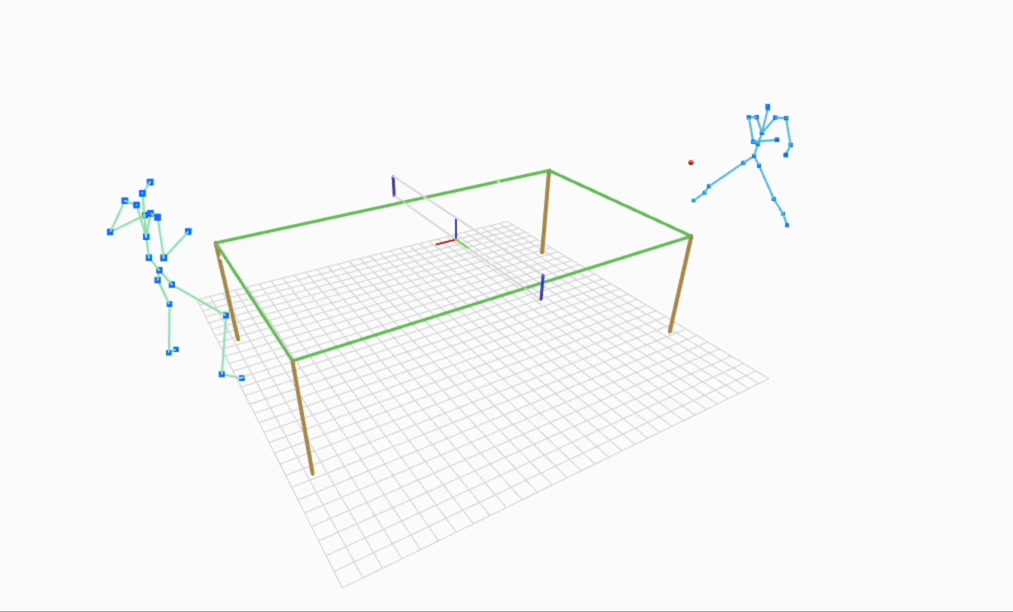} &
        \includegraphics[width=0.24\textwidth]{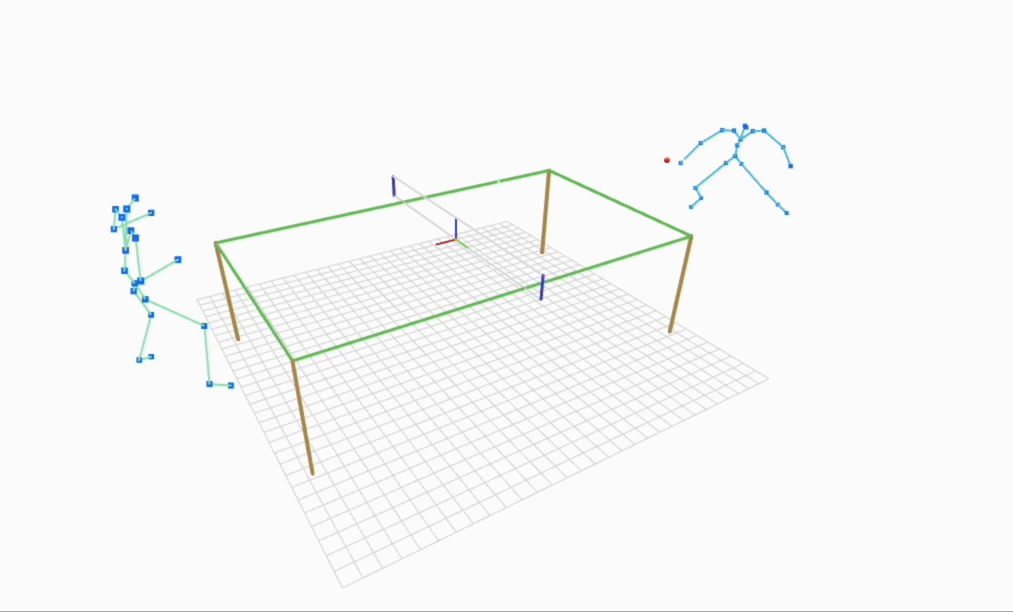} &
        \includegraphics[width=0.24\textwidth]{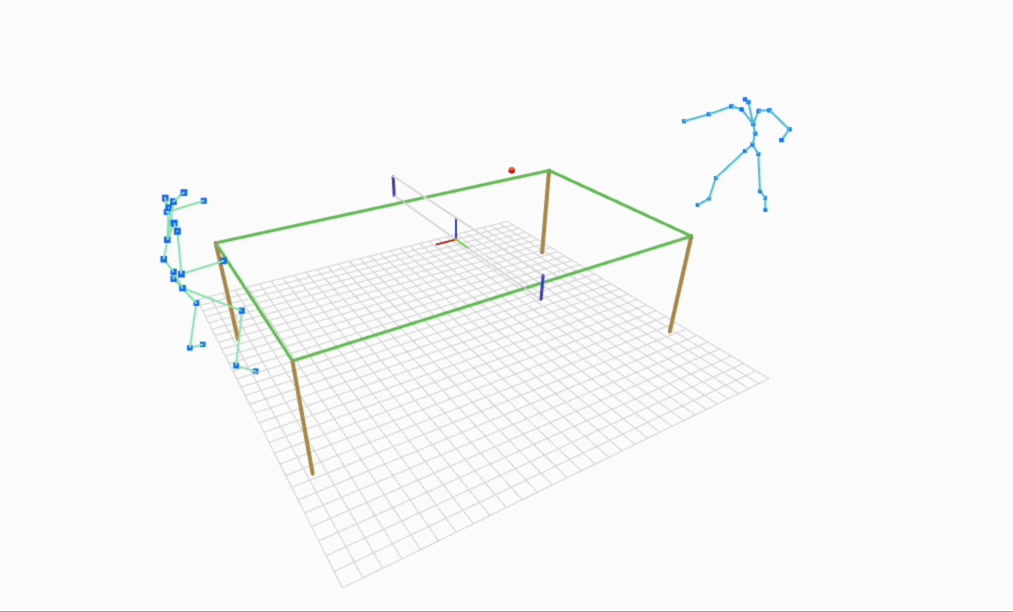} \\

        \includegraphics[width=0.24\textwidth]{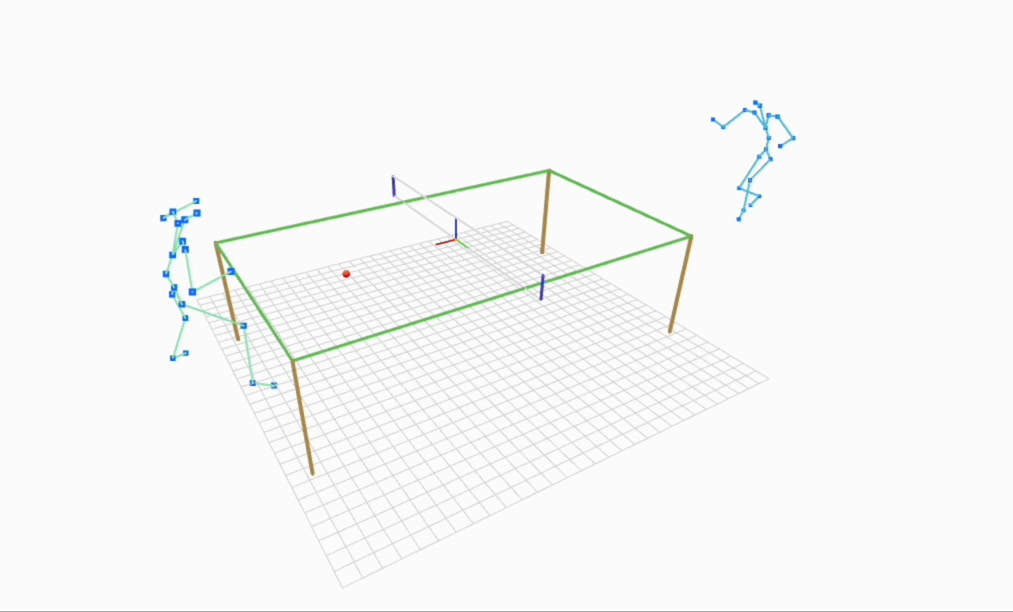} &
        \includegraphics[width=0.24\textwidth]{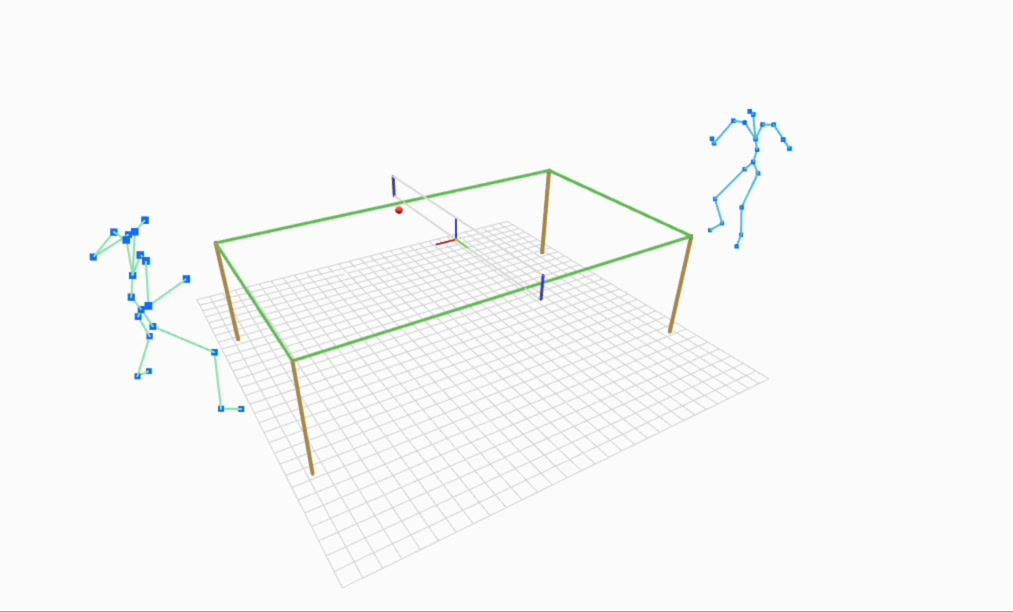} &
        \includegraphics[width=0.24\textwidth]{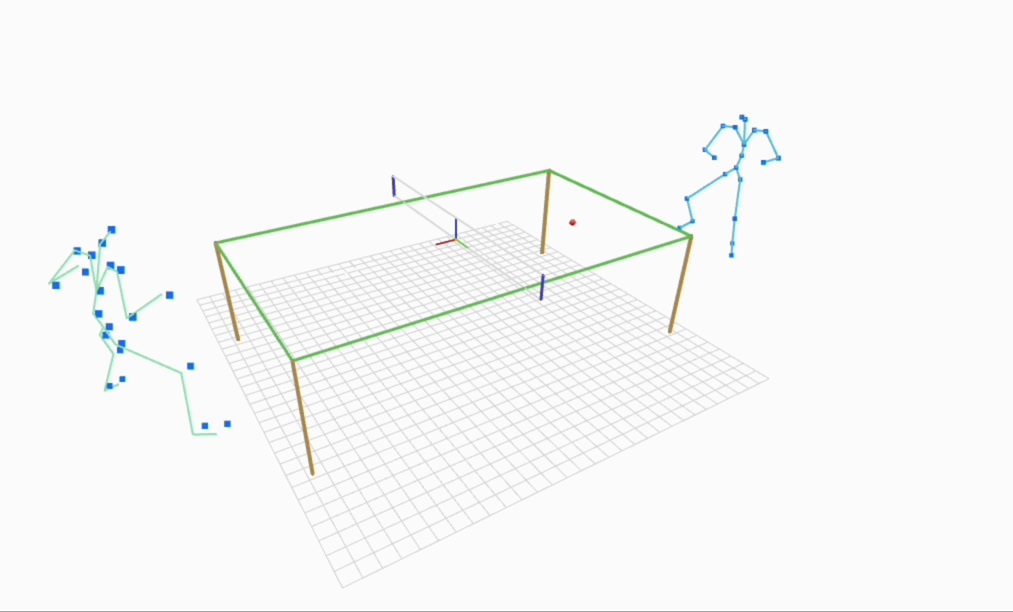} &
        \includegraphics[width=0.24\textwidth]{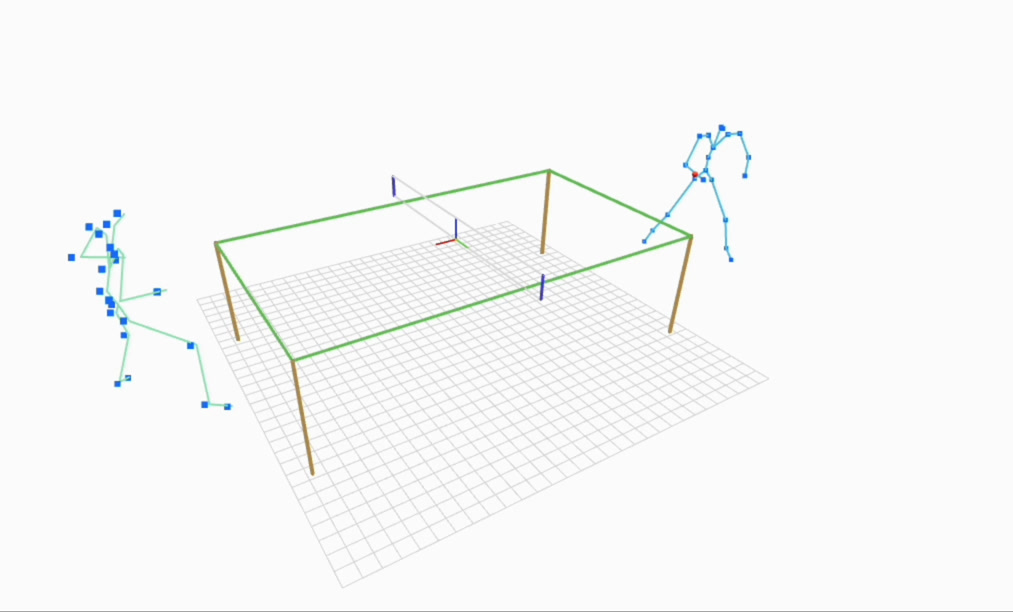} \\

        \includegraphics[width=0.24\textwidth]{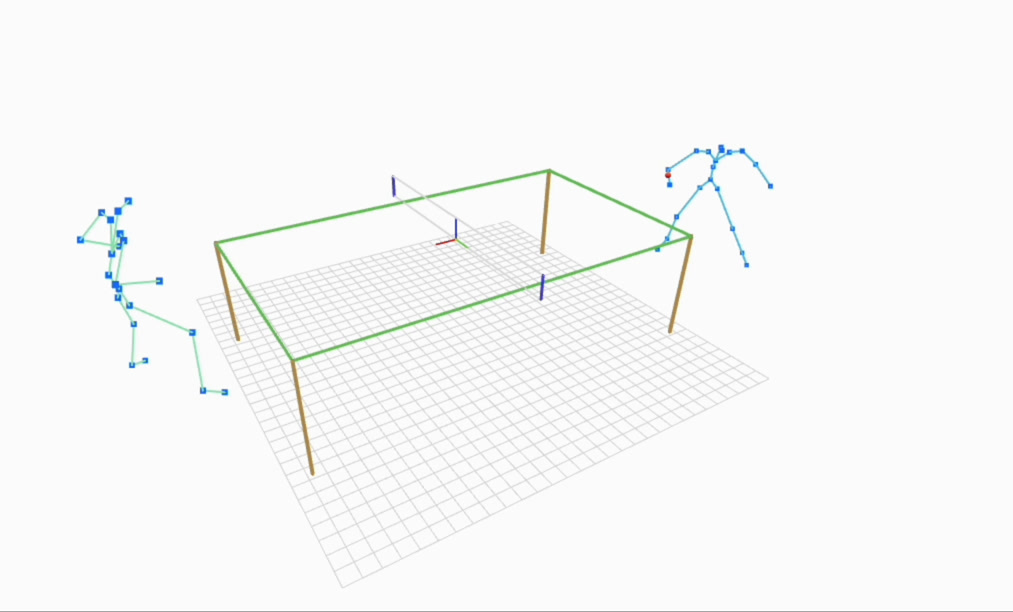} &
        \includegraphics[width=0.24\textwidth]{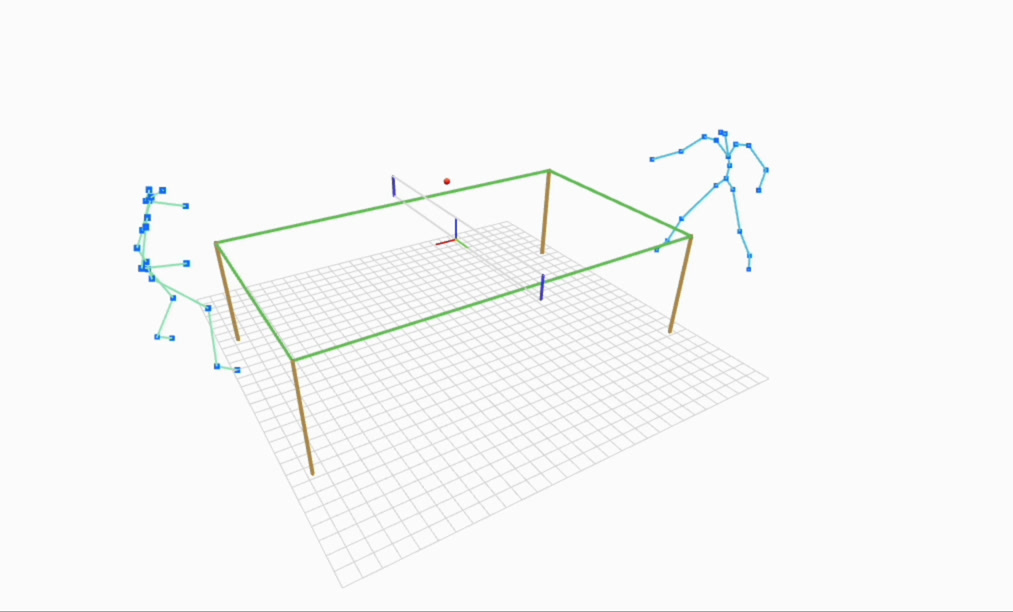} &
        \includegraphics[width=0.24\textwidth]{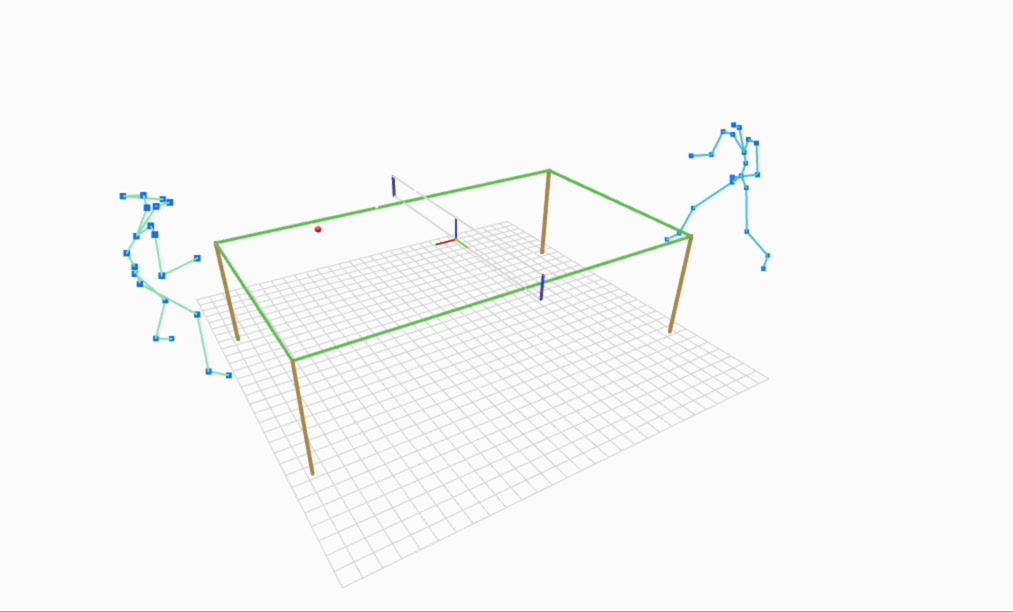} &
        \includegraphics[width=0.24\textwidth]{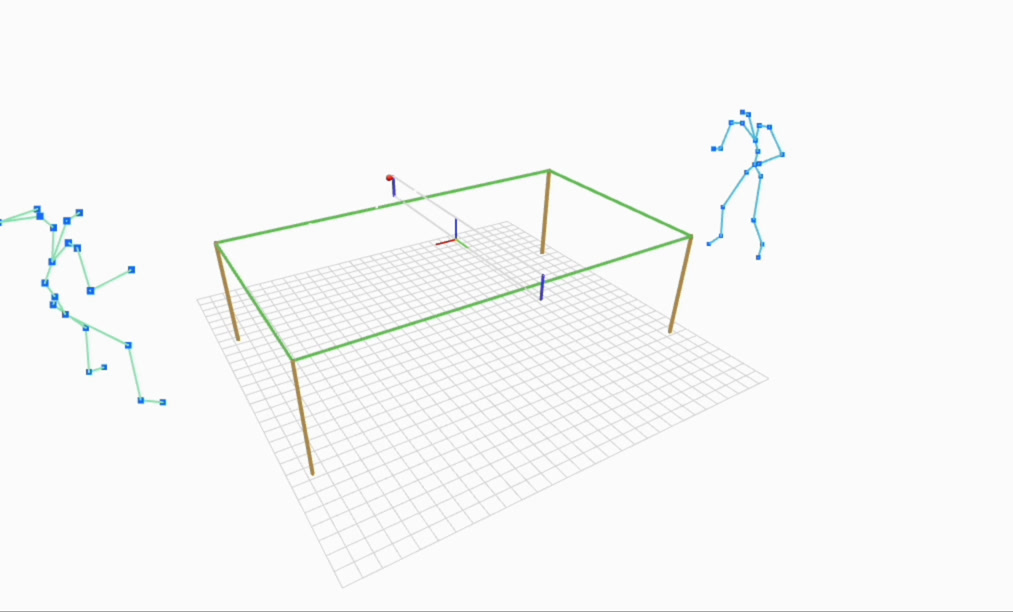} \\

        \includegraphics[width=0.24\textwidth]{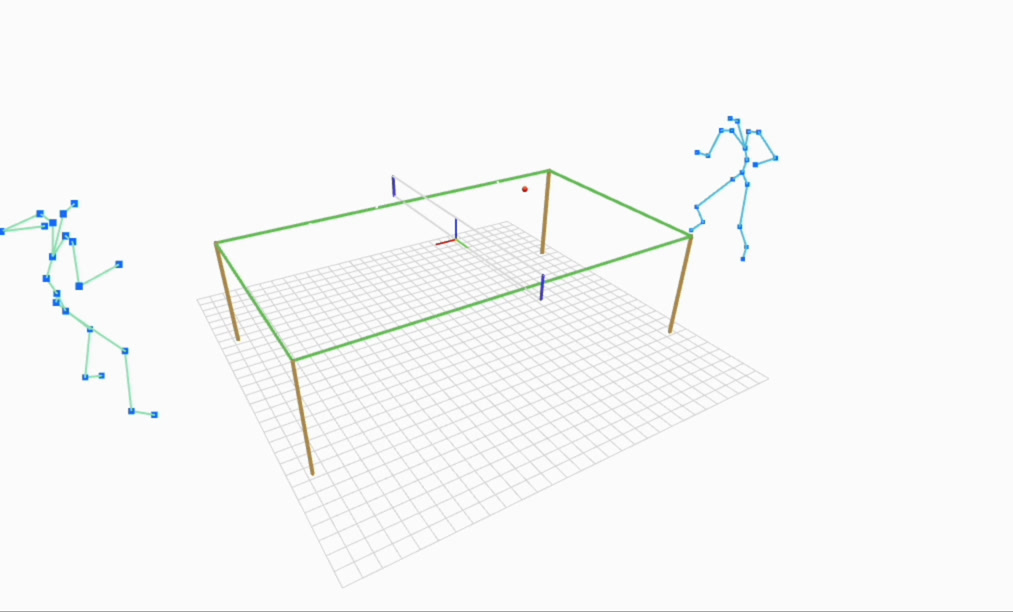} &
        \includegraphics[width=0.24\textwidth]{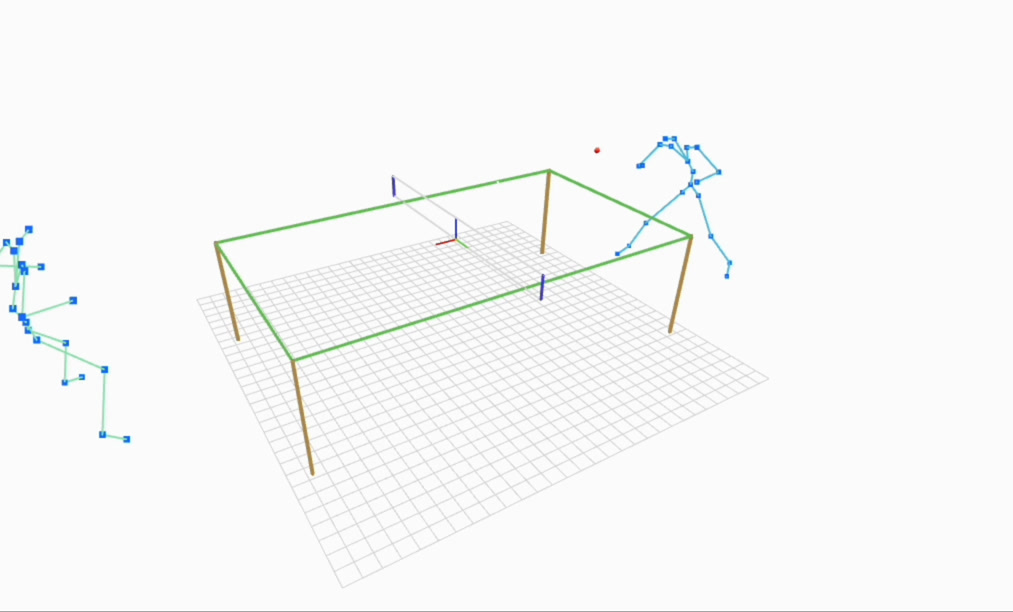} &
        \includegraphics[width=0.24\textwidth]{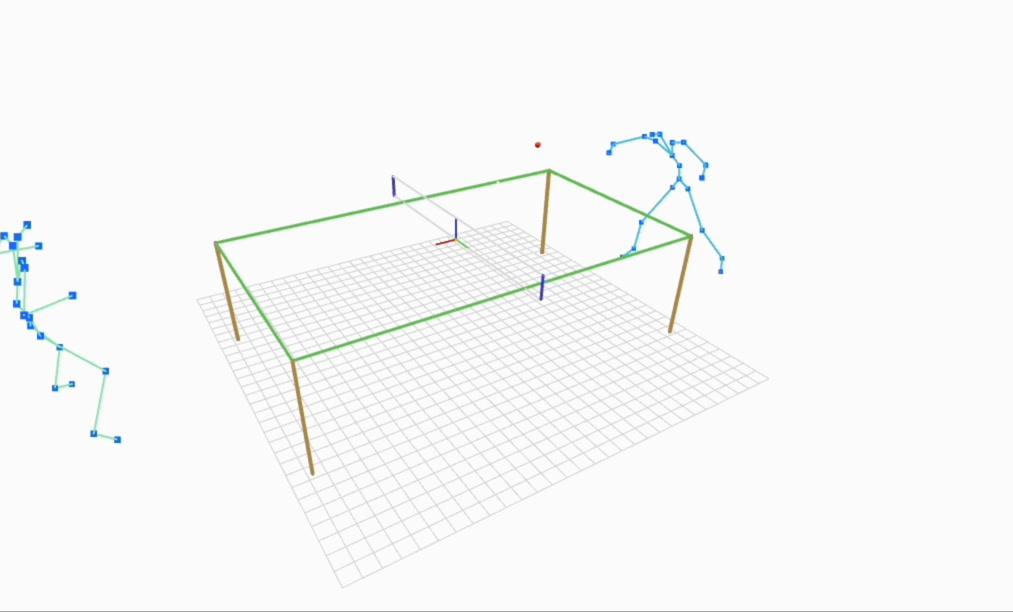} &
        \includegraphics[width=0.24\textwidth]{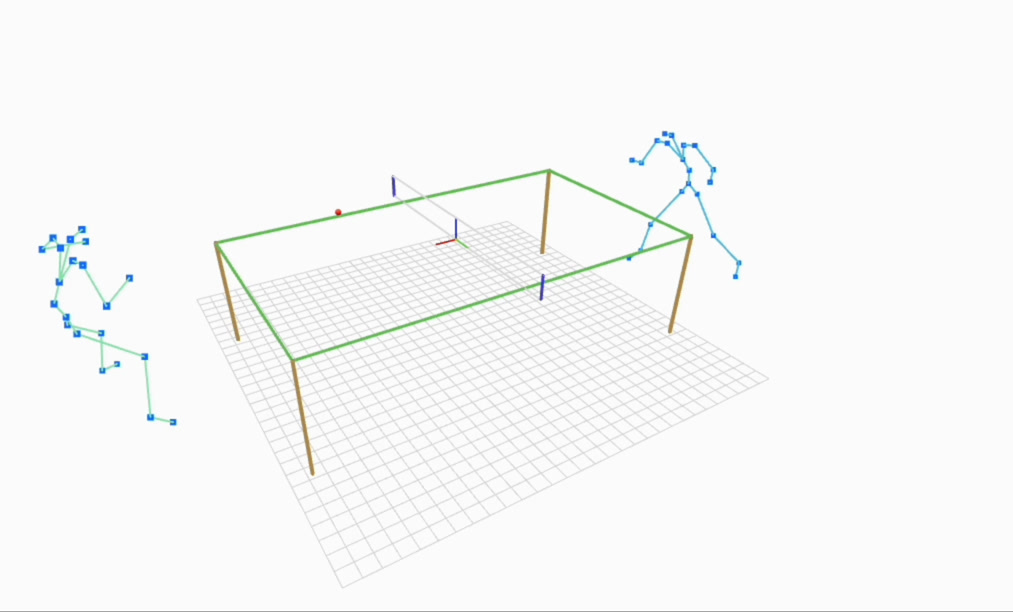} \\

        \includegraphics[width=0.24\textwidth]{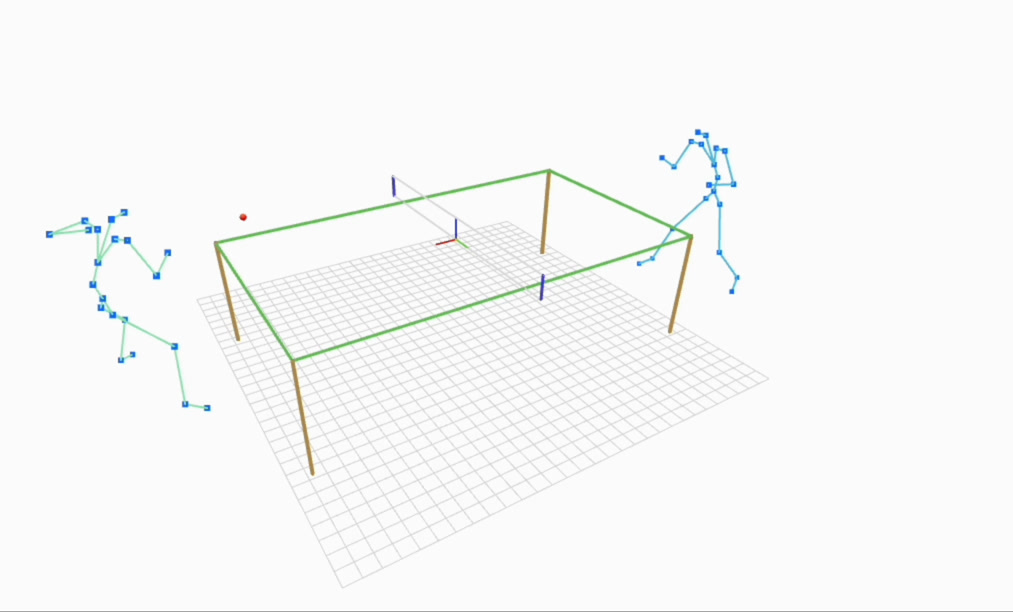} &
        \includegraphics[width=0.24\textwidth]{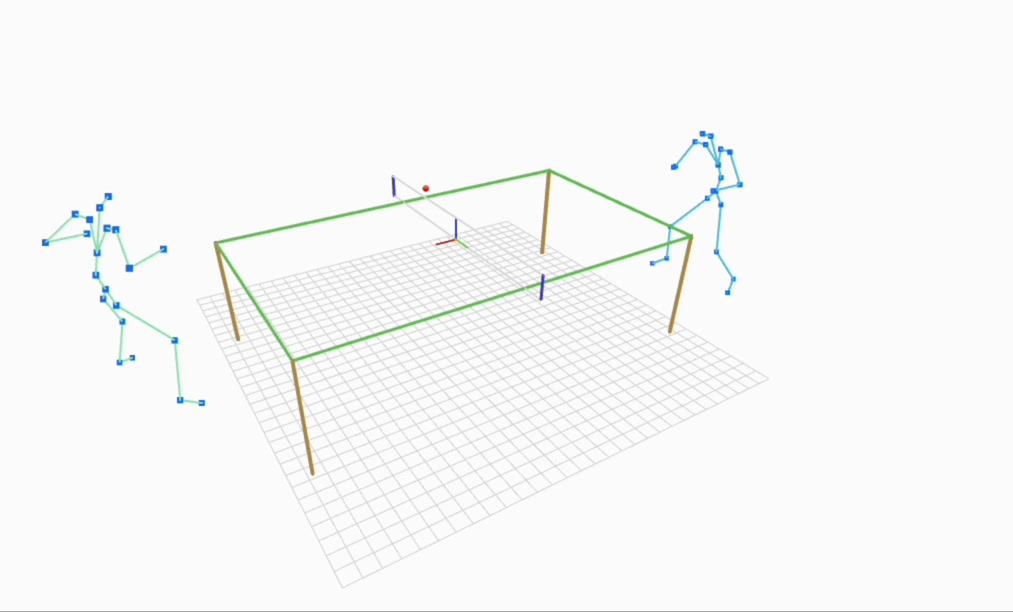} &
        \includegraphics[width=0.24\textwidth]{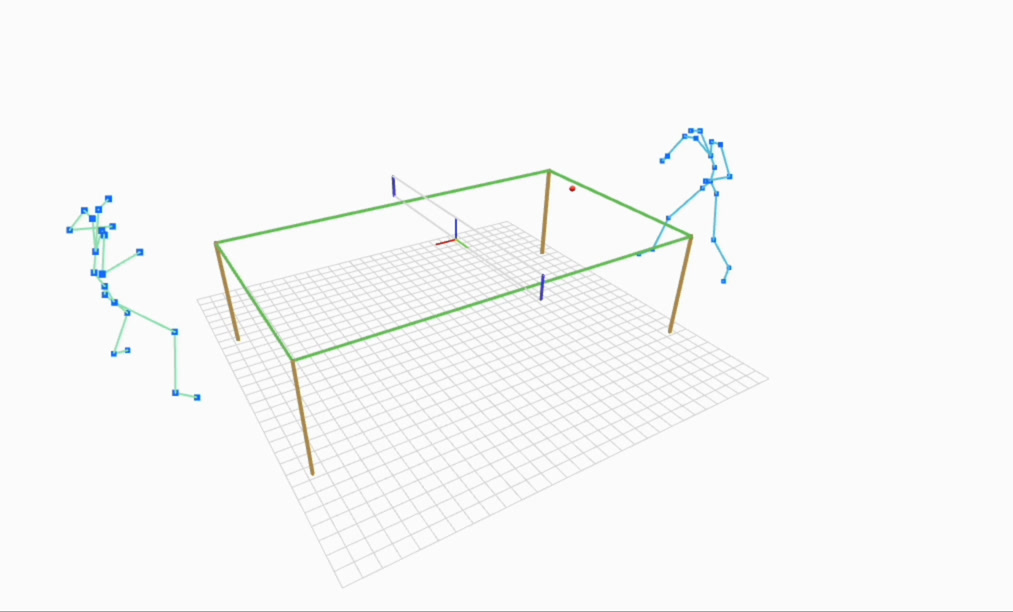} &
        \includegraphics[width=0.24\textwidth]{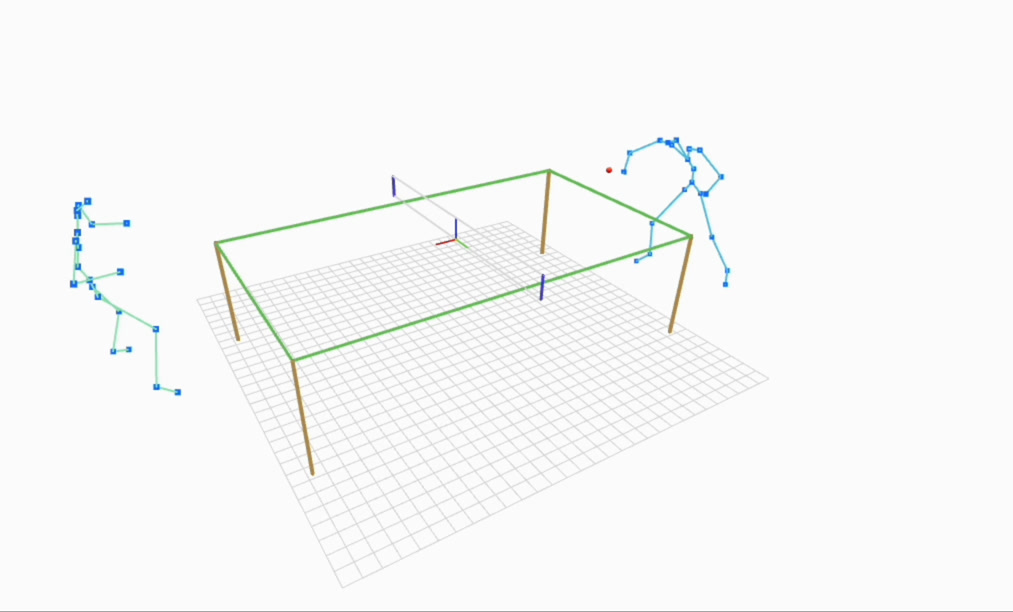} \\
    \end{tabular}

    \Description{A 6 by 4 grid of frames showing a chronological sequence of a generated table tennis rally. Each frame displays a 3D scene featuring two human skeletons rendered in light blue, positioned on opposite sides of a green wireframe table. A red dot represents the table tennis ball. Reading from left to right and top to bottom, the sequence illustrates the continuous, simulated movement of the skeletal players and the trajectory of the ball back and forth across the net.}
    \caption{One example \textbf{generated sequence} at 30 FPS. We display every 16th frame. The diagram should be read from left to right. The motion of the skeleton and ball is smooth and realistic.}
    \label{fig:supp_qualitative_flow}
\end{figure*}

\end{document}